%% file: draft.tex
\RequirePackage{fix-cm}
\documentclass[twoside,11pt]{svjour3}

\smartqed  %
\usepackage{graphicx}

\usepackage[top=.85in,bottom=.85in,left=.85in,right=.85in]{geometry}

\usepackage{graphicx}
\usepackage{amsmath,amssymb} %
\usepackage{color}

\usepackage{xspace}
\usepackage{multirow}
\usepackage{eucal}

\usepackage[square,sort,comma,numbers]{natbib}

\usepackage{algorithmic}
\usepackage[linesnumbered, algoruled, vlined]{algorithm2e}

\def\sh{{StructHash}\xspace}
\def\shh{{StructHash}\xspace}
\def\cgh{{CGHash}\xspace}

\newcommand{\ie}{i.e.\@\xspace}
\newcommand{\eg}{e.g.\@\xspace}

\newcommand{\etal}{et al.\@\xspace}

\usepackage{hyperref}

\input{macro}

\begin{document}

\title{Structured Learning of Binary Codes with  Column Generation}

\author{Guosheng Lin        %
       \and
       Fayao Liu
       \and Chunhua Shen\thanks{Corresponding author: C. Shen. E-mail: chunhua.shen@adelaide.edu.au).}
       \and Jianxin Wu
       \and
        Heng Tao Shen
       }

  \institute{G. Lin, F. Liu and C. Shen \at
        The University of Adelaide, Australia
		\and
        J. Wu \at
        Nanjing University, China
        \and
        H. T. Shen \at
        The University of Queensland, Australia
}

\date{Received: date / Accepted: date}

\maketitle

\begin{abstract}%

Hashing methods aim to learn a set of hash functions which map the original features to compact binary codes with similarity preserving in the Hamming space.
Hashing has proven a valuable tool for large-scale information retrieval.
We propose a column generation based binary code learning framework
for data-dependent hash function learning.
Given a set of triplets that encode the pairwise similarity comparison information, our column generation based method learns hash functions that preserve the relative comparison relations within the large-margin learning framework.
Our method iteratively learns the best hash functions during the column generation procedure.

Existing hashing methods optimize over simple objectives such as
the reconstruction error or graph Laplacian related loss functions, instead of the
performance evaluation criteria of interest---multivariate performance measures
such as the AUC and NDCG.
Our column generation based method can be further generalized from the triplet loss to a general structured learning based framework that allows one to directly optimize multivariate performance measures.
For optimizing general ranking measures,
the resulting optimization problem can involve exponentially or infinitely many
variables and constraints, which is more challenging than standard structured output
learning.
We use a combination of
column generation and cutting-plane techniques to solve the optimization problem.
To speed-up the training we further explore stage-wise training and propose to use a simplified NDCG loss for efficient inference.
We demonstrate the generality of our method by applying it to ranking prediction and image retrieval, and show that it outperforms a few state-of-the-art hashing methods.

\end{abstract}


\input{intro}

\input{method_cgh}

\input{method}

\input{exp}

\section{Conclusion}
We have developed a flexible column generation based hashing framework that is able to optimize general multivariate ranking measures as well as the triplet loss.
We have shown that column generation optimization is able to learn high-quality binary codes for supervised hashing.
The fact that the proposed method for optimizing ranking loss usually outperforms comparable hashing approaches is to be expected, as it more directly optimizes
the required loss function.
It is anticipated that the success of the approach may lead to a range of new
hashing-based applications with task-specified targets.
We also present two extensions of learning framework for efficient learning which are based on stage-wise training and using the proposed simplified NDCG for efficient ranking inference.

\vskip 0.2in

\bibliographystyle{unsrt}

\bibliography{structhash}

\end{document}

%% file: macro.tex
\newcommand{\cC}{\mathcal{C}}

\newcommand{\cS}{\mathcal{S}}
\newcommand{\cW}{\mathcal{W}}
\newcommand{\cX}{\mathcal{X}}

\newcommand{\cY}{\mathcal{Y}}

\newcommand{\cD}{\mathcal{D}}

\newcommand{\cT}{\mathcal{T}}

\def\argmax{\operatorname*{argmax\,}}

\def\exp{\operatorname*{exp\,}}
\def\log{\operatorname*{log\,}}
\def\sign{\operatorname*{sign\,}}
\def\inf{\operatorname*{inf\,}}
\def\sup{\operatorname*{sup\,}}

\def\bmu{{\boldsymbol \mu}}

\def\blambda{{\boldsymbol \lambda}}
\def\balpha{{\boldsymbol \alpha}}
\def\bbeta{{\boldsymbol \beta}}

\def\brho{ {\boldsymbol{\rho} } }
\def\bxi{ {\boldsymbol{\xi} } }

\def\v{{\boldsymbol v}}
\def\c{{\boldsymbol c}}
\def\x{{\boldsymbol x}}
\def\w{{\boldsymbol w}}
\def\y{{\boldsymbol y}}

\def\bx{{\x}}
\def\by{{\y}}

\def\bv{{\v}}

\newcommand{\comment}[1]{}

\def\T{{\!\top}}
\def\Real{\mathbb{R}}
\newcommand{\st}{{{\rm s.t.}\;\;}\xspace}
\newcommand{\one}{{\mathbf{1}}}
\newcommand{\zero}{{\mathbf{0}}}

\def\wnorm{\one^\T \w}
\def\wnorminf{ {\| \w \|}_{\infty} }

\def\x{{\boldsymbol x}}
\def\w{{\boldsymbol w}}

\def\wlset{{\cC}}

\def\hs{\Phi}
\def\bh{\hs}

\def\dhamm{d_{hm}}
\def\dhs{{\delta}{\bh}}
\def\dh{{\delta}{h}}

\newcommand{\figcenter}[1]{\raisebox{-0.5\height}{#1}}
\newcommand{\colseperator}{\hspace*{.4in}}

\def\wstructs{{\Psi }}

\def\dwstructs{{\delta}{\wstructs}}

\def\loss{{\Delta}}

\def\oneslack{$ 1 $-slack\xspace}

\newcommand{\nslack}{$n$-slack\xspace}

\def\hx{\widetilde{\boldsymbol x} }

%% file: intro.tex
\section{Introduction}
\label{sec:Intro}

    The ever increasing volumes of imagery available, and the benefits
    reaped through the interrogation of large image datasets, have
    increased enthusiasm for large-scale approaches to vision.  One of
    the simplest, and most effective means of improving the scale and
    efficiency of an application has been to use hashing to
    pre-process the data
\citep{kulis2009learning,weiss2008spectral,ICCV13Lin,CVPR13aShen,liu2011hashingGraphs,CVPR14Lin}.
Hashing methods construct a
set of hash functions that map the original features
into compact binary code. Hashing enables fast nearest
neighbor search by using look-up tables or Hamming
distance based ranking. Compact binary code
are also extremely efficient for large-scale data storage or
network transfer. Applications include image retrieval
\citep{torralba2008small,wang2010semi}, image matching \citep{Strecha2012}, 
large-scale object detection \citep{fastdection}, etc.

	Hash function learning aims to preserve some notion of similarity.
	We first focus on a type of similarity information that is generally presented in a
    set of triplet-based relations. 
    The triplet relations used for training can be generated in an either supervised or unsupervised fashion.  
    The fundamental idea is to learn hash functions such
    that the Hamming distance between two similar data points 
    is smaller than that between two dissimilar data points.
    This type of relative similarity comparisons have been successfully
    applied to learn quadratic distance metrics \citep{schultz2004learning,shen12jmlr}.
    Usually this type of similarity relations do not require explicit
    class labels and thus are easier to obtain
    than either the class labels or the actual distances between data points.
    For instance, in content based image retrieval,
    to collect feedback, users may be required to report whether
    one image looks more similar to another image than it is to a third one. 
    This task  is typically much easier than to label each
    individual image.
    Formally, let $ \bx$ denote one data point, 
    we are given a set of triplets:
    \begin{equation}
	    \cT = \{( i, j, k ) \, | \, d(\bx_i,\bx_j)< d(\bx_i, \bx_k) \},
    \end{equation}
    where $d(\cdot, \cdot)$ is some distance measure (e.g., Euclidean
    distance in the original space; or semantic similarity measure
    provided by a user).
    As explained, {\em one may not explicitly know $d(\cdot, \cdot)$};
    instead, one may only be able to provide sparse similarity
    relations.
    Using such a set of
    constraints, we formulate a large-margin learning problem which is a convex optimization problem but with an exponentially large number of
    variables. Column generation is thus employed to efficiently solve the formulated optimization problem.

    Our column generation based method can be further generalized to optimize more general multivariate ranking measures, not limited to the simple triplet loss.
    Depending on applications, specific measures are used to
    evaluate the performance of the generated hash codes. For example,
    information retrieval and ranking criteria \citep{mcfee10_mlr} such as
    the Area Under the ROC Curve (AUC) \citep{Joachims2005},
    Normalized Discounted Cumulative Gain (NDCG)
    \citep{JarvelinK00},
    Precision-at-K, Precision-Recall and Mean Average Precision (mAP) have been widely
    adopted to evaluate the success of hashing methods. 
    However, to date,
    most hashing methods are usually learned by optimizing simple
    errors such as the reconstruction error (e.g., binary
    reconstruction embedding hashing  \citep{kulis2009learning}), 
    the graph Laplacian related loss
    \citep{zhangSTHs,liu2011hashingGraphs,weiss2008spectral}, or the pairwise similarity loss \citep{KSH}.
    To our knowledge, none of the existing hashing methods has
    tried to learn hash codes that {\em directly} optimize a multivariate
    performance criterion. In this work, we seek to reduce the discrepancy between existing learning criteria and
the evaluation criteria (such as retrieval quality measures).

    The proposed framework accommodates various complex multivariate measures 
    as well as the simple triplet loss.
    By observing that the hash codes learning problem is essentially
    an information retrieval problem, various ranking loss functions
    can and should be applied, rather than merely pairwise distance comparisons in the triplet loss.
    This framework also allows to introduce more general definitions of
    ``similarity'' to hashing beyond existing ones.

    In summary, our main contributions are as follows.
    \begin{enumerate}
    \item
    We explore the column generation optimization technique and the large-margin learning framework for hash function learning.
    We first propose a learning framework to optimize the conventional triplet loss, which is referred to as Column Generation Hashing (CGHash). 
    Then we extend this framework to optimize complex multivariate evaluation measures 
    (e.g., ranking measures: AUC and NDCG), which is referred to as StructHash.
    This framework, {\em for the first time}, exploits the gains made in structured output learning for
    the purposes of hashing.  

    \item
    In our column generation based method for optimizing ranking measures, 
    we develop column generation and cutting-plane algorithms
    to efficiently solve the resulting optimization problem,
    which may involve exponentially or even infinitely many variables
    and constraints.

	\item
	We propose a new stage-wise training protocol to speedup the training procedure of the proposed StructHash.
	With this stage-wise learning approach, we are able to use the efficient unweighted hamming distance on the learned hash functions. 
	Experimental evaluations show that the stage-wise learning approach brings orders of magnitude speedup in training 
	while being equally or even more effective in retrieval accuracy.

	\item
	The proposed StructHash learning procedure requires an inference algorithm for finding the most violated ranking,
	which is the most time consuming part in the training procedure.
 	We propose to optimize a new ranking measure, termed as  Simplified NDCG (SNDCG), 
 	which allow efficient inference in the training procedure, 
 	and thus significantly speedup the training. 
 	Experimental results show that optimizing this new ranking measure leads to around $2$ times faster inference.

    \item
    Applied to ranking prediction for image retrieval, the proposed method demonstrates
    state-of-the-art performance on hash function learning.

    \end{enumerate}

We have released the training code of our CGHash \footnote{CGHash is available at https://bitbucket.org/guosheng/column-generation-hashing}
and \sh \footnote{\sh is available at https://bitbucket.org/guosheng/structhash} on-line, which also includes the recent extensions of \sh on efficient stage-wise training and using simplified NDCG loss.

    This paper is organized as follows: we first present our method 
    for optimizing triplet loss in Sec. \ref{sec:cgh}, then we generalize our method for optimizing complex ranking loss in Sec. \ref{sec:sh}, finally we present empirical evaluation in Sec. \ref{sec:exp}.

\section{Related work}  
Our method provides a unified framework of the column generation technique and large-margin based structured learning for binary code learning. 
Preliminary results of our work appeared in \cite{ICML13a} and \cite{LinEccv14}. 
In the following, we give a brief introduction to the most closely related work.

\paragraph{Binary code learning}
Compact binary code learning, or hashing aims to preserve some notation of similarity in the Hamming space. These methods can be roughly categorized into unsupervised and (semi-) supervised approaches. 
Unsupervised methods attempt to preserve the similarities calculated in the original feature space.  
Examples fall into this category are locality sensitive    hashing (LSH) \citep{Gionis1999}, 
spectral hashing (SPH) \citep{weiss2008spectral}, anchor graph hashing (AGH) \citep{liu2011hashingGraphs}, 
iterative quantization hashing (ITQ) \citep{gong2012iterative}. 
Specifically, LSH \citep{Gionis1999} uses random projection to generate binary codes; 
SPH \citep{weiss2008spectral} aims to preserve the neighbourhood relation by optimizing the Laplacian affinity;    
AGH \citep{liu2011hashingGraphs} makes the original SPH much more scalable;
ITH \citep{gong2012iterative} first performs linear dimensionality reduction and then conduct binary quantization in the resulting space.

As for the supervised approaches,  they aim to preserve the label based similarities. 
Binary reconstruction embedding (BRE) \citep{kulis2009learning} aims to minimize the expected distances;    
semi-supervised sequential projection learning hashing (SPLH) \citep{wang2010semi} enforces the smoothness of similar data  points and the separability of dissimilar data points;  
kernelized LSH, proposed by Kulis and Grauman \citep{KLSH}, 
randomly samples training data as support vectors, and randomly draws the dual
coefficients from a Gaussian distribution. 
Later on, Liu et al.\citep{KSH} extended kernelized LSH to
kernelized supervised hashing (KSH).
Lin \etal \citep{ICCV13Lin,CVPR14Lin} present a general two step framework for hashing learning.
In \cite{Norouzi11}, Norouzi \etal propose a latent variables based structured SVM formulation to optimize a  hinge-like loss function. 
Their method attempts to preserve similarities between pairs of training exemplars.
They further generalize the method in \citep{Norouzi11} to optimize a triplet ranking loss designed to preserve relative similarities \citep{Norouzi12}.
Our method belongs to supervised approaches. 
Unlike existing approaches, we formulate the binary code learning as a structured output 
learning problem, in order to directly optimize a wide variety of ranking evaluation measures.
The hashing method in \citep{Wang15} proposes to optimizes the NDCG ranking loss with a gradient decent method,
which comes out after the publication of our preliminary version of \sh in \cite{LinEccv14}.

\paragraph{Learning to rank}
Our method is primarily inspired by recent advances in metric learning for ranking \citep{mcfee10_mlr,Shalit12,lim2014efficient}.  
In \citep{mcfee10_mlr}, McFee \etal propose a structured SVM based method to directly optimize several different ranking measures. However, it can not be scaled to large, high-dimensional datasets due to the spectral decomposition at each iteration and the expensive constraint generation step.  Later on, Shalit \etal \citep{Shalit12} propose a scalable method for optimizing a ranking loss, though they only consider the Area Under the ROC Curve (AUC) loss.
In \citep{lim2014efficient}, Lim \etal propose to optimize a Mahalanobis metric with respect to a top-heavy ranking loss, \ie, the Weighted Approximate Pairwise Ranking (WARP) loss \citep{weston2010large}.
We extend the structured learning based ranking optimization to hash function learning.

\paragraph{Column generation}
Column generation is widely applied in boosting methods \cite{lpboost,shen10,Shen2014SBoosting}.
LPBoost \citep{lpboost} is a linear programming boosting method that iteratively learn weak classifiers to form a strong classifier. 
StructBoost \citep{Shen2014SBoosting} provides a general structured learning framework using column generation for structured prediction problems.
We here exploit the column generation technique for hash functions learning.

%% file: method_cgh.tex
\section{Hashing for optimizing the triplet loss} %

\label{sec:cgh}

    We first describe our column generation based approach for optimizing the triplet loss.
    We refer to this approach as \cgh.
        Given a set of training examples $\cX=\{\bx_1, \bx_2, \dots, \bx_n\} \subset
    \Real^d$, the task is to learn a set of hash functions
        $ [ h_1( \bx  ), h_2 ( \bx), \dots, h_m (\bx)  ]  $. 
        The domain of hash functions is denoted by $\wlset$: $h(\cdot) \in \wlset$.
    The output of one hash function is a binary value: $h(\bx) \in \{0, 1\}$.
        With the learned functions, an input $ \bx $ is mapped into a
        binary code of length $ m $.
        We use $ \hx \in
        \{0, 1\}^m$ to denote the
        hashed values of $  \bx  $, i.e.,
        \begin{equation}
         \hx  =  [ h_1( \bx  ), h_2( \bx ), \dots, h_m(\bx)     ]^\T.
        \label{EQ:Hash}
        \end{equation}
    The resulting binary code are supposed to preserve the similarity information.
    Formally, suppose that we are given a set of triplets
    $\cT=\{(i, j, k)\}$ as the supervision information for learning.
    These triplets encode the similarity comparison information
    in which the distance/dissimilarity
    between $\bx_{i}$ and ${\bx}_{j} $ is smaller than that
    between $\bx_{i}$ and ${\bx}_{k} $.
    We define the weighted Hamming distance for the learned binary codes as:
\begin{equation}
  \label{eq:whamming}
        \dhamm(\bx_i, \bx_j; \w) = \sum_{r=1}^{m}w_{r}|h_{r}(\bx_i)-h_{r}(\bx_j)|,
\end{equation}
where $w_{r}$ is a non-negative weighting coefficient associated with the $r$-th hash function.
        Such weighted hamming distance is used in 
        multi-dimension spectral hashing \citep{MDSH}.
        It is expected that after hashing, the distance between
        relevant data points should be smaller than the distance between
        irrelevant data points, that is
\begin{equation}
    \dhamm (\x_i, \x_j) <  \dhamm(\x_i, \x_k).
\end{equation}
For notational simplicity, we define
\begin{equation}
  \dh(i,j,k)= | h(\bx_i) - h(\bx_k) | - | h(\bx_i) - h(\bx_j) |
\end{equation}
and
\begin{equation}
  \label{eq:delta-whamming}
  \dhs(i, j, k)=[\dh_1(i,j,k), \dh_2(i,j,k), \dots, \dh_m(i,j,k)].
\end{equation}
With the above definitions, the weighted Hamming distance comparison of a triplet can be written as:
\begin{equation}
\dhamm( \bx_{i}, \bx_{k}) - \dhamm( \bx_{i}, \bx_{j} )
 = \w^{\T}\dhs(i, j, k). 
\end{equation}
We propose a large-margin learning framework to optimize for the
weighting parameter $ \w $ as well as the hash functions.
In what follows, we describe the details of our hashing algorithm using different types of
convex loss functions and regularization norms. 

\subsection{Learning hash functions using column generation}
\label{sec:cg_learning}

As a starting point, we first discuss using the squared hinge loss function and $\ell_1$ norm regularization for hash function learning. 
Using the squared hinge loss, we define the following large-margin
optimization problem: %
\begin{align}
\label{eq:cgh-shinge-l1-tmp} %
  \min_{\w, \bxi} \;\; & \wnorm + C \sum_{(i,j,k) \in \cT}\xi_{(i,j,k)}^2 \\
  \st \;\; 
  & \forall (i, j, k)\in \cT: \notag \\
  & \dhamm( \bx_{i}, \bx_{k}; \w) - \dhamm( \bx_{i}, \bx_{j}; \w ) \geq 1 - \xi_{(i,j,k)},  \notag \\
  &  \w \geq \zero, \;\; \bxi \geq \zero. \notag
\end{align}
Here we have used the $ \ell_1 $ norm on $\w$ as the regularization term
to control the complexity of the learned model;
the weighting vector $\w$ is defined as:
\begin{equation}
  \w = [w_{1}, w_{2}, \ldots, w_{m}]^{\T};
\end{equation}
$\bxi$ is the slack variable; $C$ is a  parameter
controlling the trade-off between the training error and model complexity.
With the definition of weighted Hamming distance in \eqref{eq:whamming} 
and the notation in \eqref{eq:delta-whamming}, 
the optimization problem in \eqref{eq:cgh-shinge-l1-tmp} can be rewritten as: 
\begin{align}
\label{eq:cgh-shinge-l1} 
  \min_{\w, \bxi} \;\; & \wnorm + C \sum_{(i,j,k) \in \cT}\xi_{(i,j,k)}^2 \\
  \st \;\;
  & \forall (i, j, k)\in \cT: \;\;
   \w^\T \dhs (i,j,k) \geq 1 - \xi_{(i,j,k)} \notag \\
  &  \w \geq \zero, \;\; \bxi \geq \zero. \notag
\end{align}
We aim to solve the 
above optimization to obtain the weighting vector $\w$ 
and the set of hash functions $[h_1, h_2, \dots]$. 
If the hash functions are obtained, the optimization 
can be easily solved for $\w$, e.g., using LBFGS-B \citep{lbfgs}. 
In our approach, we apply the column generation technique 
to alternatively solve for $\w$ and learn hash functions.
Basically, we construct a working set of hash functions 
and repeat the following two steps until converge: 
first we solve for the weighting vector 
using the current working set of hash functions, 
and then generate new hash function and add to the working set.
    
Column generation is a technique originally used for large scale
linear programming problems. 
 LPBoost \citep{demiriz2002linear} applies this technique
to design boosting algorithms.  In each iteration, one column---a
variable in the primal or a constraint in the dual problem---is added.
Till one cannot find any 
violating constraints in the dual, the current solution is the optimal solution.
In theory, if we run the column generation  with a
sufficient number of iterations, one can obtain a sufficiently
accurate solution. Here we only need to run a small number of 
column generation iteration (e.g, 60) to learn a compact set of hash functions.

To apply column generation technique for learning hash functions, 
we derive the dual problem of the  optimization in \eqref{eq:cgh-shinge-l1}. 
The optimization in \eqref{eq:cgh-shinge-l1} can be equally written as:
\begin{align}
  \label{eq:cgh-shinge-l1-short} 
  \min_{\w, \brho} \;\; & \wnorm + C \sum_{(i,j,k) \in \cT} 
  \biggr[ \max ( 1-\rho_{(i,j,k)}, \, 0 ) \biggr]^2 \\
  \st \;\; 
  & \forall (i,j,k) \in \cT: \rho_{(i,j,k)}  =  \w^{\T} \dhs (i,j,k), \label{eq:cgh-shinge-con}\\
  & \w \geq \zero. \notag
\end{align}
The Lagrangian of \eqref{eq:cgh-shinge-l1-short} can be written as:
\begin{align}
\label{eq:cgh-lag}
L(\w, \brho, \bmu, \balpha) = & \wnorm + C \sum_{(i,j,k) \in \cT} 
  \biggr[ \max ( 1 - \rho_{(i,j,k)}, \, 0 ) \biggr]^2 \notag \\
  & + \sum_{(i,j,k) \in \cT }\mu_{(i,j,k)} \biggr[ \rho_{(i,j,k)} - \w^{\T} \dhs (i,j,k) \biggr] - \balpha^\T \w, 
\end{align}
where $\bmu$, $\balpha$ are Lagrange multipliers and $\balpha \geq \zero$.
For the optimal primal solution, the following must hold: 
$\frac{\partial L}{\partial \w} = \zero$ and $\frac{\partial L}{\partial \brho} = \zero$.
Therefore we have:
\begin{align}
\label{eq:cgh-deriv-w}
\frac{\partial L}{\partial \w} = \zero & \Longrightarrow \one - \sum_{(i,j,k) \in \cT}  
 \mu_{(i,j,k)} \dhs (i,j,k) - \balpha = \zero.
\end{align}
\begin{align}
\label{eq:cgh-deriv-rho}
\frac{\partial L}{\partial \rho_{(i,j,k)}} = 0 &\Longrightarrow -2C \max ( 1 - \rho_{(i,j,k)}, \, 0 ) + \mu_{(i,j,k)}=0  \notag \\
&\Longrightarrow \mu_{(i,j,k)} = 2C \max ( 1 - \rho_{(i,j,k)}, \, 0 ).
\end{align}
With Eq. \eqref{eq:cgh-deriv-w}, Eq. \eqref{eq:cgh-deriv-rho} and Eq. \eqref{eq:cgh-lag}, we can derive the dual problem as:
\begin{align}
  \label{eq:cgh-shinge-l1-dual} 
  \max_{\bmu} \;\; & \sum_{(i,j,k) \in \cT} \mu_{(i,j,k)} - \frac{\mu_{(i,j,k)}^2}{4 C} \\
  \st \;\; 
  & \forall \, h(\cdot) \in \wlset: \;\; \sum_{(i, j, k)\in \cT} \mu_{(i,j,k)} \dh(i,j,k) \leq 1. \notag
\end{align}
Here $\mu$ is one dual variable, which corresponds to one constraint in \eqref{eq:cgh-shinge-con}.

    The core idea of column generation is to generate a small subset
    of dual constraints by finding the
    most violated dual constraint in \eqref{eq:cgh-shinge-l1-dual}.  
    This process is equivalent to adding primal variables into the
    primal optimization problem \eqref{eq:cgh-general-l1}.  
    Here finding the most violated dual constraint is learning
    one hash function, which can be written as:
\begin{align}
  h^{\star}(\cdot)  =  &
    \argmax_{h(\cdot) \in \wlset} \sum_{(i, j, k)\in \cT} \mu_{(i,j,k)} \dh(i,j,k) \notag \\
    = & \argmax_{h(\cdot) \in \wlset} \sum_{(i, j, k)\in \cT} \mu_{(i,j,k)} 
    \biggr[ | h(\bx_i) - h(\bx_k) | - | h(\bx_i) - h(\bx_j) | \biggr].
    \label{eq:cgh-wl} 
\end{align}
In each column generation iteration,
we solve the above optimization to generate one hash function.

Now we give an overview of our approach.
Basically, we repeat the following two steps until converge: 
\begin{enumerate}
  \item
  Solve the reduced primal problem in \eqref{eq:cgh-shinge-l1-short} 
  using the current working set of hash functions.
  We obtain the primal solution $\w$ and the dual solution $\bmu$ in this step.
  \item
  With the dual solution $\bmu$,
  we solve the subproblem in \eqref{eq:cgh-wl} to learn one hash function,
  and add to the working set of hash functions.
\end{enumerate}
Our method is summarized in Algorithm \ref{alg:cgh}.
We describe more details for running these two steps as follows.

\begin{algorithm}[ht]

\caption{CGHash: Hashing using column generation (with squared hinge loss)}
\label{alg:cgh}

\KwIn{
    training triplets: $ \cT = \{ (i, j, k) \}$, training examples: ${\bx_1, \bx_2, \dots}$,
    the number of bits: $m$.
}

\KwOut{
 Learned hash functions $\{h_{1}, h_{2}, \dots, h_{m} \}$ and the
 associated weights $\w$.
}

{\bf Initialize:} $\bmu \leftarrow \frac{1}{ | \cT | }$.

\For{$r = 1 $ to $m$}{
    find a new hash function $h_{r}(\cdot)$ by solving the
    subproblem: \eqref{eq:cgh-wl}\;
    add $h_{r}(\cdot)$ to the working set of hash functions\;
    solve the primal problem in \eqref{eq:cgh-shinge-l1-short} for $ \w $ (using LBFGS-B\citep{lbfgs}),
    and calculate the dual solution $\bmu$ by \eqref{eq:cgh-kkt-shinge}\;
}

\end{algorithm}

In the first step, we need to obtain the dual solution $\bmu$, 
which is required for solving the subproblem in \eqref{eq:cgh-wl} of the second step to
learn one hash function.
  In each column generation iteration, 
  we can easily solve the optimization in \eqref{eq:cgh-shinge-l1-short} 
  using the current working set of hash functions to obtain the primal solution $\w$,
  for example, using the efficient LBFGS-B solver \citep{lbfgs}.
  According to the Karush-Kuhn-Tucker (KKT) conditions in Eq. \eqref{eq:cgh-deriv-rho}, 
  we have the following relation:
    \begin{equation}
        \label{eq:cgh-kkt-shinge}
        \forall (i,j,k) \in \cT: \mu_{(i,j,k)}^\star = 
          2 C \max \biggr[ 1 - \w^{\star \T} \dhs (i,j,k), 0 \biggr].
    \end{equation}
    From the above,
    we are able to obtain the dual solution $ \bmu^\star $ 
    for the primal solution $ \w^\star $.

In the second step, 
we solve the subproblem in \eqref{eq:cgh-wl} for learning one hash function.
The form of hash function $h(\cdot)$ can be any function that have binary output value. 
When using a decision stump as the hash function,
usually we can exhaustively enumerate all possibility and find the globally best one. 
However, for many other types of hash functions, e.g., perceptron and kernel functions, 
globally solving \eqref{eq:cgh-wl} is difficult. 
In our experiments, we use the perceptron hash function:
\begin{equation}
  \label{eq:cgh-hash-fun}
  h(\x)=0.5( {\rm sign}(\v^\T\x+ b)+1).
\end{equation}
In order to obtain a smoothly differentiable objective function,
we reformulate \eqref{eq:cgh-wl}
into the following equivalent form:
\begin{equation}
  h^{\star}(\cdot)  = 
    \argmax_{h(\cdot) \in \wlset} \sum_{(i, j, k)\in \cT} \mu_{(i,j,k)} 
    \biggr[ \big( h(\bx_i) - h(\bx_k) \big)^2 - \big( h(\bx_i) - h(\bx_j) \big)^2 \biggr].
    \label{eq:cgh-wl-smooth} 
\end{equation}
The non-smooth $\sign$ function in \eqref{eq:cgh-hash-fun} 
brings the difficulty for optimization.
We replace the $\sign$ function by a smooth
sigmoid function, and then locally solve the above optimization \eqref{eq:cgh-wl-smooth} 
(e.g., using LBFGS) for learning the parameters of a hash function.
We can apply a few initialization heuristics for solving \eqref{eq:cgh-wl-smooth}. For example,
similar to LSH, we can generate a number of random planes and choose the best
one, which maximizes the objective in \eqref{eq:cgh-wl-smooth}, as the initial solution. 
We can also train a decision stump by searching a best dimension and threshold to maximize the
objective on the quantized data. Alternatively, we can employ the spectral relaxation method
\citep{liu2011hashingGraphs} which drops the $\sign$ function and solves
a generalized eigenvalue problem to obtain a solution for initialization.
In our experiments, we use the spectral relaxation method for initialization.

\subsection{Hashing with general smooth convex loss functions}

  The previous discussion for squared hinge loss is an example of using smooth convex loss function in our framework.
  To take a step forward, here we describe
  how to incorporate general smooth convex loss functions. 
    We encourage the following constraints to be satisfied as far as possible:
    \begin{equation}
        \forall (i, j, k) \in \cT: 
        \dhamm( \bx_{i}, \bx_{k}) - \dhamm( \bx_{i}, \bx_{j} ) = \w^{\T} \dhs (i,j,k) \geq 0
    \end{equation}
    These constraints do not have to be all strictly satisfied.
    Here we define the margin:
    \begin{equation}
      \rho_{(i,j,k)}  =  \w^{\T} \dhs (i,j,k), 
    \end{equation}
    and we want to
    maximize the margin with regularization.
    We denote by $ f ( \cdot ) $ as a general convex loss function which is assumed to be smooth
     (e.g., exponential, logistic, squared hinge loss).
    Using $\ell_1$ norm for regularization, we define
    the primal optimization problem as: %
\begin{align}
  \label{eq:cgh-general-l1} 
  \min_\w \;\; & \wnorm + C \sum_{(i,j,k) \in \cT} f(\rho_{(i,j,k)}) \\
  \st \;\; 
  & \forall (i,j,k) \in \cT: \rho_{(i,j,k)}  =  \w^{\T} \dhs (i,j,k), \notag \\
  & \w \geq \zero. \notag
\end{align}
    $C$ is a parameter controlling the trade-off between the training error and model complexity.
    Without the regularization, one can always make  $ \w $
    arbitrarily large to make the convex loss approach zero when all constraints are satisfied.

The squared hinge loss which we discussed before is an example of $f(\cdot)$. 
We can easily recover the formulation in \eqref{eq:cgh-shinge-l1-short} for squared hinge loss 
by using the following definition:
\begin{equation}
  \label{eq:cgh-general-shinge}
  f(\rho_{(i,j,k)})=\biggr[ \max ( 1-\rho_{(i,j,k)}, \, 0 ) \biggr]^2.
\end{equation}

For applying column generation, we derive the dual problem of \eqref{eq:cgh-general-l1}.
The Lagrangian of \eqref{eq:cgh-general-l1} can be written as:
\begin{align}
  L(\w, \brho, \bmu, \balpha) = & \wnorm + C \sum_{(i,j,k) \in \cT} 
  f(\rho_{(i,j,k)}) \notag \\
  & + \sum_{(i,j,k)\in \cT }\mu_{(i,j,k)} 
  \biggr[ \rho_{(i,j,k)} - \w^{\T} \dhs (i,j,k) \biggr] - \balpha^\T \w, 
\end{align}
    where $\bmu$, $\balpha$ are Lagrange multipliers and $\balpha \geq \zero$.
    With the definition of Fenchel conjugate \citep{boyd}: 
$f^{\star}(z):=\sup_{x \in \text{dom} f} x^{\T}z - f(x)$ 
( here $f^\star(\cdot) $ is the Fenchel conjugate of the function $ f(\cdot) $ ), 
    we have the following
   dual objective:
 \begin{align}
  \inf_{\w, \brho}{L(\w, \brho, \bmu, \balpha)} 
    = & \inf_{\brho} \biggr( C \sum_{(i,j,k) \in \cT} f(\rho_{(i,j,k)}) 
    + \sum_{(i,j,k) \in \cT }\mu_{(i,j,k)} \rho_{(i,j,k)} \biggr) \notag \\
    = & - \sup_{\brho} \biggr( -C \sum_{(i,j,k) \in \cT} f(\rho_{(i,j,k)}) 
      - \sum_{(i,j,k) \in \cT }\mu_{(i,j,k)} \rho_{(i,j,k)} \biggr) \notag \\
    = & - C \sup_{\brho}  \biggr(
      \sum_{(i,j,k) \in \cT } \frac{-\mu_{(i,j,k)}}{C} \rho_{(i,j,k)} 
    - \sum_{(i,j,k) \in \cT} f(\rho_{(i,j,k)}) \biggr) \notag \\
    = & - C \sum_{(i,j,k) \in \cT} f^{\star} \left( \frac{-\mu_{(i,j,k)}}{C} \right).
 \end{align}
For the optimal primal solution, the condition: $\frac{\partial L}{\partial \w} = 0$ must hold; 
hence we have the following relation:
\begin{equation}
  \one - \balpha^\T -  
  \sum_{(i,j,k) \in \cT } \mu_{(i,j,k)} \dhs (i,j,k)  = \zero.
\end{equation}
Consequently, the corresponding dual problem
of \eqref{eq:cgh-general-l1} can be written as:
\begin{align}
  \label{eq:cgh-general-l1-dual} 
  \max_{\bmu} \;\; & - \sum_{(i,j,k) \in \cT} f^{\star} \left( \frac{-\mu_{(i, j, k)}}{C} \right) \\
  \st \;\; 
  & \forall \, h(\cdot) \in \wlset: \;\; \sum_{(i, j, k)\in \cT} \mu_{(i,j,k)} \dh(i,j,k) \leq 1. \label{eq:cgh-general-cons}
\end{align}
With the above dual problem for general smooth convex loss functions,
we generate a new hash function 
by finding the most violating constraints in \eqref{eq:cgh-general-cons},
which is the same as that for squared hinge loss.
Hence, we solve the optimization in \eqref{eq:cgh-wl} to generate a new hash function.
Using different loss functions will result in different dual solutions.
The dual solution is required for generating hash functions.

As aforementioned, 
in each column generation iteration, 
we need to obtain the dual solution before solving \eqref{eq:cgh-wl} to generate a hash function.
    Since we assume that $ f(\cdot) $ is smooth, the
    Karush-Kuhn-Tucker (KKT) conditions establish the connection
    between the primal solution of \eqref{eq:cgh-general-l1} and
    the dual solution of \eqref{eq:cgh-general-l1-dual}:
    \begin{equation}
        \label{eq:cgh-general-kkt}
        \forall (i,j,k) \in \cT: \mu_{(i,j,k)}^\star = - C f' \left( \rho_{(i,j,k)}^\star \right)
    \end{equation}
    in which,
    \begin{equation}
        \rho_{(i,j,k)}^\star = \w^{\star \T} \dhs (i,j,k).
    \end{equation}
    In other words, the dual variable is determined by the gradient of
    the loss function in the primal. 
    According to \eqref{eq:cgh-general-kkt}, 
    we are able to obtain the dual solution $ \bmu^\star $ using the primal solution $ \w^\star $.

\subsection{Discussion on extensions}

We can easily incorporate different kinds of loss functions and regularization in our learning framework.
In this section, we discuss the case of using the logistic loss and the $\ell_\infty$ norm regularization.

\subsubsection{Hashing with logistic loss}

It has been shown in \eqref{eq:cgh-general-shinge} that 
formulation for the squared hinge loss is an example 
of the general formulation in \eqref{eq:cgh-general-l1} with smooth convex loss functions.
Here we describe using the logistic loss as another example of the general formulation.
The learning algorithm is similar to 
the case of using the squared hinge loss which is described before.
We have the following definition for the logistic loss:
\begin{equation}
  \label{eq:cgh-general-logistic}
  f(\rho_{(i,j,k)})= \log ( 1 + \exp (-\rho_{(i,j,k)})).
\end{equation}
The general result for smooth convex loss function can be applied here. 
The primal optimization problem can be written as:
\begin{align}
\label{eq:cgh-logistic-l1-short} 
  \min_{\w, \brho} \;\; & \wnorm + C \sum_{(i,j,k) \in \cT} 
  \log ( 1 + \exp (-\rho_{(i,j,k)})) \\
  \st \;\; 
  & \forall (i,j,k) \in \cT: \rho_{(i,j,k)}  =  \w^{\T} \dhs (i,j,k), \notag \\
  & \w \geq \zero. \notag
\end{align}
The corresponding dual problem can be written as:
\begin{align}
\label{eq:cgh-logistic-l1-dual} 
  \max_{\bmu} \;\; & \sum_{(i,j,k) \in \cT} 
  ( \mu_{(i,j,k)} - C ) \log ( C-\mu_{(i,j,k)} ) - 
  \mu_{(i,j,k)} \log ( \mu_{(i,j,k)} )
  \\
  \st \;\; 
  & \forall \, h(\cdot) \in \wlset: \;\; \sum_{(i, j, k)\in \cT} \mu_{(i,j,k)} \dh(i,j,k) \leq 1. \notag
\end{align}
The dual solution can be calculated by:
\begin{equation}
        \label{eq:cgh-kkt-logistic}
    \forall (i,j,k) \in \cT: \mu_{(i,j,k)}^\star = 
      \frac{C}{\exp(\w^{\star \T} \dhs (i,j,k)) + 1}.
\end{equation}

\subsubsection{Hashing with $\ell_\infty$ norm regularization}
The proposed method is flexible that it is easy to incorporate different types of regularizations.
Here we discuss the $\ell_\infty$ norm regularization as an example.
For general convex loss, the optimization can be written as:
\begin{align}
\label{eq:cgh-general-linf} 
  \min_{\w, \brho} \;\; & \wnorminf + C \sum_{(i,j,k) \in \cT} f(\rho_{(i,j,k)}) \\
  \st \;\; 
  & \forall (i,j,k) \in \cT: \rho_{(i,j,k)}  =  \w^{\T} \dhs (i,j,k), \notag \\
  & \w \geq \zero. \notag 
\end{align}
This optimization problem can be equivalently written as:
\begin{align}
\label{eq:cgh-general-linf2} 
  \min_{\w, \brho} \;\; & \sum_{(i,j,k) \in \cT} f(\rho_{(i,j,k)}) \\
  \st \;\; 
  & \forall (i,j,k) \in \cT: \rho_{(i,j,k)}  =  \w^{\T} \dhs (i,j,k), \notag  \\
  & \zero \leq \w \leq C' \one, \notag 
\end{align}
where $C'$ is a constant that controls the regularization trade-off.
This optimization can be efficiently solved using quasi-Newton methods such as LBFGS-B
by eliminating the auxiliary variable $ \brho$.
The Lagrangian can be written as:
\begin{align}
  L(\w, \brho, \bmu, \balpha, \bbeta) = & \sum_{(i,j,k) \in \cT} 
  f(\rho_{(i,j,k)}) 
  - \balpha^\T \w + \bbeta^\T (\w - C' \one)
  \notag \\
  & + \sum_{(i,j,k)\in \cT }\mu_{(i,j,k)} 
  \biggr[ \rho_{(i,j,k)} - \w^{\T} \dhs (i,j,k) \biggr], 
\end{align}
where $\bmu$, $\balpha$, $\bbeta$ are Lagrange multipliers 
and $\balpha \geq \zero$, $\bbeta \geq \zero$.
Similar to the case for $ \ell_1 $ norm,
the dual problem can be written as:
\begin{align}
  \label{eq:cgh-general-linf-dual} 
  \max_{\bmu, \bbeta} \;\; & - C' \one^\T \bbeta - \sum_{(i,j,k) \in \cT} f^{\star}(-\mu_{(i, j, k)}) \\
  \st \;\; 
  & \forall \, h(\cdot) \in \wlset: \;\; \sum_{(i, j, k)\in \cT} \mu_{(i,j,k)} \dh(i,j,k) \leq \beta_h, \notag \\
  & \bbeta \geq \zero. \notag 
\end{align}

As the same with the case of $\ell_1$ norm,
the dual solution $\bmu$ can be calculated using \eqref{eq:cgh-general-kkt},
and the rule for generating one hash function is to solve the subproblem in \eqref{eq:cgh-wl}.

Similar to the discussion for $\ell_1$ norm, 
different loss functions, including the squared hinge loss in \eqref{eq:cgh-general-shinge} and 
the logistic loss in \eqref{eq:cgh-general-logistic}, 
can be applied here to incorporate the $\ell_\infty$ norm regularization.
As the flexibility of our framework, 
we also can use the non-smooth hinge loss with the $\ell_\infty$ norm regularization.

\comment{
\subsubsection{Extension of regularization}
     To demonstrate the flexibility of the proposed framework, we
     show an example that considers additional pairwise
     information for hashing learning.
    Assume that we are given the pairwise similarity information.
    It is expected that the distance of similar data pairs should be minimized. 
    We can easily add a new regularization term in our objective function to leverage 
    this additional information. 
    Formally, let us denote the set of pairwise relations by:
    \begin{equation}
      \cD = \{  (  i,  j  ) \; | \; \bx_i \text{ is similar to } \bx_j \}.
    \end{equation}
    We want to minimize the divergence:
    \begin{equation}
      \sum_{(i,j) \in \cD} \dhamm ( \bx_i, \bx_j  )
        = \sum_{r=1}^m  \sum_{(i,j) \in \cD}  w_r | h_r( \bx_i ) - h_r ( \bx_j )|.
    \end{equation}
    If we use this term to replace the $ \ell_1 $ regularization term:
    $ \wnorm $ of the optimization in \eqref{eq:cgh-general-l1},
    all of our analysis still holds and  Algorithm \ref{alg:cgh} is still applicable with small
    modification, because the new term can be simply seen as weighted $ \ell_1 $ norm.
}

%% file: method.tex
\section{Hashing for optimizing ranking loss}
\label{sec:sh}

  Our column generation based approach \cgh can be extended to optimize the more general ranking loss,
  which is more complex than the simple triplet loss.
  This extension is a structured learning based approach for binary code learning.
  Hence we referred to this extension as \sh in this paper.
  Before describing details of \sh, we first present
  a preliminary technique which applies large-margin based structured learning for optimize ranking loss.

	\subsection{Structured SVM for learning to rank}

	First we provide a brief overview of structured SVM.
    Let $\{( \bx_i; \by_i ) \}$, $ i = 1, 2 \cdots$, denote a set of input-output pairs.
    The discriminative function for structured output prediction is
    $F(\bx,\by):
    \cX\times\cY\mapsto \Real$, which measures the compatibility of the input and output pair
    $(\bx,\by)$.
     Structured SVM enforces that the score of the
     ``correct'' model $ \by'$ should be larger than all other ``incorrect''
     model $ \by $,  $ \forall \by \neq \by' $, which writes:
     \begin{equation}
         \label{EQ:svm1}
         \forall \by \in \cY:\quad  %
            \w^\T [ \wstructs (  \bx, \by'  )  -
                    \wstructs (  \bx, \by  ) ]
                    \geq \loss( \by, \by'  ) - \xi.
     \end{equation}
     Here $  \xi $ is a slack variable (soft margin) corresponding to the hinge
     loss.
     $ \wstructs( \bx, \by   )  $ is a vector-valued joint feature mapping.
     It plays a key role in structured learning and specifies the
     relationship between an input $ \bx $ and output $ \by $. $\w$ is the model parameter. The label loss $\loss(\by,\by') \in \Real$ measures the discrepancy of the predicted $\by$ and the true
    label $\by'$.
    A typical assumption is that $\loss(\by, \by)=0, \loss(\by, \by')>0$ for any $\by \neq \by'$,
    and $\loss(\by, \by')$ is upper bounded.
     The prediction $\by^\star$ of an input $\x$ is achieved by
      \begin{align}
      \label{eq:infer1}
       \by^\star=\argmax_{\by \in \cY }F(\bx, \by) = \w ^\T \wstructs( \bx, \by ).
     \end{align}

    For structured problems, the size of the output $|\cY|$ is typically very large or infinite.
    Considering all possible constraints in \eqref{EQ:svm1} is generally intractable.
    The cutting-plane method \citep{kelley1960} is commonly employed, which
    allows to maintain a small working-set of constraints and obtain an approximate solution of the original problem 
    up to a pre-set precision.  
    To speed up, the 1-slack reformulation is proposed
    \citep{JoachimsSVM}.
    Nonetheless the cutting-plane method needs to find the most violated
    label (equivalent to an inference problem) by solving the following optimization:
      \begin{align}
        \label{EQ:SSVM2}
        \by^\star=\argmax_{ \by \in \cY   }  \; \w^\T \wstructs ( \bx, \by  )
        + \loss( \by, \by'  ).
    \end{align}
    Structured SVM typically requires: 1) a well-designed feature representation $ \wstructs(\cdot, \cdot)
    $; 2) an appropriate label loss $ \loss( \cdot, \cdot  ) $;
    3) solving inference problems \eqref{eq:infer1} and \eqref{EQ:SSVM2} efficiently.
    In a retrieval system, given a test data point $\x$, the goal is to predict a ranking of data points in the database. For a
    ``correct'' ranking, relevant data points are expected to be
    placed in front of irrelevant data points.
    A ranking output is
    denoted by $\y$.
    Given a query $ \bx_i $, we use $ \cX_i^+  $ and $ \cX_i^-  $
    to denote the subsets of relevant and irrelevant data points in
    the training data.
    Given two data points: $\x_i$ and $\x_j$,
    $\x_i \!\!   \prec_{\y} \!\!  \x_j $
     ($\x_i \!\! \succ_{\y} \!\! \x_j $) means that $\x_i $ is placed before
     (after) $\x_j $ in the ranking $\y$.
    Let us introduce a symbol  $ y_{jk} = 1  $ if
    $\x_j \!\!   \prec_{\y} \!\!  \x_k $ and $ y_{jk} = -1 $
    if  $\x_j \!\!   \succ_{\y} \!\!  \x_k $.
    The ranking can be evaluated by various measures such as AUC, NDCG,
    mAP. These evaluation measures can be
    optimized directly as label loss
    $\loss$ \citep{Joachims2005,mcfee10_mlr}.
    Here $\wstructs(\x_i, \y)$ can be defined as:
  \begin{align}
	\wstructs(\bx_i, \by)
	= %
	\sum_{\bx_j \in \cX_i^+} \sum_{\bx_k \in \cX_i^-}
    y_{jk}
    \Bigl[
        \frac{\phi( \bx_i, \bx_j  )  -  \phi( \bx_i, \bx_k )
        }
        {
             |\cX_i^+| \cdot | \cX_i^- |
        }
    \Bigr].
    \label{EQ:rk1}
  \end{align}
$\cX_i^+$ and $\cX_i^-$ are the sets of relevant and irrelevant
neighbours of data point $\x_i$ respectively.
Here $  | \cdot | $ is the set size.
The feature map $ \phi( \bx_i, \bx_j  )   $ captures the relation between a query $ \bx_i $ and point $ \bx_j $.

        We have briefly reviewed how to
        optimize ranking criteria using structured prediction. Now we review some
        basic concepts of hashing before introducing our framework.

    For the time being, let us assume that we have already learned all
    the hashing functions. In other words, {\em given a data point $ \bx $,
    we assume that we have access to its corresponding hashed values $ \hx  $,
    as defined in \eqref{EQ:Hash}.
    }
    Later we will show how this mapping can be explicitly learned
    using column generation. Now let us focus on how to optimize for
    the weight $ \w $.
    When the weighted hamming distance is used, we aim to learn an
    optimal weight $ \w $.
    Distances are calculated in the learned space and ranked
    accordingly.
    A natural choice for the vector-valued mapping function $ \phi $ in \eqref{EQ:rk1} is
    \begin{equation}
        \label{EQ:phi}
         \phi( \bx_i, \bx_j   )  =  -  |  \hx_i - \hx_j   |.
    \end{equation}
    Note that we have flipped the sign, which preserves the ordering
    in the standard structured SVM.
    Due to this change of sign, sorting the data by ascending $
    d_{\sf hm} ( \x_i, \x_j   )    $ is equivalent to sorting
    by descending $ \w^\T  \phi ( \bx_i, \bx_j   ) = -
    \w^\T | \hx_i - \hx_j | $.

    The loss function $ \loss( \cdot, \cdot ) $ depends on the
    metric, which we will discuss in detail in the next section.
    For ease of exposition, let us define
    \begin{equation}
        \label{EQ:delta}
        \dwstructs_i ( \by )
              = \wstructs ( \x_i, \by_i ) - \wstructs ( \x_i, \by ),
    \end{equation}
    with $\wstructs ( \x_i, \by )$ defined in \eqref{EQ:rk1}.
    We consider the following problem,
    \begin{align}
    \label{eq:sh}
          \min_{ {\w \geq 0, \bxi \geq 0}}   \;\; &
    \|\w \|_1 +  {\tfrac{C}{m}} \, \sum_{i=1}^m  \xi_{i} \\
    \st  \;\;
	&
    \forall i=1, \dots, m
    \text{ and } \forall \by \in \cY: \notag \\
    &
    \w^\T \dwstructs_i ( \by )  \geq
         \loss(\by_i,\by) - \xi_i. \label{eq:sh_con}
	\end{align}
    Unlike standard structured SVM, here we use the
    $ \ell_1 $ regularisation (instead of $ \ell_2 $) and enforce that $ \w $ is non-negative.
    This is aligned with boosting methods
    \citep{demiriz2002linear,Shen2011Totally}, and enables us to learn hash functions efficiently.

\subsection{Weighting learning via cutting-plane} %
\label{SUBSEC:OptW}
Here we show how to learn the weighting coefficient $\w$.
Inspired by \citep{JoachimsSVM},
we first derive the \oneslack formulation of the original \nslack  formulation \eqref{eq:sh}:

\begin{align}
\label{eq:sh-1s}
  & \min_{\w \geq 0, \xi \geq 0}  \;\;
  \| \w \|_1 + C \xi \\
  \st  \;\;
  &
  \forall \c \in\{0,1\}^m \text{ and }
  \forall \y \in \cY, i=1,\cdots, m: \notag \\
  &
   \frac{1}{m} \w^ \T
  \biggl[
             \sum_{i=1}^m c_i
            \cdot
                    \dwstructs_i ( \y )
    \biggr]
            \geq
            \frac{1}{ m }  \sum_{i=1}^m c_i\loss(\y_i,\y ) - \xi.
  \label{eq:sh-1s_con}
\end{align}
Here $ \c $ enumerates all possible $ \c \in \{ 0, 1 \}^n$.
As in \citep{JoachimsSVM}, cutting-plane methods can
    be used to solve the \oneslack primal problem
    \eqref{eq:sh-1s} efficiently.
    Specifically, we need to solve a maximization for every $\x_i$
    in each cutting-plane iteration to find the most violated
    constraint of \eqref{eq:sh-1s_con},
given a solution $\w$:
\begin{align}
\y_i^\star= \argmax_{\y} \loss( \y_i, \y ) - \w ^ \T  \dwstructs_i (\y ).
\label{eq:infer}
\end{align}
The cutting-plane algorithm is summarized in Algorithm \ref{ALG:sh-cp}.

\begin{algorithm}[ht]
\caption{Cutting planes for solving the \oneslack primal}
\label{ALG:sh-cp}
\KwIn{cutting-plane tolerance: $\epsilon_{\rm cp}$;
  inputs from Algorithm \ref{ALG:sh-cg}.
}
\KwOut{$\w$ and $\bmu$.}
{\bf Initialize:} working set: $\cW\leftarrow \emptyset
$;  $c_i = 1$, $\y_i' \leftarrow $ any element in $\cY $, for $
i=1,\dots, n$.

    \Repeat{
$
      \frac{1}{n} \w^ \T
  \biggl[
             \sum\limits_{i=1}^n c_i
                    \dwstructs_i ( \y_i' )
    \biggr]
            \geq
            \frac{1}{n} \sum\limits_{i=1}^n c_i\loss(\y_i,\y_i' ) - \xi -
            \epsilon_{\rm cp}
$
     }{

    $\cW \leftarrow \cW \cup \{ (c_1,
    \dots, c_n, \y_1',\dots,\y_n')  \} $\;

    obtain primal and dual solutions $\w,\xi$; $\blambda $
    by solving \eqref{eq:sh-1s} (e.g., using MOSEK \citep{mosek}) on current working set $\cW$ \;

    \For{$i=1, \dots, n$}{

      $ \y_i'= \argmax_{\y} \loss( \y_i, \y ) - \w ^ \T  \dwstructs_i ( \y ) $\;

      $ c_i = \left\{
        \begin{array}{l l}
          1 & \quad \loss( \y_i, \y_i' ) - \w ^ \T  \dwstructs_i ( \y_i' ) > 0\\
          0 & \quad {\rm otherwise} \\
        \end{array}
      \right. $ \;

      }%

  } %

    update $ \mu_{(i,\y)} = \sum_{\c}\lambda_{ ( \c, \y ) } c_i$
    for $\forall ( \c, \y ) \in \cW$ \;

\end{algorithm}

We now know how to efficiently learn $\w$
using cutting-plane methods. However, it remains unclear how to learn
hash functions (or features). Thus far, we have taken for granted that
the hashed values $ \hx $ are given.
    We would like to learn the hash functions and $ \w $
    in a single optimization framework.  Next we show how this is
    possible using the column generation technique from boosting.

\subsection{Learning hash functions using column generation} %
\label{SUBSEC:CG}

 Note that the dimension of $ \w $ is the same as
 the dimension of $ \hx $ (and of $  \phi ( \cdot, \cdot )$, see Equ.\
 \eqref{EQ:phi}), which is the number of hash bits by the definition
 \eqref{EQ:Hash}. If we were able to
 access all hash functions, it may be possible to select a subset of
 them and learn the corresponding $ \w $ due to the sparsity introduced by the $ \ell_1 $
 regularization in \eqref{eq:sh}.
    Unfortunately,
    the number of possible hash functions can be infinitely
    large. In this case it is in general infeasible to solve the optimization
    problem exactly.
    We here develop a column generation algorithm for \sh to iteratively learn the hash functions and weights,
    which is similar to \cgh.

 To learn hash functions via column generation, we derive the dual problem of
 the above \oneslack optimization, which is,
  \begin{align}
  \label{eq:sh-1s-dual}
    \max_{ \blambda \geq 0 }
    \;\;
    &
    \sum_{ \c , \y}
    \lambda_{ (\c, \y ) }
    \sum_{i=1}^m c_i  \loss( \y_i, \y ) \\
   \st \;\;
   &  \frac{1}{m} \sum_{ \c, \y } \lambda_{ (\c, \y) }
     \biggl[
          \sum_{i=1}^m c_i \cdot  \dwstructs_i( \y )
     \biggr]
     \leq \one, \label{EQ:11b}
\\
      &  0 \leq \sum_{ \c, \y} \lambda_{ (\c, \y ) } \leq C. \notag
\end{align}
We denote by $ \lambda_{( \c, \y)} $ the \oneslack dual variable associated with one constraint in \eqref{eq:sh-1s_con}.
Note that \eqref{EQ:11b} is a set of constraints
    because $ \dwstructs( \cdot )$ is a vector of the same dimension
    as $ \phi(\cdot, \cdot)  $ as well as $  \hx  $, which can be infinitely large.
One dimension in the vector $ \dwstructs( \cdot )$ corresponds to one constraint in \eqref{EQ:11b}.
Finding the most violated constraint in the dual form
\eqref{eq:sh-1s-dual} of the \oneslack formulation for
    generating one hash function is to maximize the l.h.s.\ of
    \eqref{EQ:11b}.

    The calculation of $ \dwstructs( \cdot )$ in \eqref{EQ:delta} can be simplified as follows.
    Because of the subtraction of $\wstructs( \cdot )$ (defined in \eqref{EQ:rk1}),
    only those incorrect ranking pairs will appear in the calculation.
    Recall that the true ranking is $ \by_i $ for $ \bx_i $.
    We define $ \cS_i(\y ) $ as a set of incorrectly ranked pairs:
    $ (j, k) \in \cS_i(\y )   $, in which the incorrectly ranked pair $(j, k)$ means
    that the true ranking is $  \bx_j \!\! \prec_{\y_i} \!\!  \bx_k  $
    but $ \bx_j \! \succ_{\y}  \! \bx_k  $. So we have
    \begin{align}
            \dwstructs_i ( \by  )
               & =
               \tfrac{2}{ | \cX^+_i | |\cX^-_i | }
               \sum_ {
               (j,k) \in \cS_i(\y)
               }
\bigl[
            \phi( \bx_i, \bx_j ) - \phi( \bx_i, \bx_k )
\bigr] \notag
    \\
    &           =
               \tfrac{2}{ | \cX^+_i | |\cX^-_i | }
               \sum_ {
               (j,k) \in \cS_i(\y)
               }
\bigl(
| \hx_i - \hx_k | - |  \hx_i - \hx_j |
\bigr).
\end{align}
With the above equations and the definition of $\hx$ in \eqref{EQ:Hash},
the most violated constraint in  \eqref{EQ:11b}
can be found by solving the following problem:
\begin{align}
    \label{EQ:CG-Sub-tmp}
    h^\star ( \cdot )  =  \argmax_ { h(\cdot) } &
    \sum_{\c, \by } \lambda_{ (\c, \y ) }
    \sum_i
      \frac{2 c_i}{ | \cX^+_i | |\cX^-_i | }
      \cdot
      \notag    \\
     & \sum_ { (j,k) \in \cS_i(\y) }
\bigl(
| h( \x_i ) - h(\x_k) | - |  h(\x_i) - h(\x_j) |
\bigr).
\end{align}
By exchanging the order of summations,
the above optimization can be further written in a compact form:
\begin{align}
    \label{EQ:CG-Sub}
    h^\star ( \cdot ) &  =  \argmax_ { h(\cdot) } \sum_{i, \y}
     \sum_ { (j,k) \in \cS_i(\y)  } \mu_{(i, \y)} \,
\bigl( | h( \x_i ) - h(\x_k) | - |  h(\x_i) - h(\x_j) |\bigr), \\
 & \text{where, }
	\mu_{(i, \y)} = \frac{2}{ | \cX^+_i | |\cX^-_i | } \sum_{\c} \lambda_{ (\c, \y ) } c_i.
\end{align}
The objective in the above optimization is a summation of weighted triplet $(i,j,k)$ ranking scores,
in which $\mu_{(i, \y)}$ is the triplet weighting value.
Solving the above optimization provides the best hash function
            for the current solution $\w$.
            Once a hash function is generated, we learn $ \w $ using cutting-plane in Sec. \ref{SUBSEC:OptW}.
The column generation procedure for hash function learning is
summarised in Algorithm \ref{ALG:sh-cg}.

\begin{algorithm}[ht]
\caption{\shh: Column generation for hash function learning}
\label{ALG:sh-cg}
  \KwIn{
  training examples: $ (\x_1; \y_1), (\x_2; \y_2) ,\cdots
    $; trade-off parameter: $C$;
    the maximum iteration number (bit length $m$).
  }
  \KwOut{
  learned hash functions $[h_1, \dots, h_m]$ and weighting coefficients $ \w $.
  }
    {\bf Initialize:}
    working set of hashing functions $\cW_{\mathrm{H}}\leftarrow \emptyset$;
    for each $i$, ($i=1,\dots,n $),
    randomly pick any  $\y_i^{(0)} \in \cY$,
    initialize $ \mu_{(i,\y)} = \frac{C}{n} $ for $ \y = \y_i^{(0)}$, and
    $ \mu_{(i,\y)} = 0 $ for all $ \y \in \cY
    \backslash \y_i^{(0)}$.

    \Repeat{the maximum number of iterations is reached
    }
    {
      Find a new hashing function $h^\star ( \cdot )$ by solving Equ.\ \eqref{EQ:CG-Sub};
      add $h^\star$ into the working set of hashing functions: $\cW_{\mathrm{H}}$\;
      Solve the structured SVM problem \eqref{eq:sh} or the equivalent
    \eqref{eq:sh-1s} using cutting-plane in Algorithm \ref{ALG:sh-cp}, to obtain $\w$ and $\bmu$\;
    }
\end{algorithm}

In most of our experiments, we use the linear perceptron hash function with the output in $\{0, 1\}$:
\begin{align}
\label{eq:hash_learn}
h(\x)=0.5( {\rm sign}(\bv^\T\x+ b)+1).
\end{align}
We apply a similar way as \cgh for learning the hash function.
Please refer to the learning procedure of \cgh in Sec. \ref{sec:cg_learning} for details.
Basically, we replace the $ \rm sign(\cdot) $ function by a smooth
sigmoid function, and then locally solve the above optimization \eqref{EQ:CG-Sub} (e.g.,
LBFGS \citep{lbfgs}) for learning the parameters of a hash function.
We apply the spectral relaxation \citep{liu2011hashingGraphs} to obtain an initial point for solving \eqref{EQ:CG-Sub},
which drops the $ \rm sign(\cdot) $ function and solves a generalized eigenvalue problem.

Next, we discuss some widely-used information retrieval evaluation
criteria, and show how they can be seamlessly incorporated into \shh.

\subsection{Ranking measures} %
\label{SEC:RM}

Here we discuss
a few ranking measures for loss functions, including AUC and NDCG. %
Following \citep{mcfee10_mlr}, we define the loss function over two rankings $\loss \in [0 \; 1]$ as:
\begin{align}
\label{eq:loss_score}
\loss(\y,\y')=1-\text{score}(\y,\y').
\end{align}
Here $\y'$ is the ground truth ranking and $\y$ is the prediction. We define $\cX_{\y'}^+$ and $\cX_{\y'}^-$ as the indexes of relevant and irrelevant neighbours respectively in the ground truth ranking $\y'$.

{\bf AUC}. The area under the ROC curve
is to evaluate the performance of correct ordering of data pairs,
which can be computed by counting the proportion of correctly ordered data pairs:
\begin{align}
\mathrm{score}_{\mathrm{AUC}}(\y,\y')=\frac{1}{|\cX_{\y'}^+||\cX_{\y'}^-|}\sum_{i\in\cX_{\y'}^+}\sum_{j\in\cX_{\y'}^-}\delta(i \prec_\y j).
\end{align}
$\delta(\cdot) \in \{0, 1\}$ is the indicator function.
For using this AUC loss,
the maximization inference in \eqref{eq:infer} can be solved efficiently by sorting the distances of data pairs, as described in \citep{Joachims2005}.
Note that the loss of a wrongly ordered pair is not related to their positions in the ranking list, thus AUC is a position insensitive measure.
It clearly shows that AUC loss is to calculate the portion of correctly ranked triplets.
Hence optimizing AUC loss in \sh is equivalent to optimize the triplet loss in \cgh.

\comment{
{\bf Precision-at-K}.
Precision-at-K is to evaluate the quality of top-K retrieved examples in a ranking. It is computed by counting the number of relevant data points within top-K positions and divided by K:
\begin{align}
\text{score}_{\mathrm{P@K}}(\y,\y')=\frac{1}{K}\sum_{i=1}^{K}\delta(\y(i) \in \cX_{\y'}^+).
\end{align}
Here $\y(i)$ is the example index on the $i$-th position of a ranking $\y$;
$\delta(\cdot)$ is an indicator.
An algorithm for solving the inference in
\eqref{eq:infer} is proposed in \citep{Joachims2005}.
}

{\bf NDCG}. Normalized Discounted Cumulative Gain \citep{JarvelinK00} is to measure the ranking quality of the first K returned neighbours. A similar measure is Precision-at-K which is the proportion of top-K relevant neighbours. NDCG is a position-sensitive measure which considers the positions of the top-K relevant neighbours.
Compared to the position-insensitive measure: AUC, NDCG assigns different importances on the ranking positions,
which is a more favorable measure for a general notion  of a ``good'' ranking in real-world applications.
In NDCG, each position of the ranking is assigned a score in a decreasing way. NDCG can be computed by accumulating the scores of top-K relevant neighbours:
\begin{align}
\mathrm{score}_{\mathrm{NDCG}}(\y, \y')=\frac{1}{\sum_{i=1}^{K}{S(i)}}\sum_{i=1}^{K}{S(i)}\delta(\y(i) \in
\cX_{\y'}^+).
\end{align}
Here $\y(i)$ is the example index on the $i$-th position of the ranking $\y$.
$S(i)$ is the score assigned to the $i$-th position in the ranking.
$S(1)=1$, $S(i)=0$ for $i>K$ and $S(i)=1/\log_2(i)$ for other cases.
A dynamic programming algorithm is proposed in \citep{Chakrabarti2008} for solving the maximization inference in \eqref{eq:infer}.

\comment{
{\bf mAP}. Mean average precision (mAP) is the averaged precision-at-K scores over all positions of relevant data points in a ranking, which is computed as:
\begin{align}
\mathrm{score}_{\mathrm{mAP}}(\y, \y')=\frac{1}{|\cX_{\y'}^+|}\sum_{i=1}^{|\cX_{\y'}^+|+ |\cX_{\y'}^-|}\delta(\y(i)\in \cX_{\y'}^+) \mathrm{score}_{\mathrm{P@K}(K=i)}(\y, \y').
\end{align}
For using this mAP loss, an efficient algorithm for solving the inference in
\eqref{eq:infer} is proposed in \citep{Yue2007}.
}

\subsection{Speedup training}
\label{sec:speedup}
In this section, we propose two strategies to speedup the training procedure of our \sh model, both from the aspects of training and inference.

\subsubsection{Stage-wise training}

When learning a new hash function,
the original \sh model needs to solve for all the weights of all hash functions in each column generation iteration.
As the number of hashing functions increase, the dimension of the weights which need to learn is also increase,
When the dimension of weights increases, 
we usually need to perform a large number of inference operations (see \eqref{eq:infer}) in the cutting-plane algorithm for the convergence, 
which is generally computation expensive.
The learning procedure becomes more and more expensive as the number of bits increases.

Here we exploit the stage-wise learning strategy to speedup the training.
In column generation based totally-corrective boosting methods \citep{lpboost,shen10,Shen2014SBoosting}, 
all the hash function weights $\w$ are updated during each column generation iteration. 
In contrast, in stage-wise boosting, \eg, AdaBoost, 
only the weight of the newly added weak learner is updated in the current boosting iteration and weights of all previous weak learners are fixed.  
This leads to more efficient training and is less prone to overfit.
Inspired by the stage-wise boosting,
we here exploit a new training protocol based on the stage-wise training to speedup the training of \sh.
Specifically, in the $t$-th column generation iteration, we only learn two weight variables, \ie, $w_t$ and $w_{t-1}$, where $w_t$ is the weight of the current newly added hash function, and $w_{t-1}$ is the weight shared by all previous hash functions.

Using this stage-wise training, we only need to solve for two variables ($w_t$ and $w_{t-1}$) for learning one hashing function 
 using the cutting-plane algorithm (Algo .\ref{ALG:sh-cp}).
With much less variables, the cutting-plane algorithm is able to converge much faster, therefore significantly reduce the number of inference (solving \eqref{eq:infer}) need to perform. 
Since we solve the optimization problem with only two variables for every hash bit,
the optimization complexity of learning new hash function is not increasing with the number of bits.
Thus this stage-wise training could easily scale to learning for large number of bits.
In the experiment part, as shown in Table \ref{tab:cmp_time}, 
this new training protocol largely reduce the total number of inference iterations for learning one hash function and this is not increased with the number of bits, 
hence brings orders of magnitude training speedup.

One more advantage of using this stage-wise training is that by forcing all the hash functions to share the same weights, we can use unweighted hamming distance to calculate similarities of the learned binary codes. 
We observe that the unweighted hamming distance is more efficient and has better generalization performance, as demonstrated later in the experiment section (Sec. \ref{sec:eval_new}), This also indicates that the hash functions are more important to the performance than the weights of hash functions in the \sh model.

\subsubsection{Optimizing Simplified NDCG (SNDCG) score}
As discussed before, we need to solve the maximization inference in \eqref{eq:infer} for finding most violated constraints. The computational complexity for solving this inference problem mainly depends on the definition of 
$\loss(\y,\y')$ in \eqref{eq:loss_score}, of which some examples are discussed in Sec.~\ref{SEC:RM}. Usually when using position sensitive loss functions, such as mAP, NDCG, it is computational expensive to solve the maximization inference \citep{Yue2007,Chakrabarti2008}, which might  limit its application on large-scale learning. 
Inspired by the efficient metric learning method in \citep{lim2014efficient}, here we discuss a form of position-sensitive ranking loss which is capable for fast inference.
Basically, we construct a simplified NDCG (referred to as SNDCG) score, based on a number of NDCG scores which are calculated from ``simple" rankings.
\begin{align}
\label{eq:sndcg}
\mathrm{score}_\mathrm{SNDCG}(\y, \y') = \frac{1}{|\cX_{\y'}^+|}\sum_{i\in\cX_{\y'}^+} N(i, \y)
\end{align}
where,
\begin{align}
N(i, \y)=\sum_{j=1}^{|\cX_{\y}^-|+1}{S(j)}\delta(\y(j) = i).
\end{align}
Here $\y(j)$ is the example index on the $j$-th position of the ranking $\y$.
$S(j)$ is the score assigned to the $j$-th position in the ranking.
$S(j)=1/\log_2(1+j)$.

It clearly shows that the loss is decomposed over all relevant examples.
$N(i)$ represents the NDCG score corresponding to the $i$-th relevant example, which is calculated from a simple ranking: a ranking only involves one relevant example and all irrelevant examples. 
The summation over relevant examples in \eqref{eq:sndcg} allow independent inference calculation for each relevant example.
For solving the inference on the simple ranking for each relevant example, we only need to perform a simple sorting of the hamming distances which is very efficient.
Hence the maximization inference in \eqref{eq:infer} can be independently and efficiently solved for each relevant example, which is much more efficient than using the original NDCG loss. In the experiment section, we evaluate training efficiency and ranking accuracy of the proposed SNDCG loss in Sec.~\ref{sec:eval_new}.

%% file: exp.tex
\section{Experiments}
\label{sec:exp}

We evaluate our column generation learning framework for binary code learning in this section.
Specifically, we evaluate the proposed method \cgh for optimizing triplet loss 
and the more general method \sh for optimizing ranking loss.
We first compare our models with state-of-the-art methods in Sec. \ref{sec:eval_state}, 
and then in Sec. \ref{sec:eval_new} we evaluate the more efficient models proposed in Sec. \ref{sec:speedup}.

Nine datasets are used here for evaluation, including  one UCI dataset: ISOLET, 4 image datasets: 
CIFAR10\footnote{http://www.cs.toronto.edu/\~{}kriz/cifar.html},
STL10\footnote{http://www.stanford.edu/\~{}acoates/stl10/},
MNIST, USPS, and another 4 large image datasets:
Tiny-580K \citep{gong2012iterative}, Flickr-1M\footnote{http://press.liacs.nl/mirflickr/}, 
SIFT-1M \citep{wang2010semi} and GIST-1M\footnote{http://corpus-texmex.irisa.fr/}.
CIFAR10 is a subset of the 80-million tiny images and STL10 is a subset of Image-Net.
Tiny-580K consists of $580,000$ tiny images.
Flick-1M dataset consists of 1 million thumbnail images.
SIFT-1M and GIST-1M datasets contain 1 million SIFT and GIST features respectively.

For the hashing performance evaluation, we follow the common
setting in many supervised methods \citep{kulis2009learning,KSH}.
For multi-class datasets, we use class labels to
define the relevant and irrelevant semantic neighbours by label
agreement.
For large datasets: Flickr-1M, SIFT-1M, GIST-1M and Tiny-580K, the semantic ground truth is defined according to the
$\ell_2$ distance \citep{wang2010semi}.
Specifically, a data point is labeled as a
relevant data point of the query if it lies in the top 2 percentile
points in the whole dataset.
We generated GIST features for all image datasets except MNIST and USPS.
we randomly select 2000 examples for testing
queries, and the rest is used as database.
We sample 2000 examples from the database as training data for learning models.
For large datasets, we use 5000 examples for training.
To evaluate the performance of compact bits, the maximum bit length is set to 64,
as similar to the evaluation settings
in other supervised hashing methods \citep{kulis2009learning}.

\begin{table}[t]
\caption{Results using NDCG measure (64 bits).
 We compare our \sh using AUC (StructH-A) and NDCG (StructH-N) loss functions, and our \cgh for triplet loss with 
 other supervised and unsupervised methods.
 \sh using NDCG loss performs the best in most cases.}
\centering
\resizebox{.9\linewidth}{!}
  {
  \begin{tabular}{ c || c c c | c c c | c c c c c}
\hline
Dataset & StructH-N & StructH-A &CGH  &SPLH &STHs &BREs &ITQ  &SPHER  &MDSH &AGH  &LSH \\
\hline
& \multicolumn{11}{c}{NDCG ($K=100$)} \\ \hline
STL10 &\bf 0.435  &0.374  &0.375  &0.404  &0.214  &0.289  &0.337  &0.318  &0.313  &0.310  &0.228\\
USPS  &\bf  0.905 &0.893  &0.900  &0.816  &0.688  &0.777  &0.804  &0.762  &0.735  &0.741  &0.668\\
MNIST &0.851  &0.798  &\bf 0.867  &0.804  &0.594  &0.805  &0.856  &0.806  &0.100  &0.793  &0.561\\
CIFAR &0.335  &0.259  &0.258  &\bf 0.357  &0.178  &0.273  &0.314  &0.297  &0.283  &0.286  &0.168\\
ISOLET  &\bf 0.881  &0.839  &0.866  &0.629  &0.766  &0.483  &0.623  &0.518  &0.538  &0.536  &0.404\\
\hline
  \end{tabular}
  }
\label{tab:main}
\end{table}

\begin{table}[t]
\caption{Results using ranking measures of Precision-at-K, Mean Average Precision and Precision-Recall (64 bits).
 We compare our \sh using AUC (StructH-A) and NDCG (StructH-N) loss functions, and our \cgh for triplet loss with other supervised and unsupervised methods. Our method using NDCG loss performs the best on these measures}
\centering
\resizebox{.9\linewidth}{!}
  {
  \begin{tabular}{ c || c c c | c c c | c c c c c}
\hline
Dataset & StructH-N & StructH-A &CGH  &SPLH &STHs &BREs &ITQ  &SPHER  &MDSH &AGH  &LSH \\
\hline
& \multicolumn{11}{c}{Precision-at-K ($K=100$)} \\ \hline
STL10 &\bf 0.431  &0.376  &0.376  &0.396  &0.208  &0.279  &0.325  &0.303  &0.298  &0.301  &0.222\\
USPS  &\bf 0.903  &0.894  &0.898  &0.805  &0.667  &0.755  &0.780  &0.730  &0.698  &0.711  &0.637\\
MNIST &0.849  &0.807  &\bf 0.862  &0.797  &0.579  &0.790  &0.842  &0.788  &0.100  &0.780  &0.540\\
CIFAR &0.336  &0.259  &0.261  &\bf 0.354  &0.174  &0.264  &0.301  &0.286  &0.270  &0.281  &0.164\\
ISOLET  &\bf 0.875  &0.844  &0.859  &0.604  &0.755  &0.448  &0.589  &0.477  &0.493  &0.493  &0.370\\
\hline
& \multicolumn{11}{c}{Mean Average Precision (mAP)} \\ \hline
STL10 &\bf 0.331  &0.326  &0.322  &0.299  &0.155  &0.211  &0.233  &0.193  &0.178  &0.162  &0.162\\
USPS  &\bf 0.868  &0.851  &0.848  &0.689  &0.456  &0.582  &0.566  &0.451  &0.405  &0.333  &0.418\\
MNIST &\bf 0.802  &0.790  &0.789  &0.684  &0.397  &0.558  &0.585  &0.510  &0.119  &0.505  &0.343\\
CIFAR &0.294  &\bf 0.300  &0.298  &0.289  &0.147  &0.204  &0.215  &0.204  &0.181  &0.201  &0.149\\
ISOLET  &\bf 0.836  &0.796  &0.815  &0.518  &0.653  &0.340  &0.484  &0.357  &0.348  &0.298  &0.267\\
\hline
& \multicolumn{11}{c}{Precision-Recall} \\ \hline
STL10 &\bf 0.267  &0.248  &0.248  &0.246  &0.130  &0.181  &0.200  &0.174  &0.164  &0.145  &0.138\\
USPS  &\bf 0.776  &0.760  &0.760  &0.609  &0.401  &0.520  &0.508  &0.424  &0.379  &0.326  &0.375\\
MNIST &\bf 0.591  &0.574  &0.582  &0.445  &0.165  &0.313  &0.323  &0.246  &0.018  &0.197  &0.143\\
CIFAR &0.105  &0.093  &0.091  &\bf 0.110  &0.042  &0.066  &0.074  &0.069  &0.064  &0.061  &0.042\\
ISOLET  &\bf 0.759  &0.709  &0.737  &0.445  &0.563  &0.301  &0.429  &0.321  &0.320  &0.275  &0.238\\
\hline
  \end{tabular}
  }
\label{tab:main-other}
\end{table}

\begin{figure}[t]
    \centering
   \includegraphics[width=.3\linewidth]{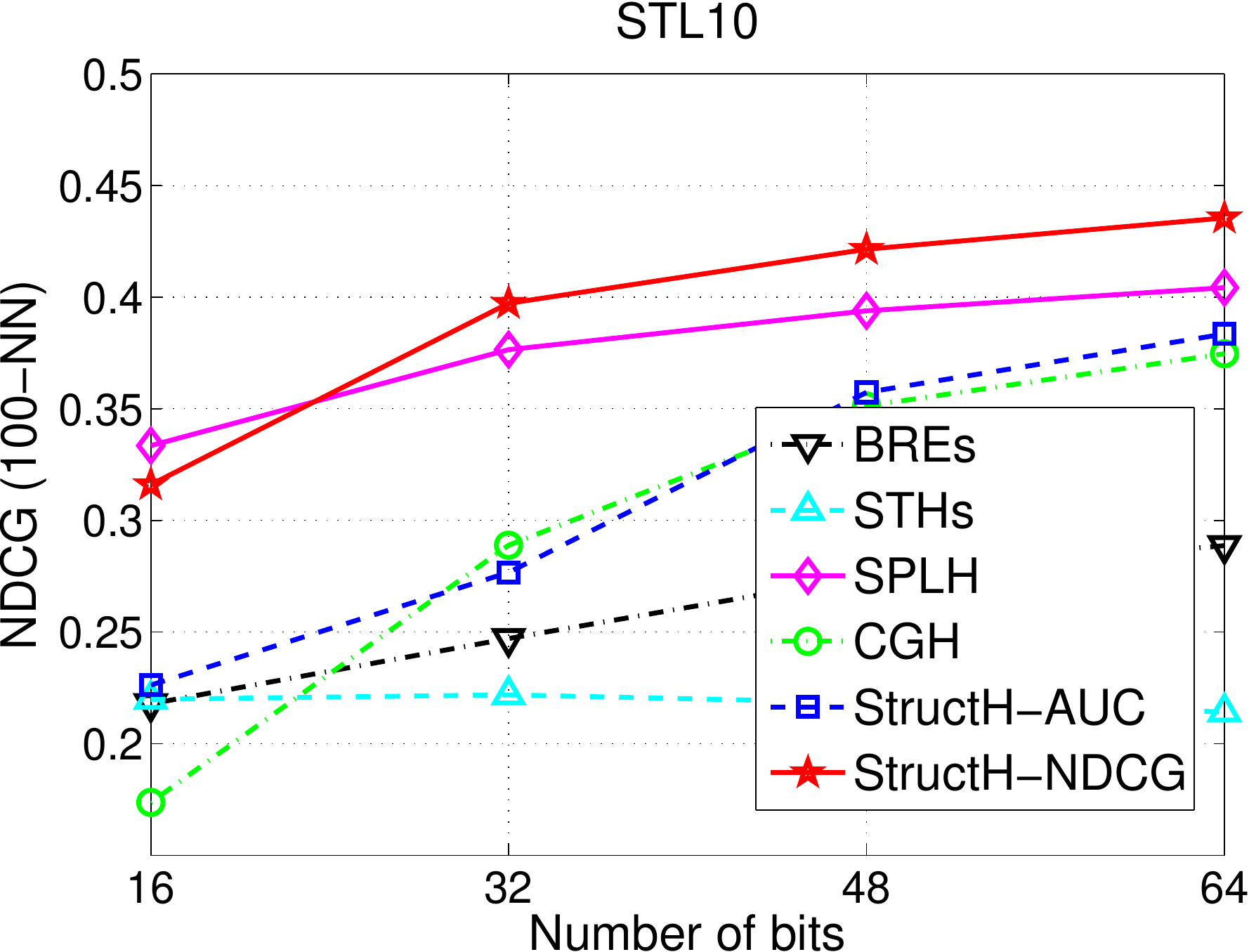}
   \includegraphics[width=.3\linewidth]{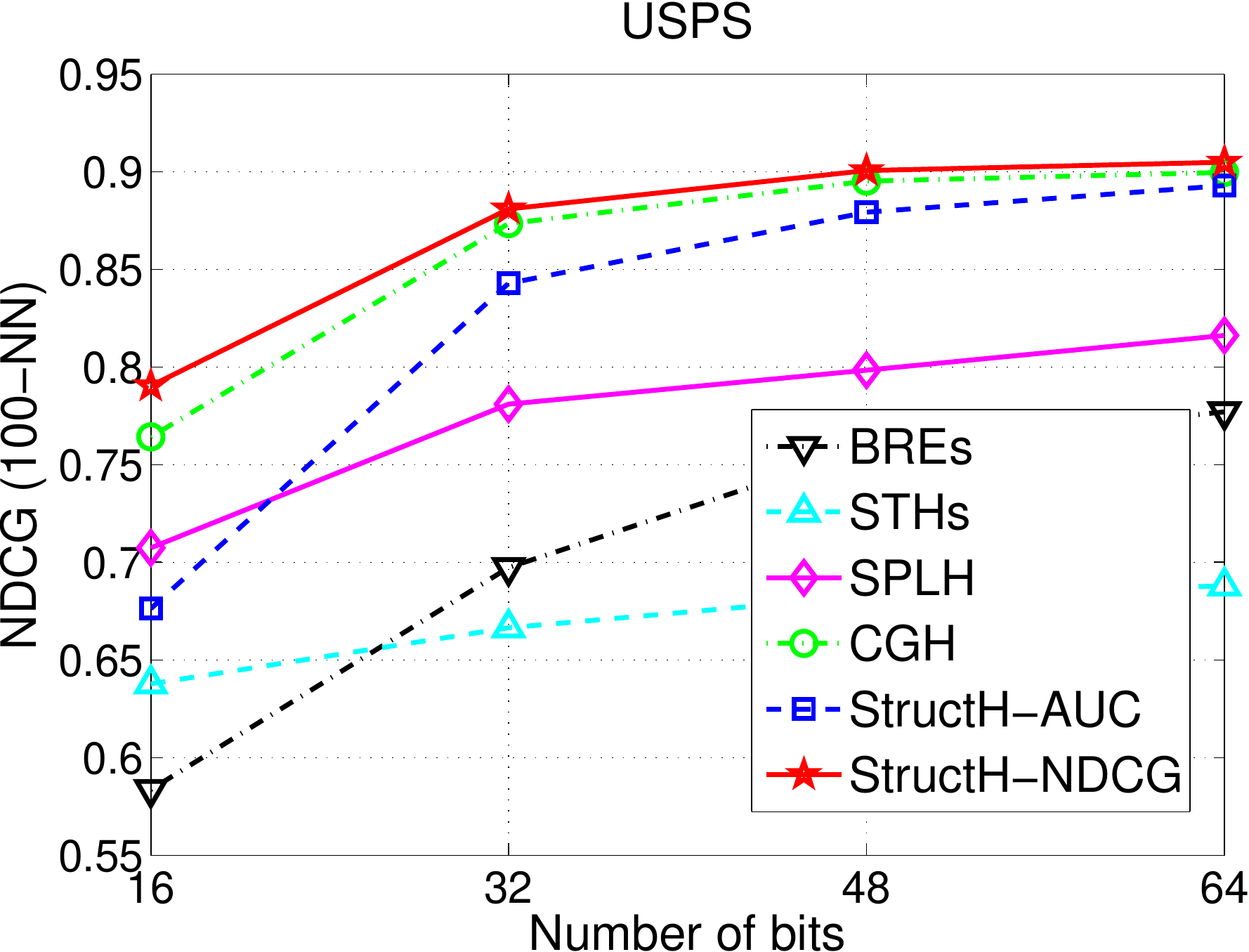}
   \includegraphics[width=.3\linewidth]{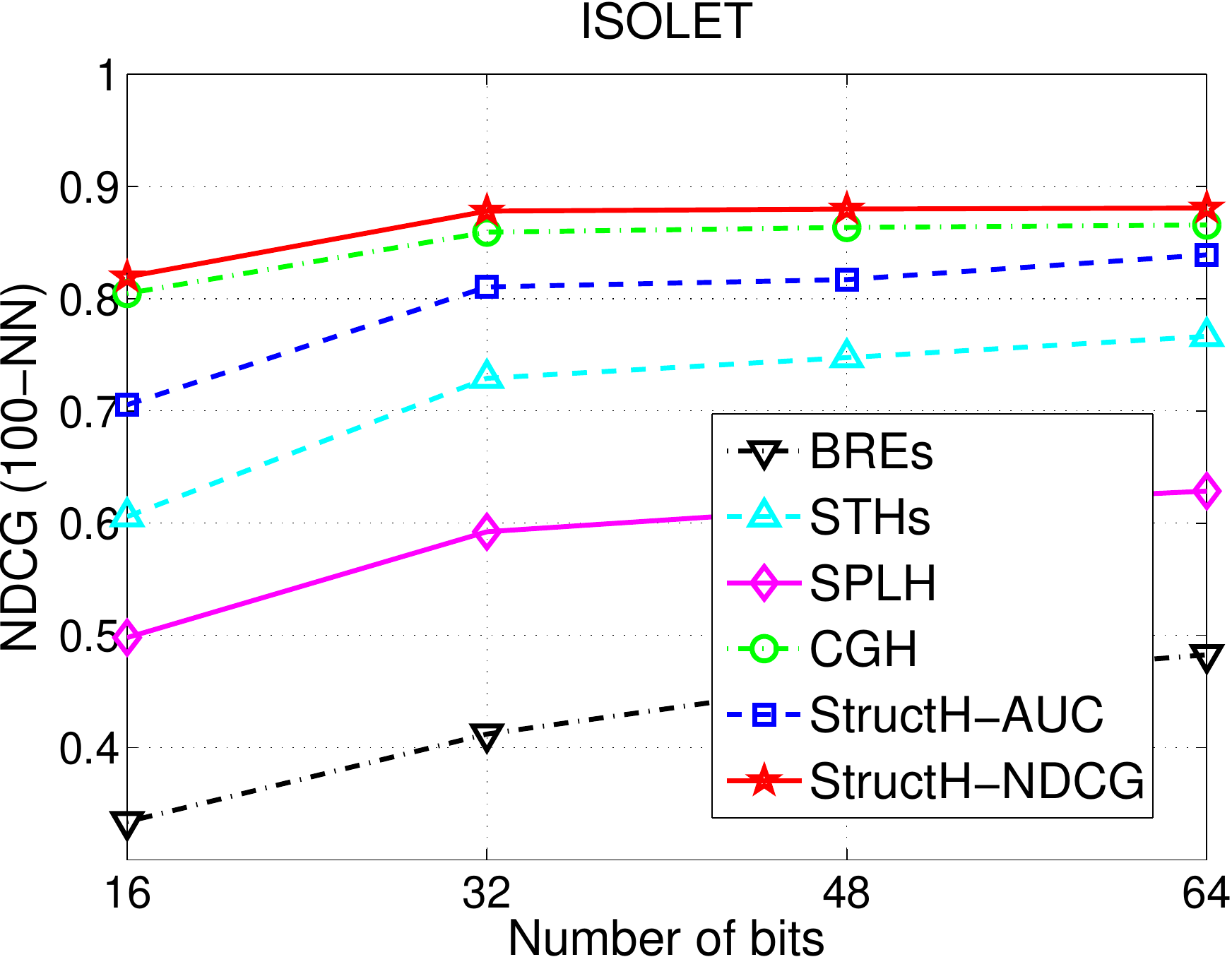}
 \caption{NDCG results on 3 datasets. Our StructHash performs the best.}
    \label{fig:ndcg_curve}
\end{figure}

\begin{figure}[t]
    \centering

  \includegraphics[width=.3\linewidth]{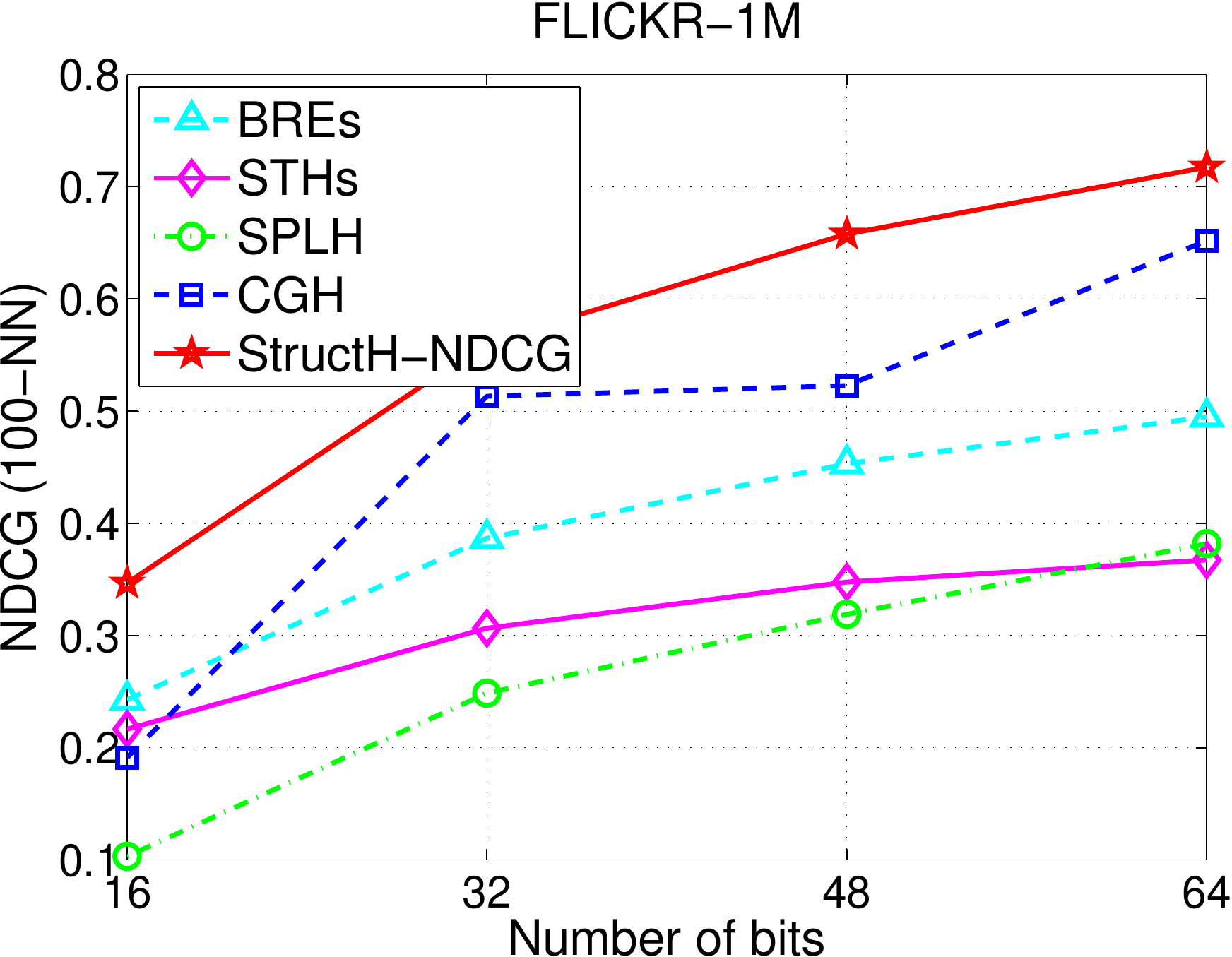}
  \includegraphics[width=.3\linewidth]{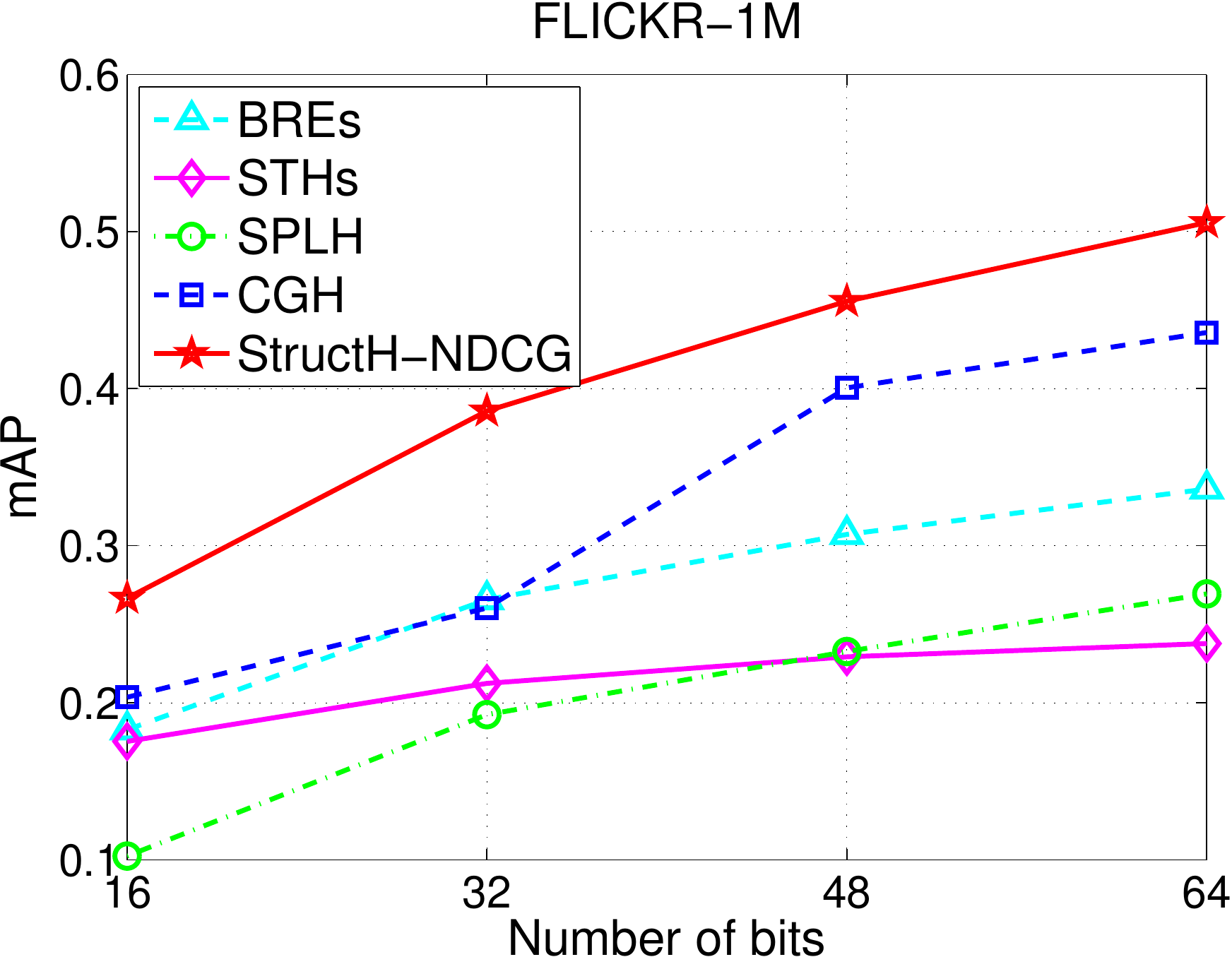}
  \includegraphics[width=.3\linewidth]{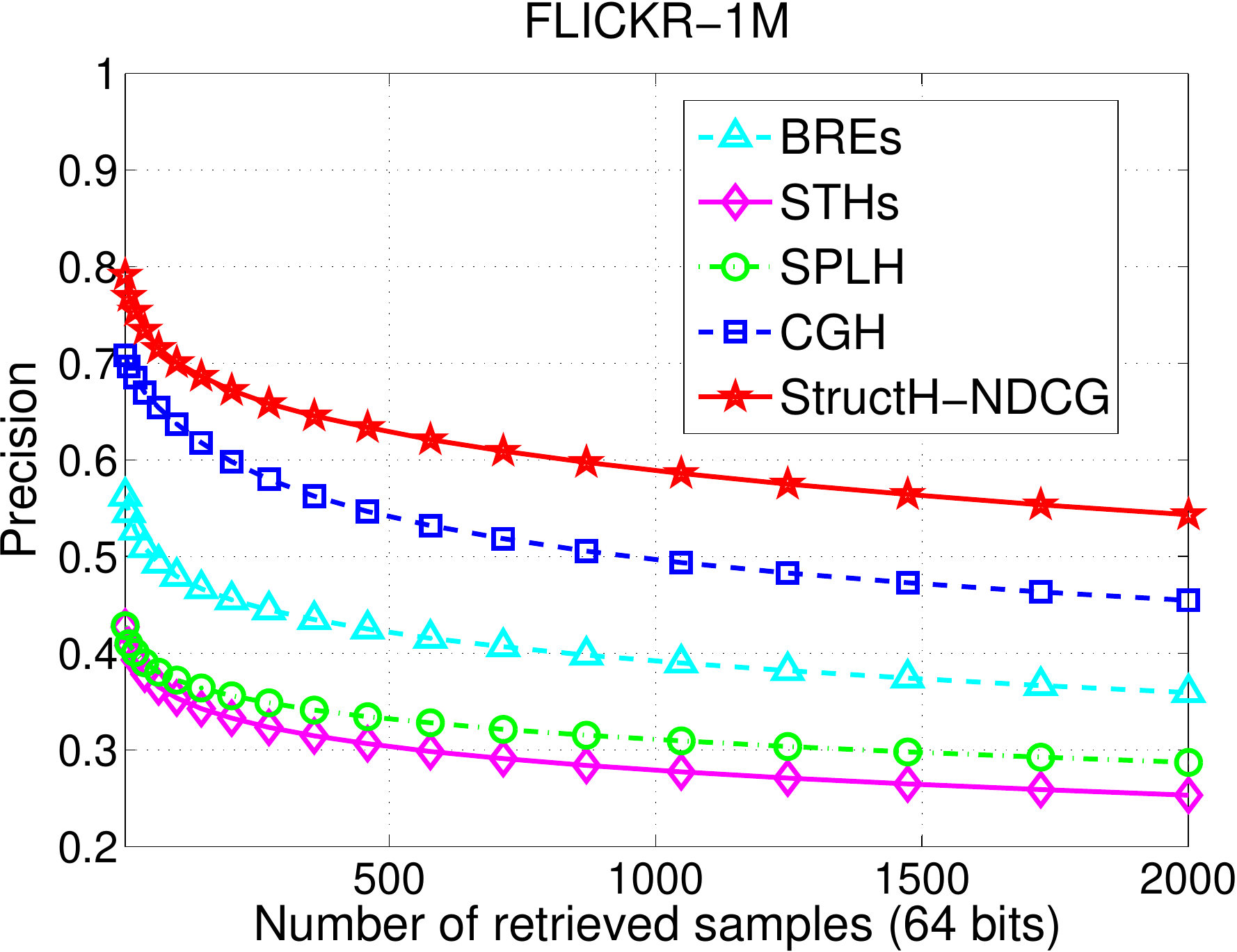}

  \includegraphics[width=.3\linewidth]{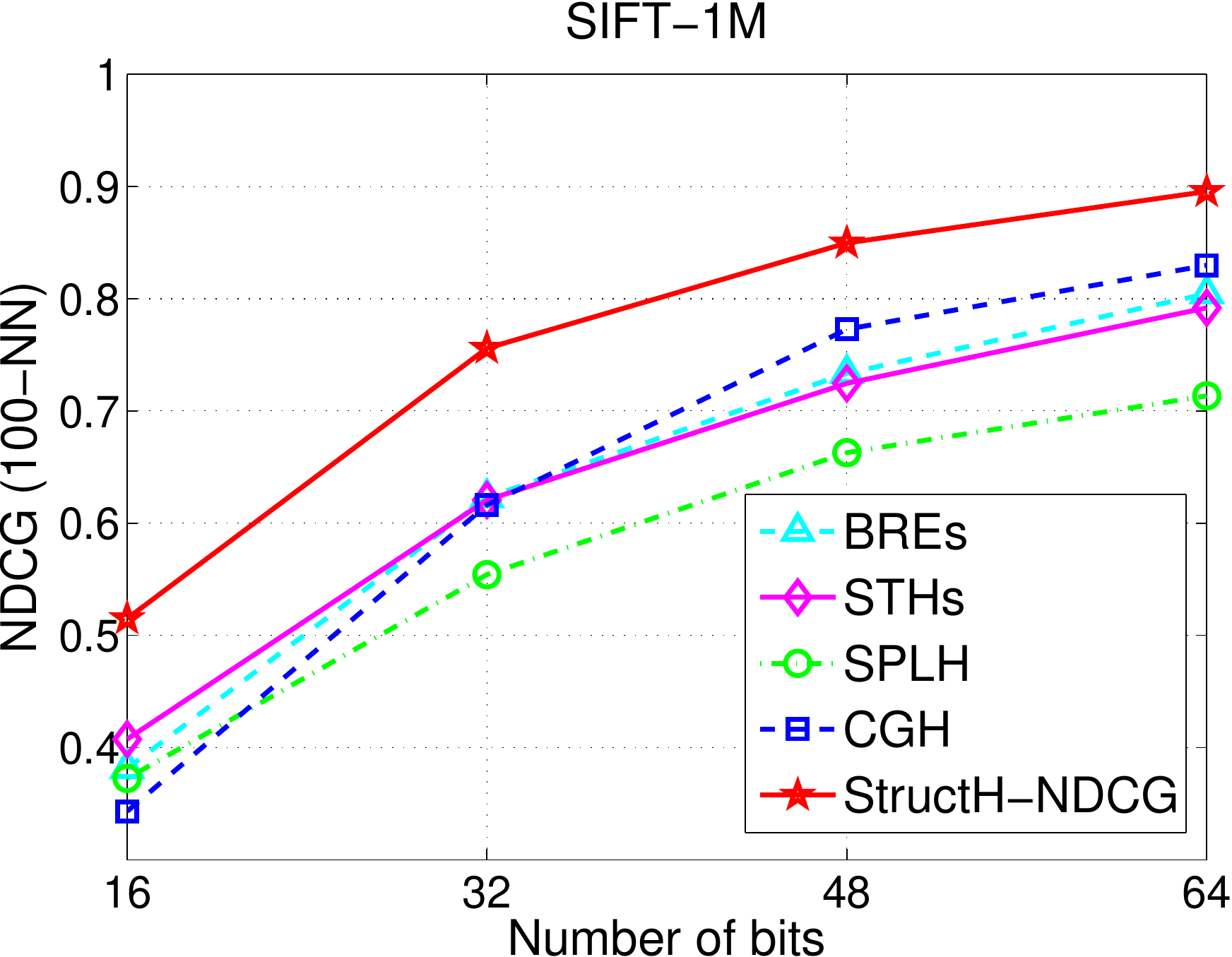}
  \includegraphics[width=.3\linewidth]{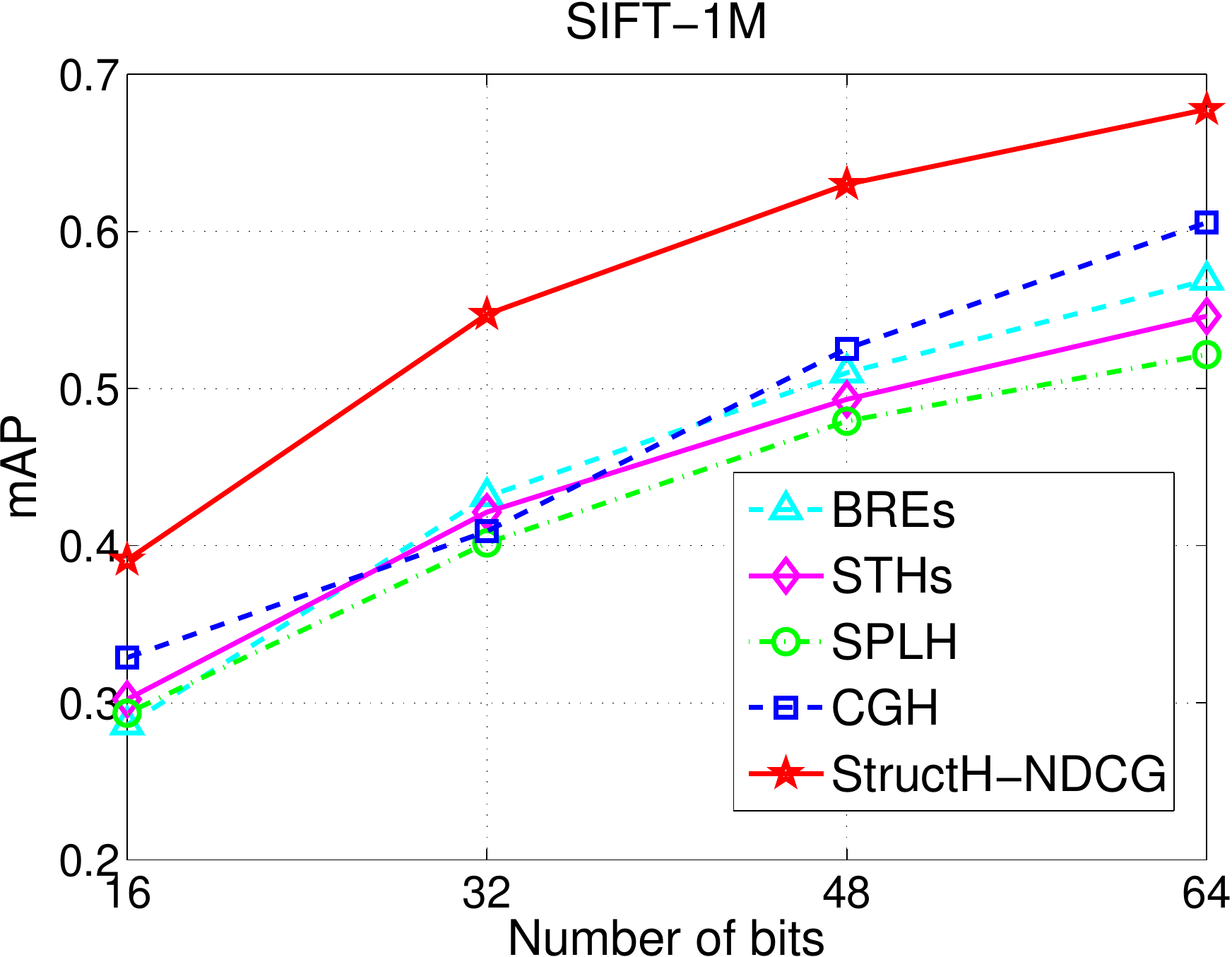}
  \includegraphics[width=.3\linewidth]{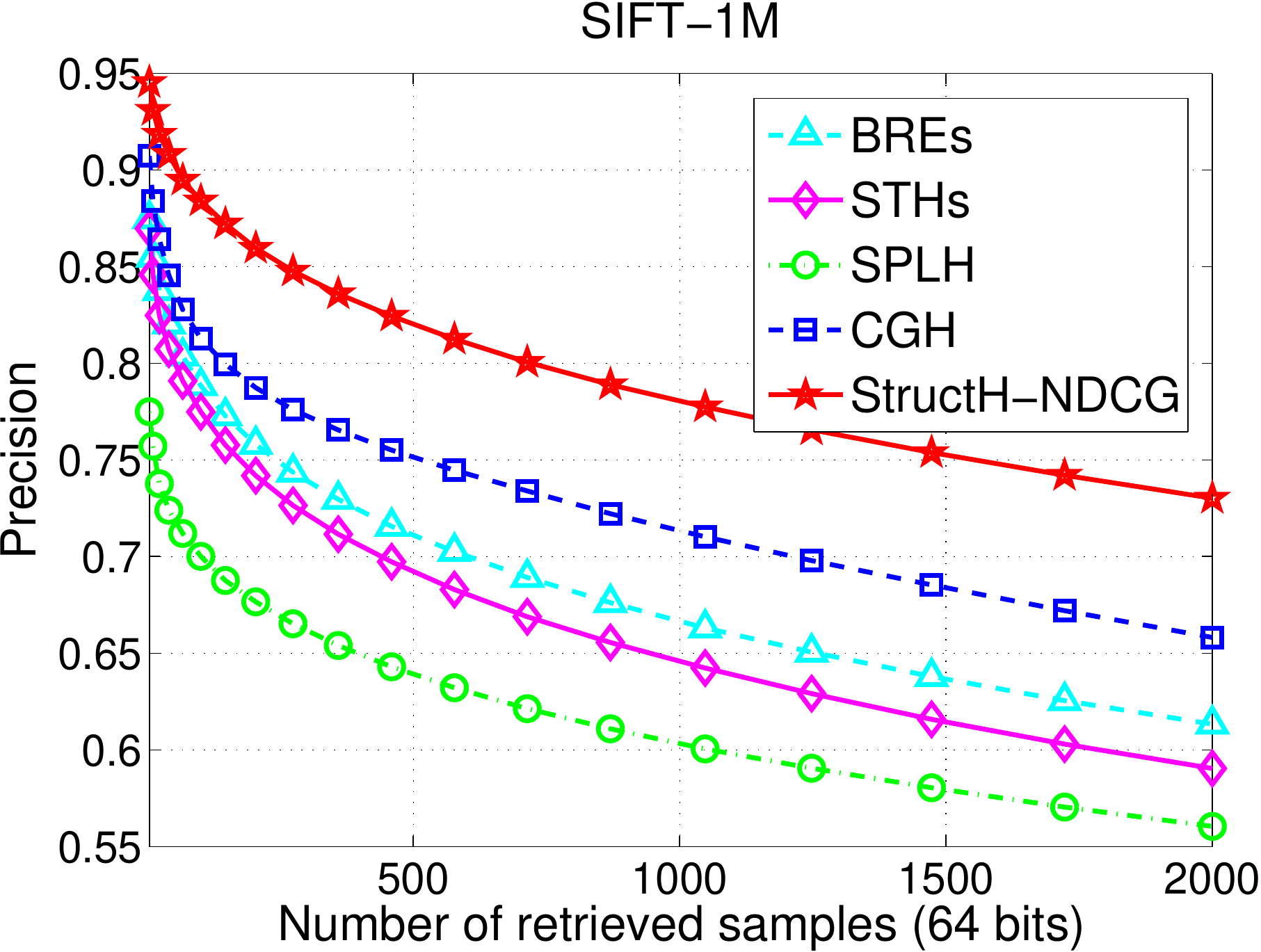}

  \includegraphics[width=.3\linewidth]{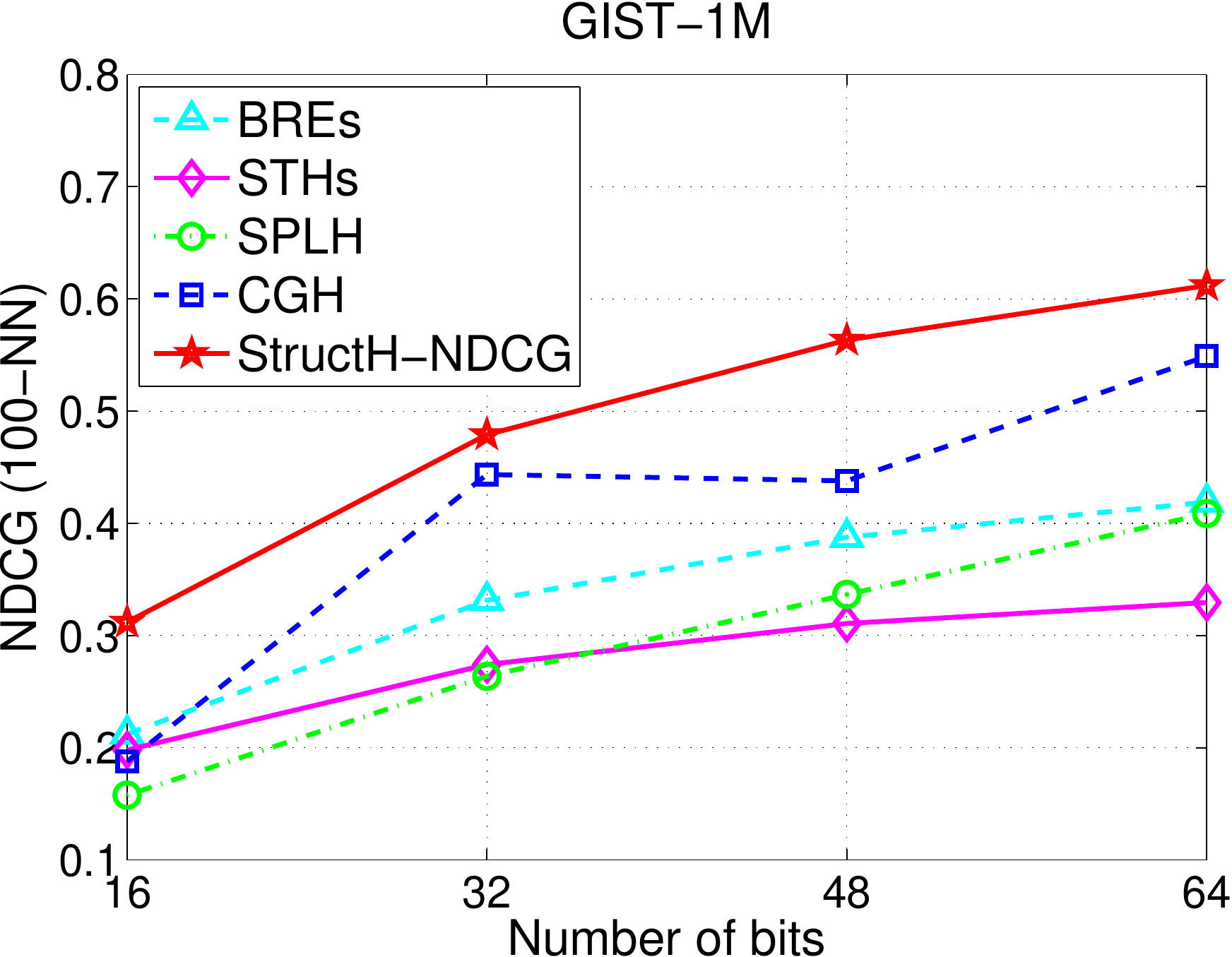}
  \includegraphics[width=.3\linewidth]{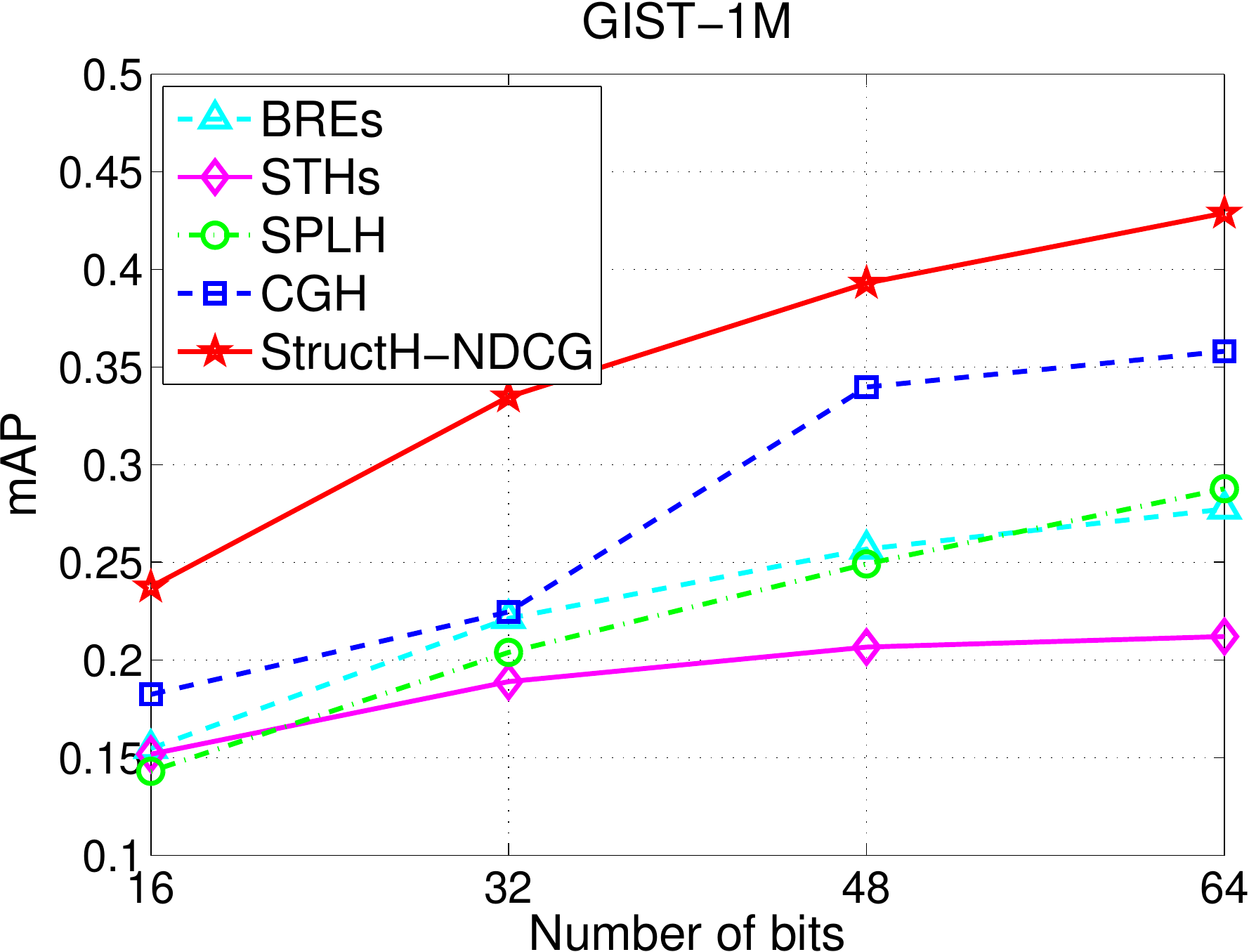}
  \includegraphics[width=.3\linewidth]{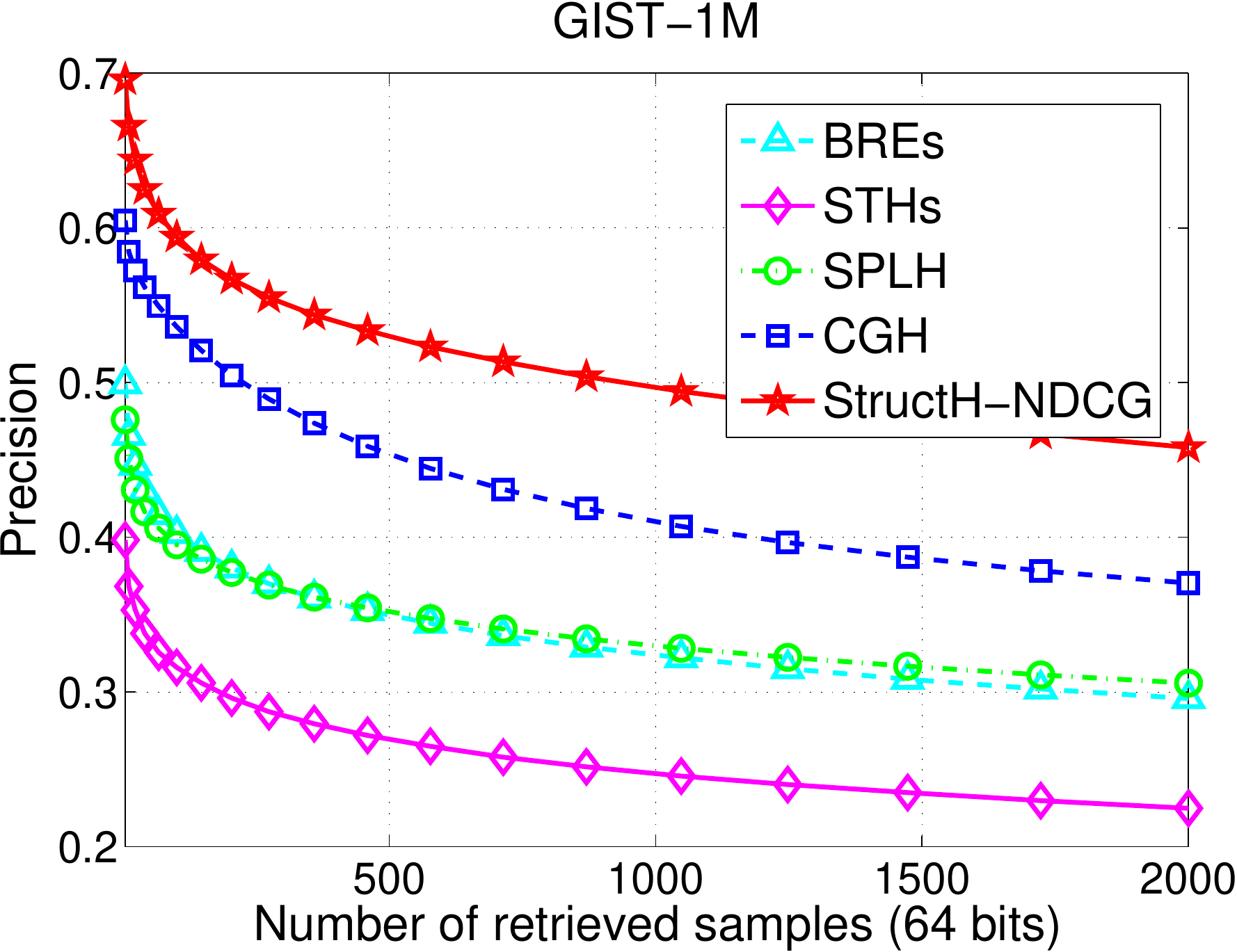}

  \includegraphics[width=.3\linewidth]{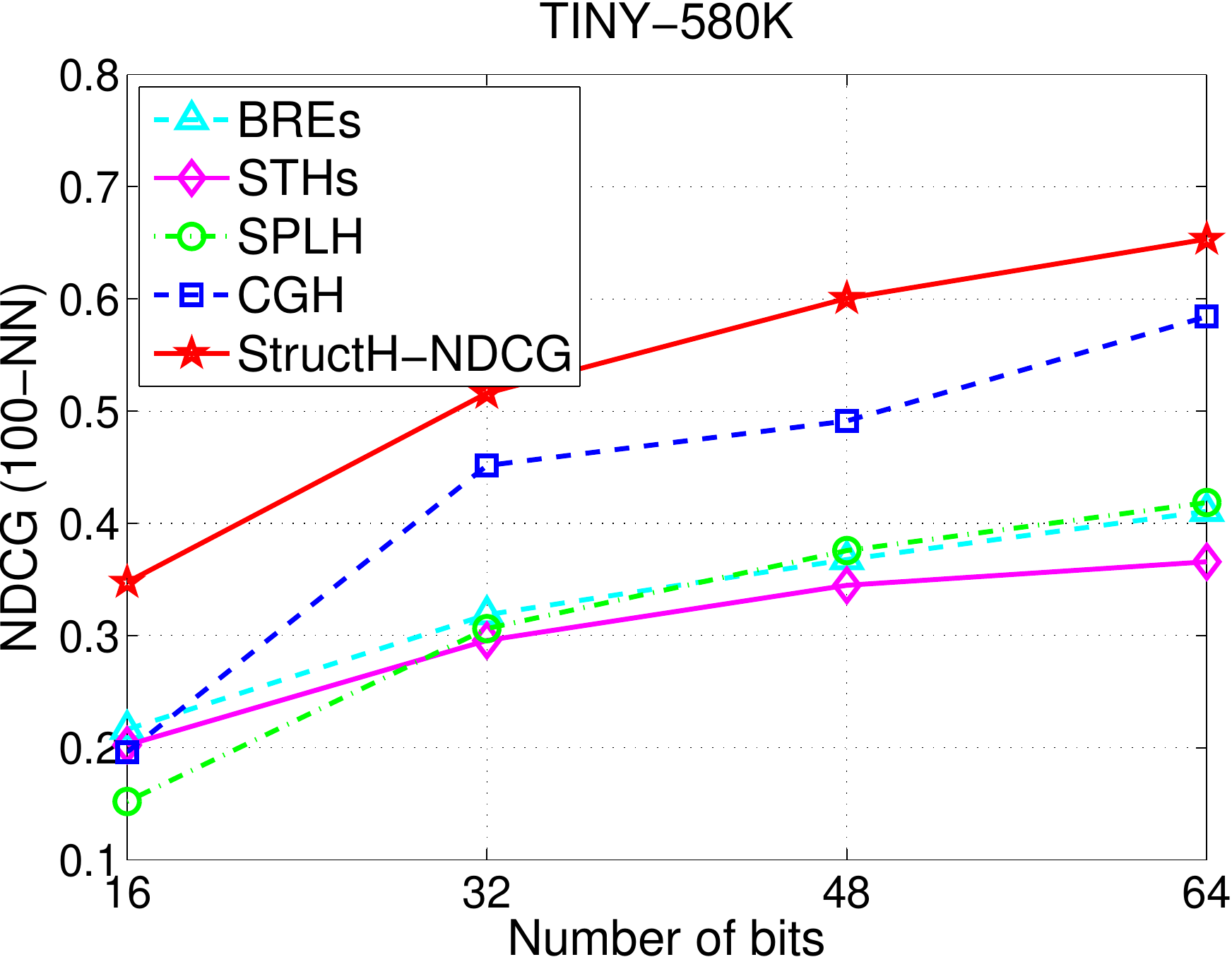}
  \includegraphics[width=.3\linewidth]{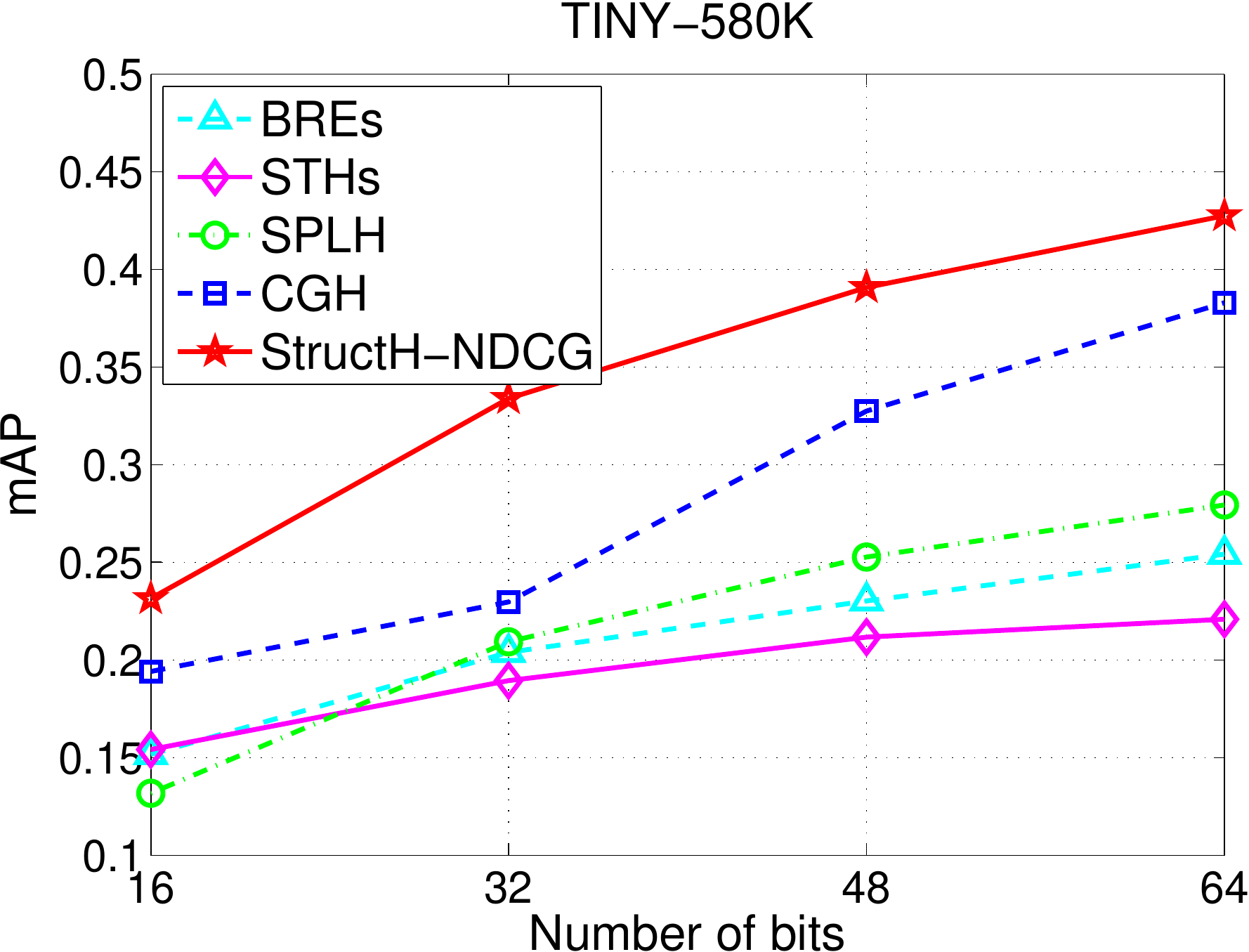}
  \includegraphics[width=.3\linewidth]{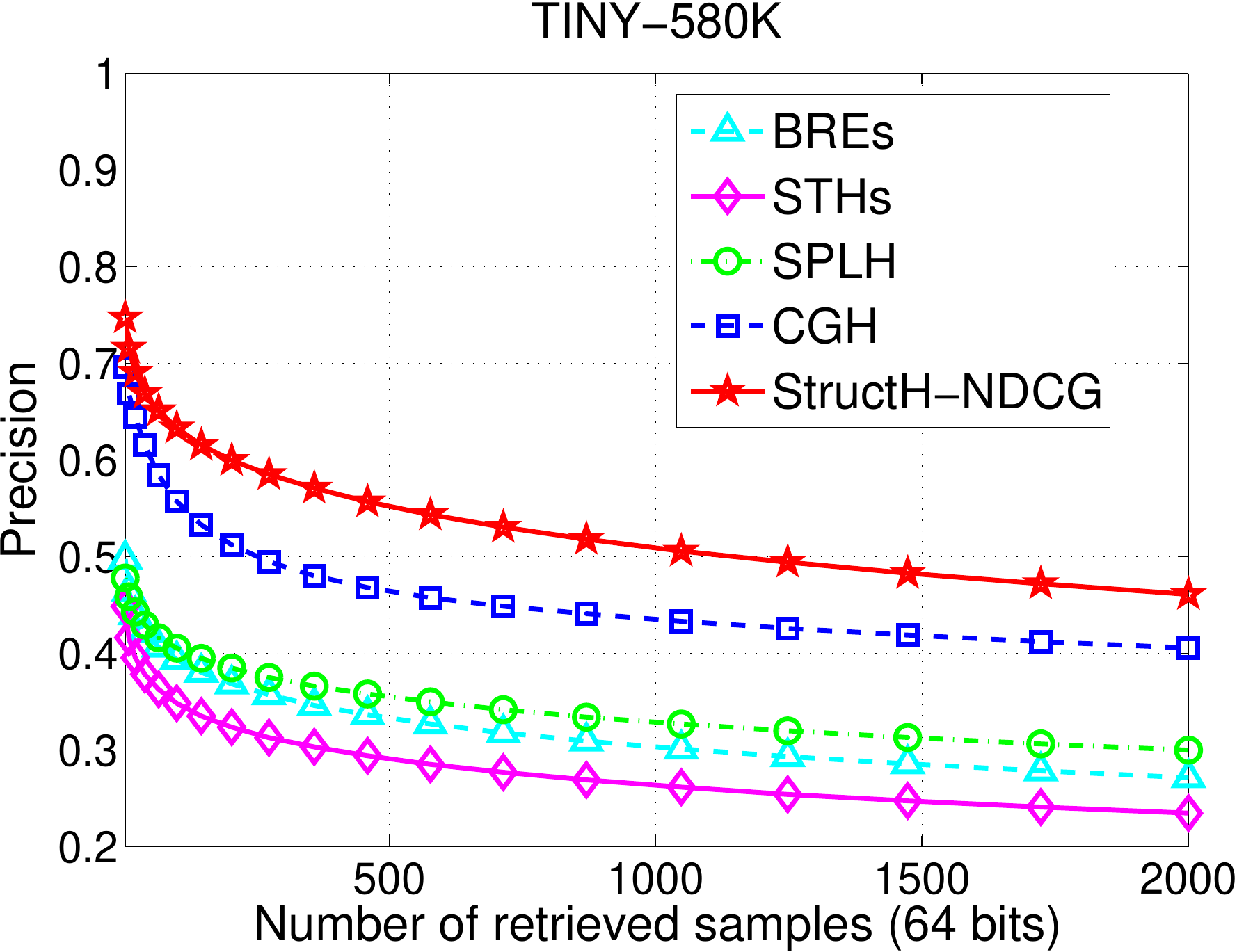}

 \caption{Results on 4 large datasests: Flickr-1M (1 million Flickr images), Sift-1M (1 million SIFT features), Gist-1M (1 million GIST features) and Tiny580K ($580,000$ Tiny image dataset). 
 We compare with several supervised methods.
 The results of 3 measures (NDCG, mAP and precision of top-K neighbours) are shown here.
 Our \sh outperforms others in most cases.} \label{fig:large}
\end{figure}

\begin{figure}[t]
    \centering

  \includegraphics[width=.3\linewidth]{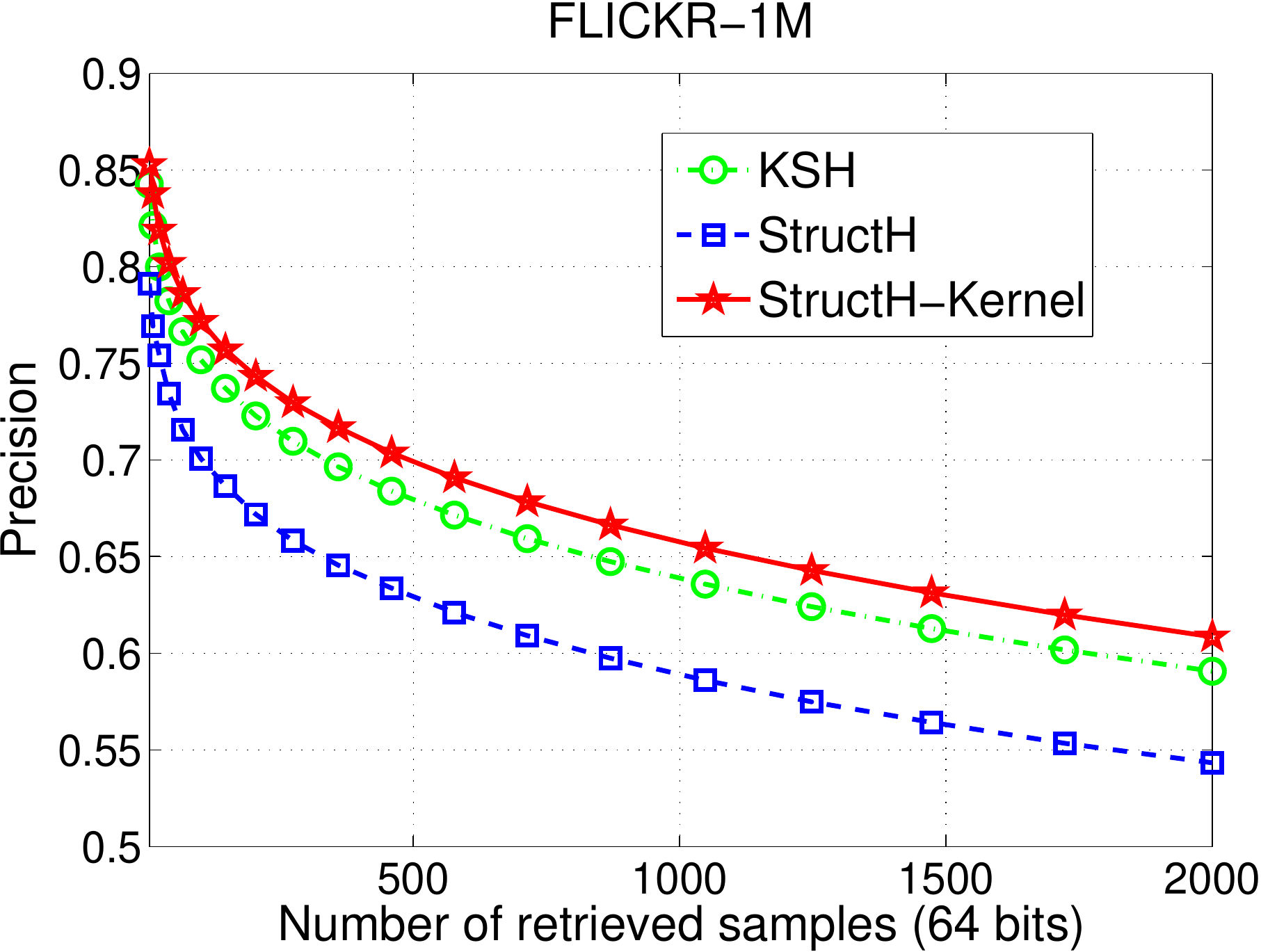}
  \includegraphics[width=.3\linewidth]{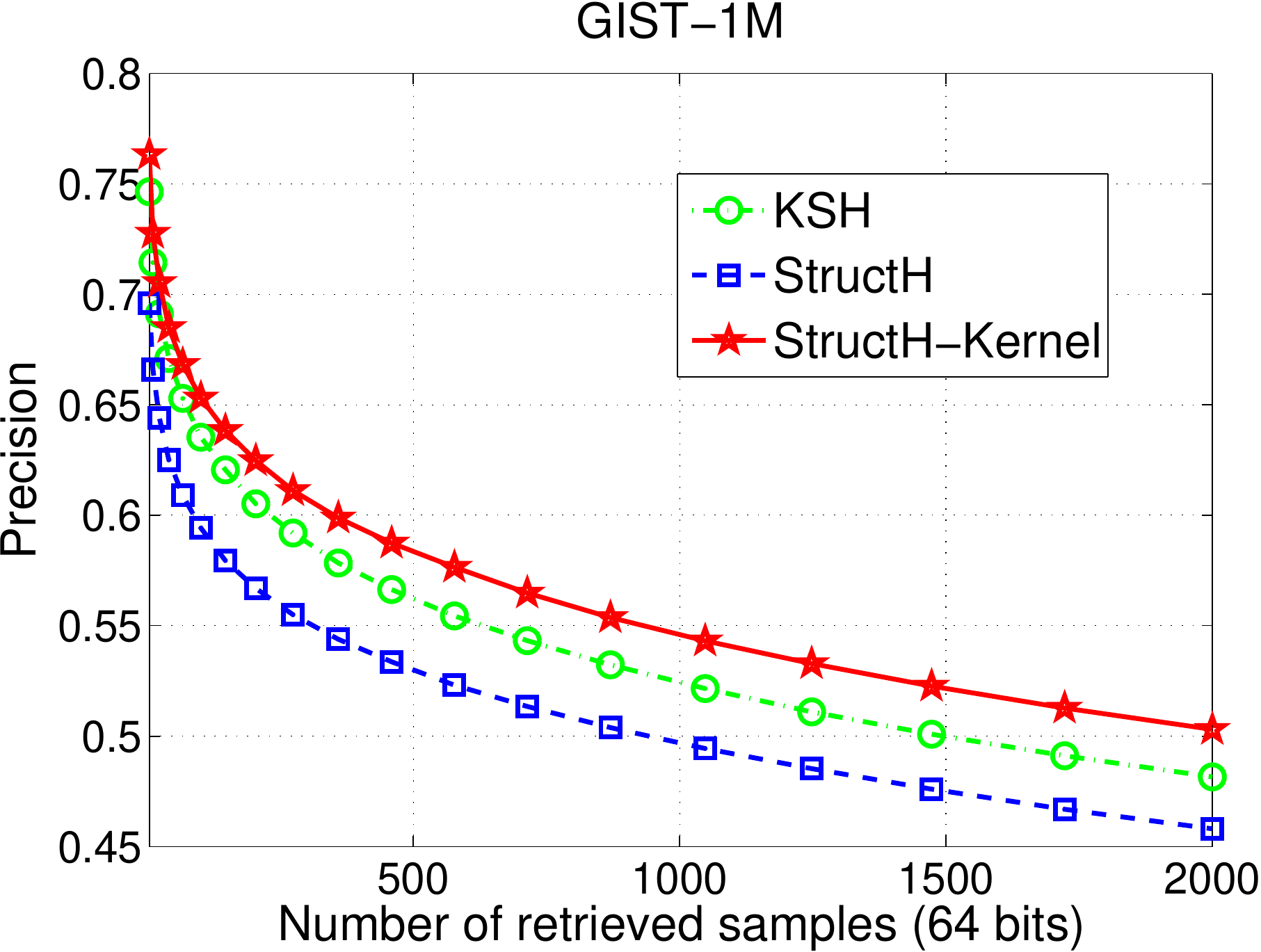}
  \includegraphics[width=.3\linewidth]{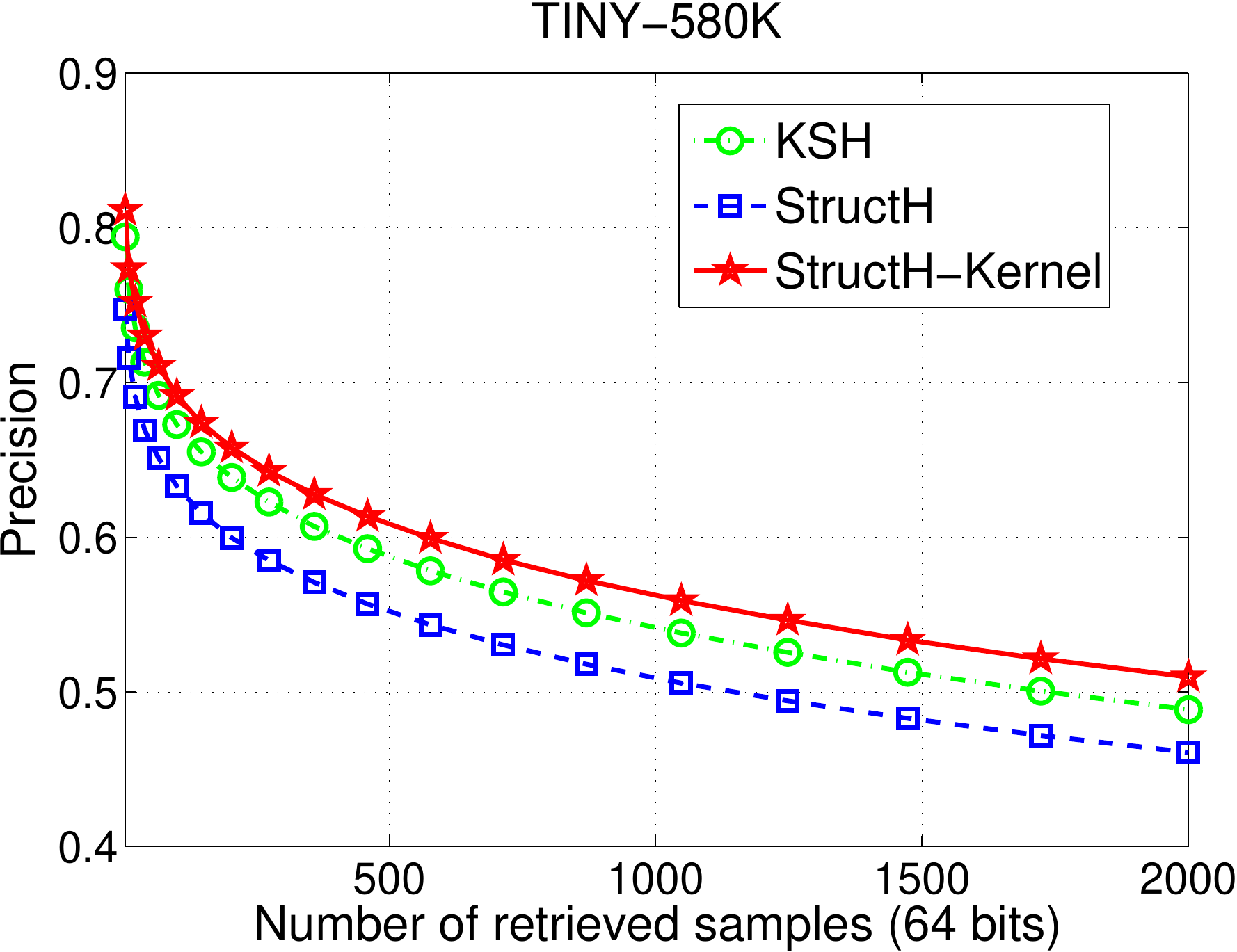}

 \caption{Comparison on large datasets of our kernel \sh (StructHash-Kernel) with our non-kernel StructHash and the relevant method KSH \citep{KSH}. Our kernel version is able to achieve better results.} \label{fig:ksh}
\end{figure}

\begin{figure}[t]
    \centering

        \figcenter{{\includegraphics[height=0.25in]{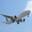}}}
                \colseperator
\figcenter{{\includegraphics[height=0.44in]{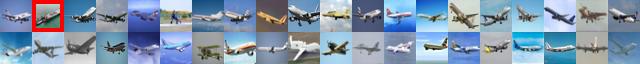}}}

        \figcenter{\includegraphics[height=0.25in]{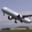}}
        \colseperator
        \figcenter{\includegraphics[height=0.44in]{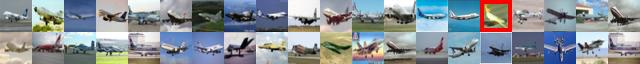}}

  \figcenter{\includegraphics[height=0.25in]{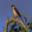}}
       \colseperator
       \figcenter{\includegraphics[height=0.44in]{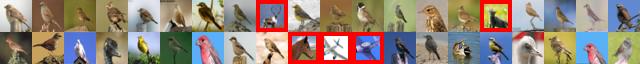}}

         \figcenter{\includegraphics[height=0.25in]{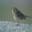}}
       \colseperator
       \figcenter{\includegraphics[height=0.44in]{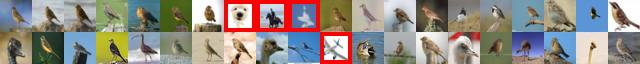}}

       \figcenter{\includegraphics[height=0.25in]{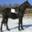}}
       \colseperator
       \figcenter{\includegraphics[height=0.44in]{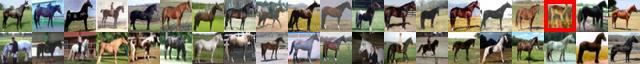}}

       \figcenter{\includegraphics[height=0.25in]{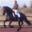}}
       \colseperator
       \figcenter{\includegraphics[height=0.44in]{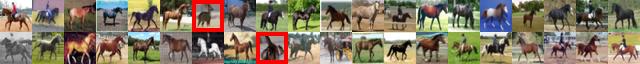}}

       \figcenter{\includegraphics[height=0.25in]{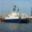}}
       \colseperator
       \figcenter{\includegraphics[height=0.44in]{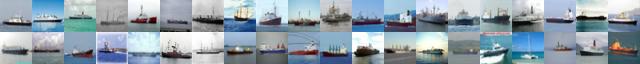}}

       \figcenter{\includegraphics[height=0.25in]{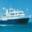}}
       \colseperator
       \figcenter{\includegraphics[height=0.44in]{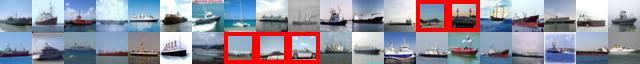}}

    \caption{Some ranking examples of \sh. The first column shows query images, and the rest are retrieved images. False predictions are marked by red boxes.}
    \label{fig:examples_cifar}
\end{figure}

\subsection{State-of-the-art comparisons}
\label{sec:eval_state}
Our method is in the category of supervised method for learning compact binary codes. 
Thus we mainly compare with 3 supervised methods: supervised binary reconstructive embeddings
(BREs) \citep{kulis2009learning}, supervised self-taught hashing (STHs)
\citep{zhangSTHs}, semi-supervised sequential projection learning
hashing (SPLH) \citep{wang2010semi}.
We also run some unsupervised methods for comparisons:
locality-sensitive hashing (LSH) \citep{Gionis1999}, anchor graph hashing
(AGH) \citep{liu2011hashingGraphs}, spherical hashing (SPHER) \citep{jae2012},
multi-dimension spectral hashing (MDSH) \citep{MDSH}, and iterative quantization (ITQ) \citep{gong2012iterative}.
We carefully follow the original authors' instruction for parameter setting.
For SPLH, the regularization parameter is
picked from $0.01$ to $1$. We use the hierarchical variant of AGH. The
bandwidth parameters of Gaussian affinity in MDSH is set as $\sigma=t\bar{d}$. Here $\bar{d}$ is the average
Euclidean distance of top $100$ nearest neighbours and $t$ is picked
from $0.01$ to $50$.
For supervised training of our \sh and \cgh, we use 50
relevant and 100 irrelevant examples
to construct similarity information for each data point.

We report the result of the NDCG measure in Table \ref{tab:main}.
We compare our \sh using AUC and NDCG loss functions, and our \cgh for triplet loss with other supervised and unsupervised methods.  \sh using NDCG loss function performs the best in most cases.
We also report the result of other common measures in Table \ref{tab:main-other},
including the result of Precision-at-K, Mean Average Precision (mAP) and Precision-Recall.
Precision-at-K is the proportion of true relevant data points in the returned top-K results.
The Precision-Recall curve measures the overall performance in all positions of the prediction ranking, which
is computed by varying the number of nearest neighbours.
It shows that our method generally performs better than other methods on these evaluation measures.
As described before, compared to the AUC measure which is position insensitive,
the NDCG measure assigns different importance on ranking positions, which is closely related to many other position sensitive ranking measures (e.g., mAP).
As expected, the result shows that on the Precision-at-K, mAP and Precision-recall measures,
optimizing the position sensitive NDCG loss performs better than the AUC loss.
\sh with AUC loss actually minimize the triplet loss, hence it achieve similar performance with 
our triplet loss based method \cgh.
\sh with the NDCG loss which is position insensitive is able to outperform \cgh in these measures.
We also plot the NDCG results on several datasets in Fig. \ref{fig:ndcg_curve} by varying the number of bits.
Some retrieval examples are shown in Fig. \ref{fig:examples_cifar}.

We further evaluate our method on 4 large-scale datasets (Flickr-1M, SIFT-1M, GIST-1M and Tiny-580K).
The results of NDCG, mAP and the precision of top-K neighbours are shown in Fig. \ref{fig:large}.
The NDCG and mAP results are shown by varying the number of bits.
The precision of top-K neighbours is shown by varying the number of retrieved examples.
In most cases, our method outperforms other competitors.
Our method with NDCG loss function succeeds to achieve good performance both on NDCG and other measures.

Applying the kernel technique in KLSH \citep{KLSH} and KSH \citep{KSH} further improves the performance of our method. As describe in \citep{KSH}, we perform a pre-processing step to generate the kernel mapping features: we randomly select a number of support vectors (300) then compute the kernel response on data points as input features.
Note that here we simply follow KSH for the kernel parameter setting.
We evaluate this kernel version of our method in Fig. \ref{fig:ksh} and compare to KSH.
Our kernel version is able to achieve better results.

\subsection{Evaluation of the extensions of \sh for efficient learning}
\label{sec:eval_new}
In this section, we evaluate the two extensions of \sh proposed in Sec.~\ref{sec:speedup} for efficient learning.
Specifically, we denote the extension of using efficient stage-wise training as StructH-NDCG-Stage, which uses the original NDCG loss;
we denote the second extension as StructH-SNDCG-stage which also applies stage-wise training but uses the proposed efficient Simplified NDCG loss instead.
We mainly compare this two extensions with the original version of the StructHash with the NDCG loss, denoted as StructH-NDCG.

Table \ref{tab:cmp-speedup} reports the compared results on 5 datasets using different ranking measures.
As we can see, the two efficient extensions, the StructH-NDCG-Stage and the StructH-SNDCG-Stage, 
generally performs better or comparable with the original method StructH-NDCG.

\begin{figure}[t]
\begin{tabular}{cccc}
    \centering
   \includegraphics[width=.3\linewidth, height=.24\linewidth]{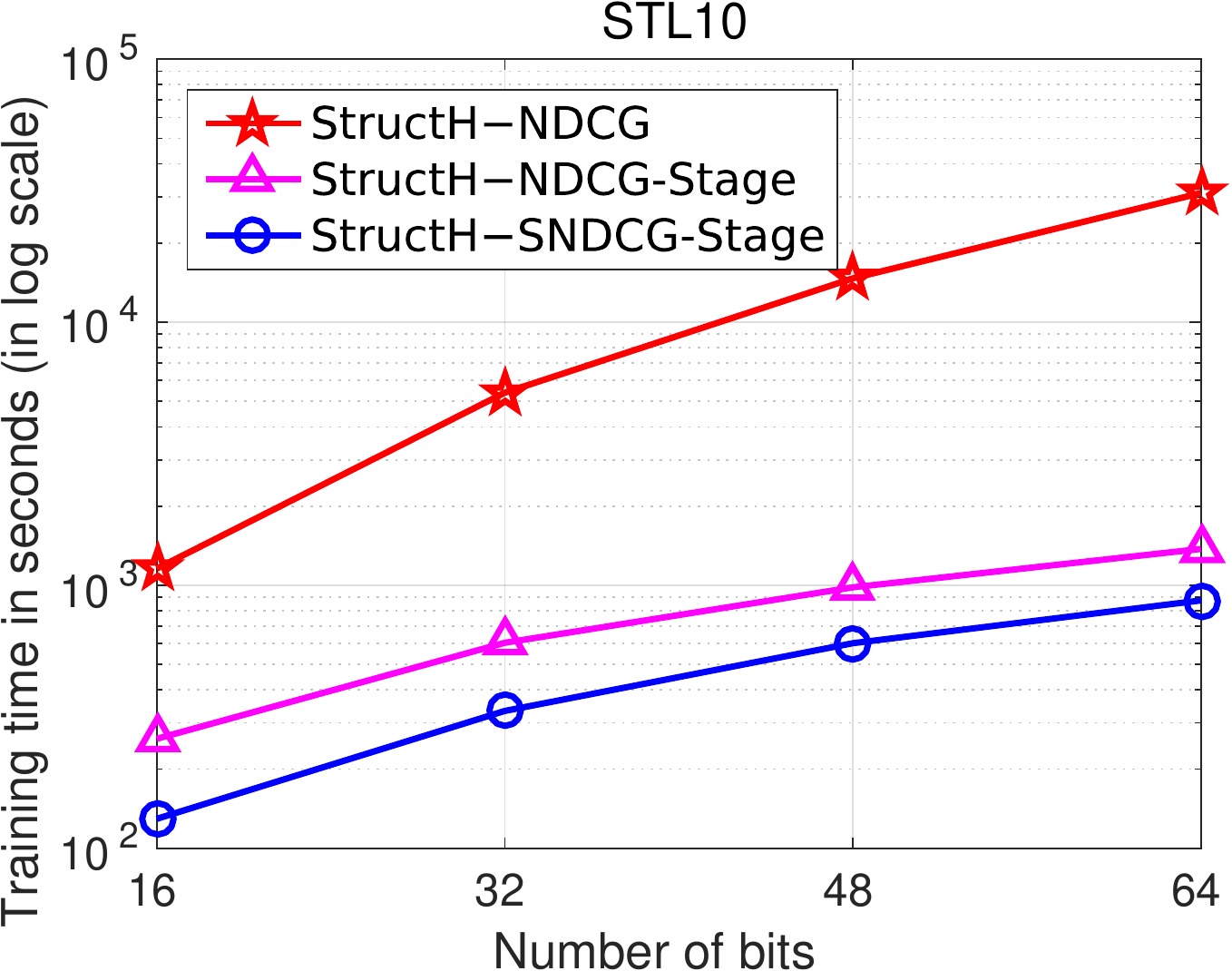}
   &\includegraphics[width=.3\linewidth, height=.24\linewidth]{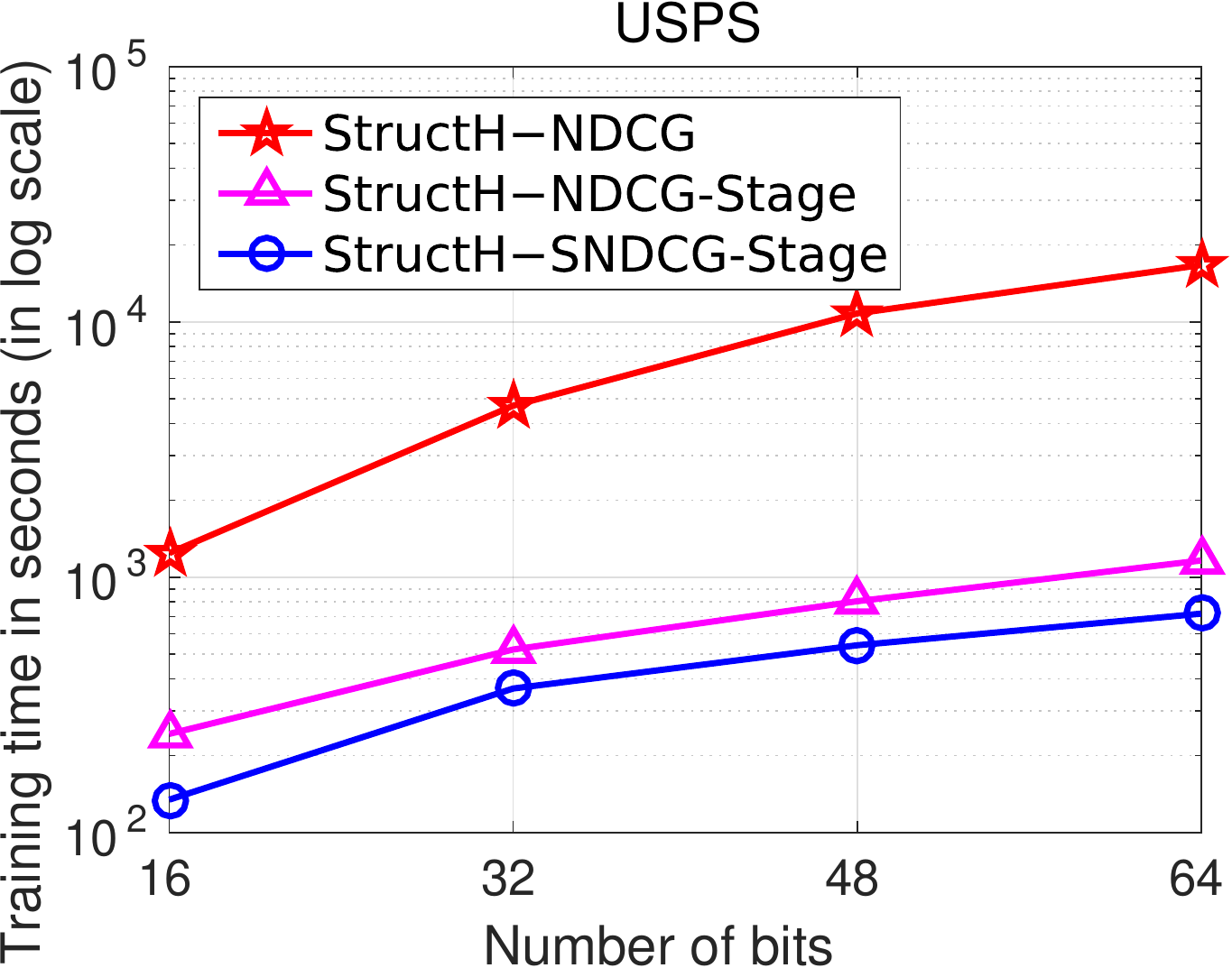}
   &\includegraphics[width=.3\linewidth, height=.24\linewidth]{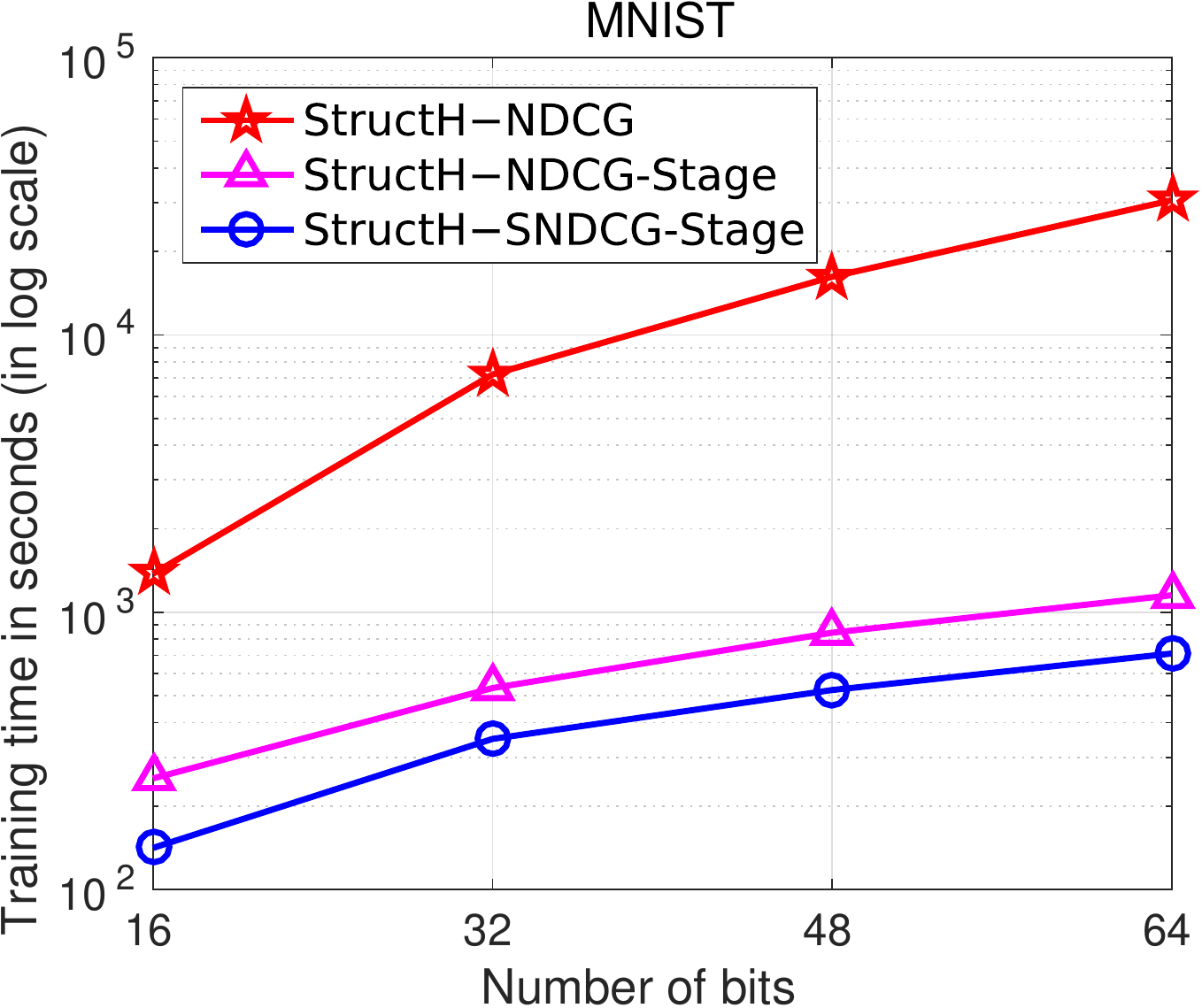} \\
   
   \includegraphics[width=.3\linewidth, height=.24\linewidth]{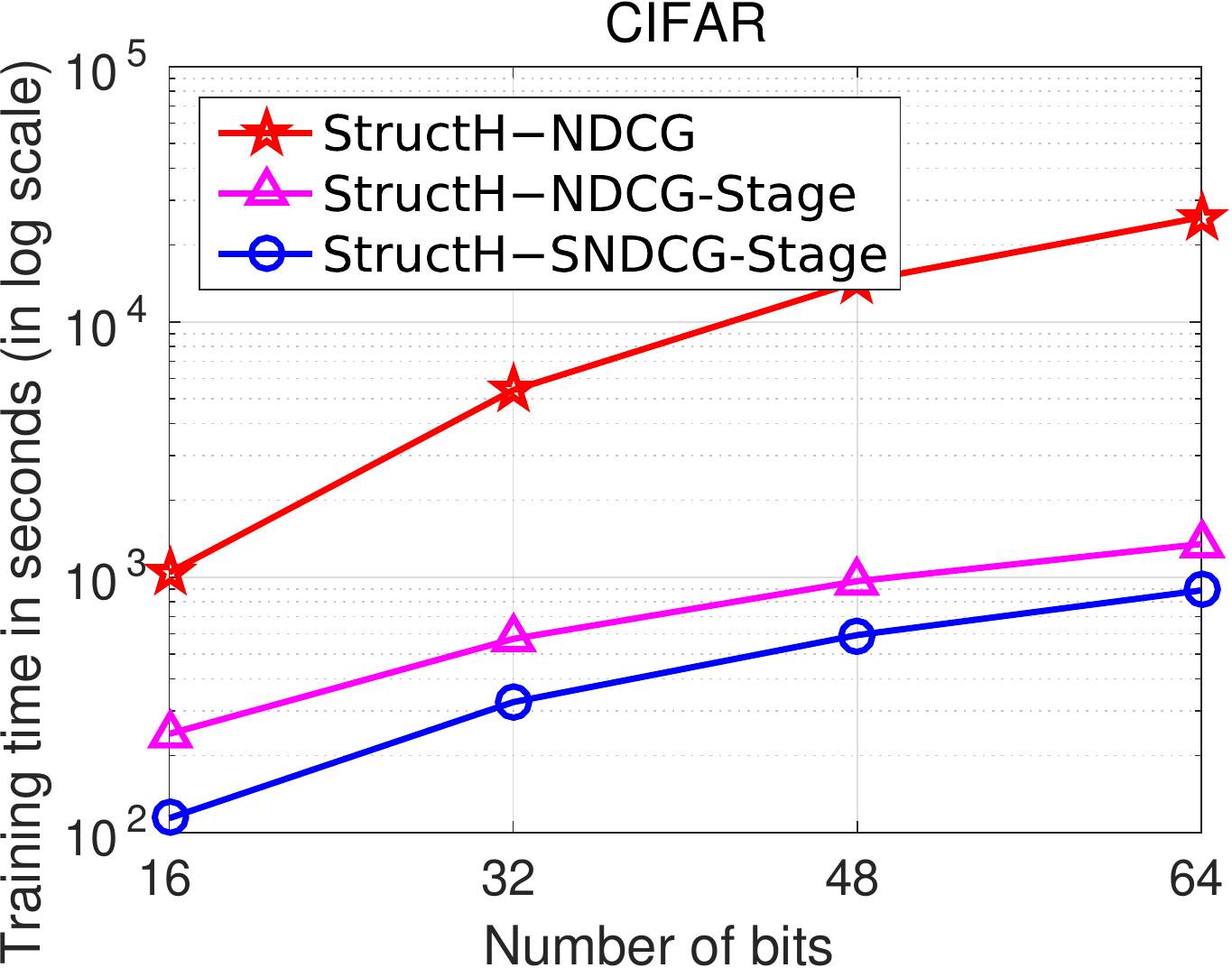}
   &\includegraphics[width=.3\linewidth, height=.24\linewidth]{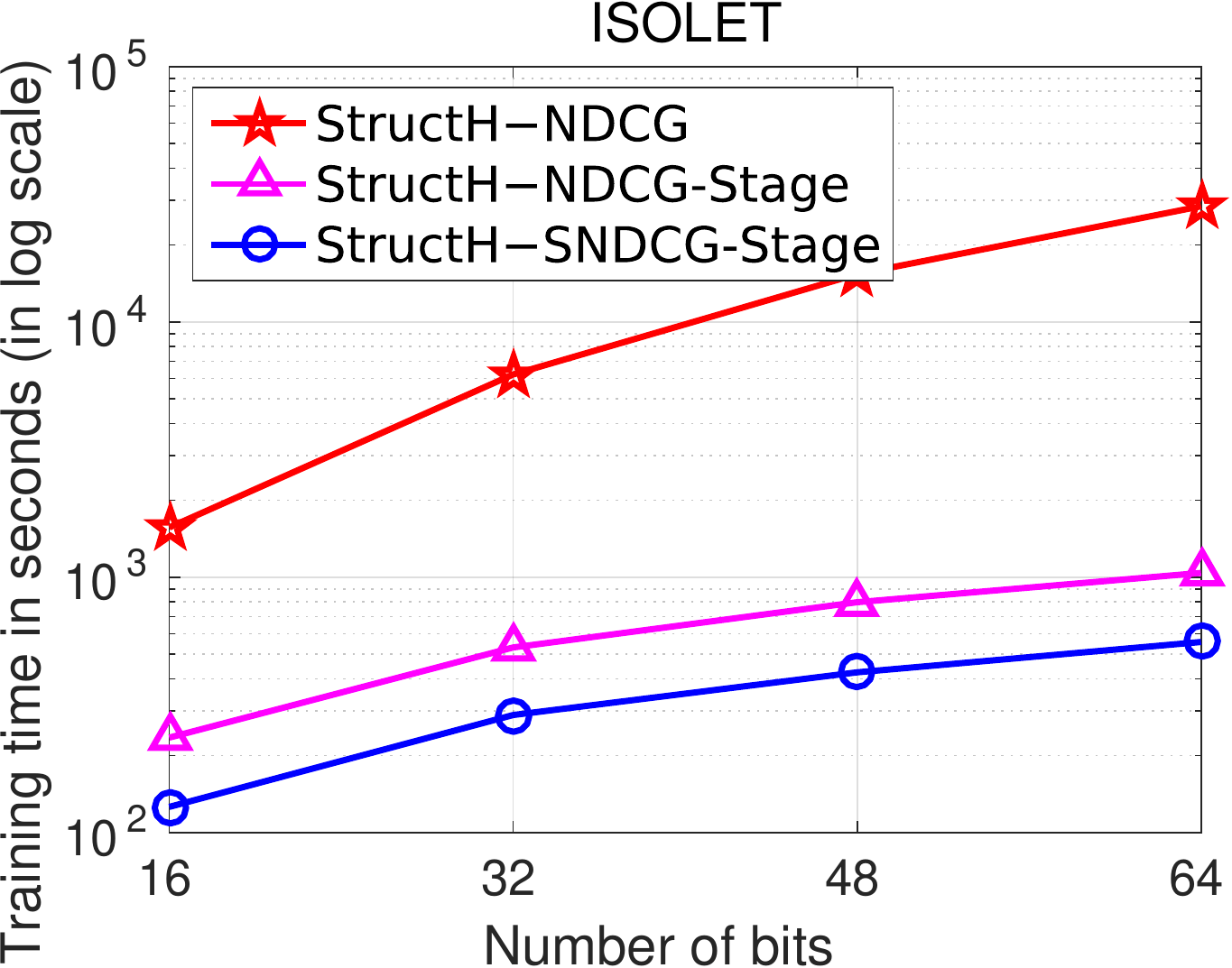}
   &\includegraphics[width=.3\linewidth, height=.24\linewidth]{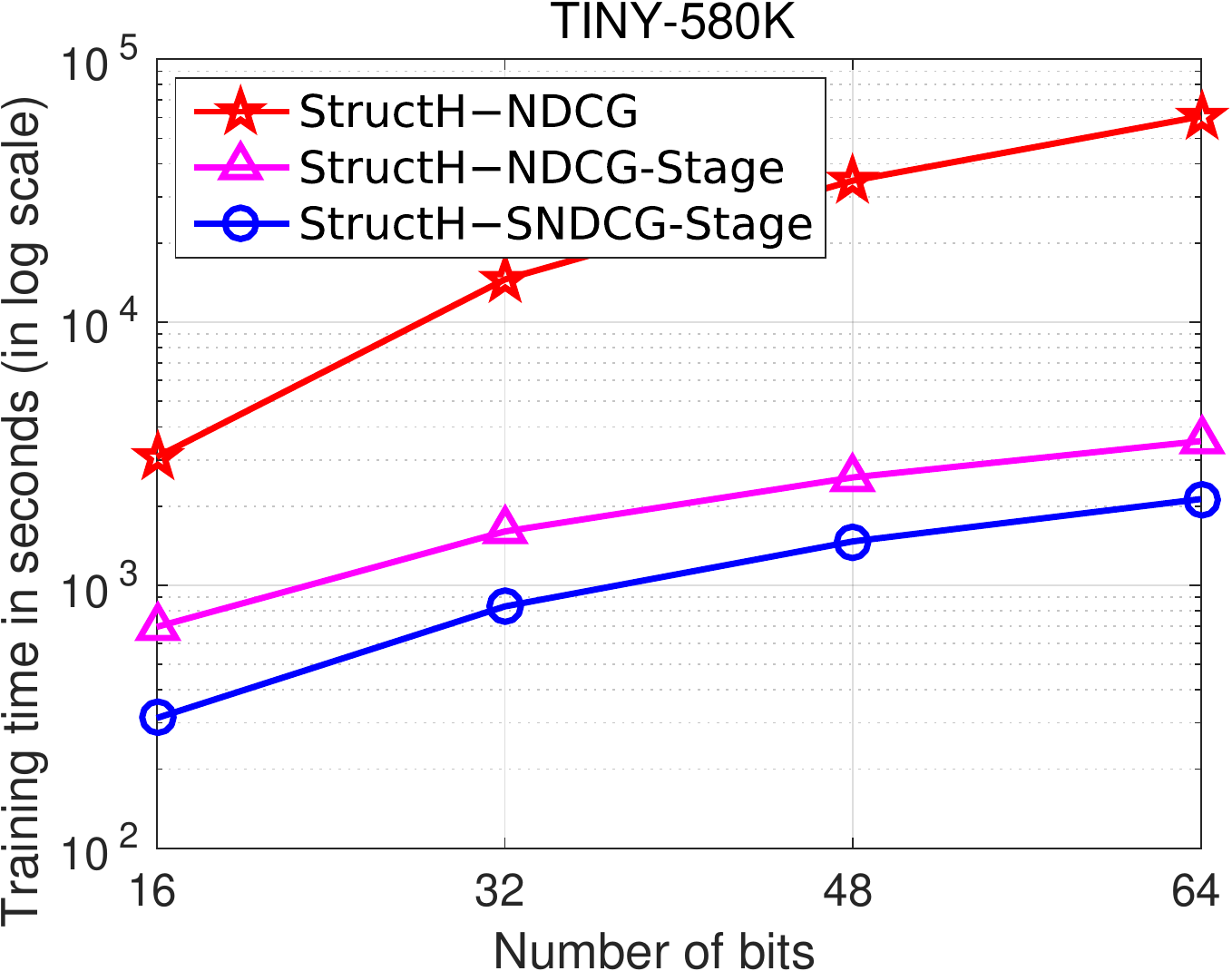}   
\end{tabular}
 \caption{Comparisons of training time in seconds (in log scale) in terms of different hashing bits on 6 datasets. Our StructHash-NDCG-Stage with stage-wise training is generally orders of magnitude faster than the original StructHash-NDCG. It also shows that using simplified NDCG loss (StructHash-SNDCG-Stage) is twice faster than using the original NDCG loss (StructHash-NDCG-Stage).}
    \label{fig:cmp_time}
\end{figure}

We further compare these two efficient models against the original model in terms of training time.
The experiments are conducted on a standard PC machine with 16G memory.
Fig. \ref{fig:cmp_time} shows the compared results. 
It clearly reveals that the \sh model with stage-wise training is orders of magnitude faster than the original \sh model.
Furthermore, compared to optimizing the NDCG score, optimizing the simplified NDCG (SNDCG) score generally reduces the training time by half,
which shows the efficient inference of SNDCG significantly improve the training speed.

\begin{table}[t]
\caption{Results using ranking measures of NDCG, Precision-at-K, Mean Average Precision and Precision-Recall (64 bits).
 We compare our \sh with stage-wise training using NDCG (StructH-N-Stage) and SNDCG (StructH-SN-Stage) loss functions against the original \sh with NDCG loss (StructH-N). The StructH-N-Stage and the StructH-SN-Stage generally performs better or comparable with the StructH-NDCG.}
\centering
\resizebox{.86\linewidth}{!}
  {
  \begin{tabular}{ c || c c c | c c c }
\hline
\multirow{2}{*}{Dataset} & \multicolumn{3}{c|}{NDCG ($K=100$)} & \multicolumn{3}{c}{Precision-at-K ($K=100$)}  \\
\cline{2-7}
& StructH-N & StructH-N-Stage & StructH-SN-Stage & StructH-N & StructH-N-Stage & StructH-SN-Stage \\
\hline
STL10 &0.435   &0.441 &\bf 0.450 &0.431 &0.436  &\bf 0.445  \\
USPS  &0.905   &0.910 &\bf 0.913 &0.903 &0.906  &\bf 0.909  \\
MNIST &0.851   &0.872 &\bf 0.873 &0.849 &0.866  &\bf 0.868 \\
CIFAR &0.335   &0.386 &\bf 0.393 &0.336 &0.380  &\bf 0.388 \\
ISOLET  &0.881  &\bf 0.886  &0.884 &0.875 &\bf 0.878  &0.874  \\
TINY-580K &0.653   &\bf 0.678 &0.676 &0.634 &\bf 0.658  &0.656 \\
SIFT-1M  &0.896 &\bf 0.898  &0.895 &0.885 &\bf 0.887  &0.885 \\
\hline \\
\hline

\multirow{2}{*}{Dataset} & \multicolumn{3}{c|}{Mean Average Precision (mAP)} & \multicolumn{3}{c}{Precision-Recall} \\
\cline{2-7}
& StructH-N & StructH-N-Stage & StructH-SN-Stage & StructH-N & StructH-N-Stage & StructH-SN-Stage \\
\hline
STL10  &0.331 &0.332  &\bf 0.339 &0.267 &0.271  &\bf 0.275 \\
USPS   &\bf 0.868 &0.862  &0.861 &\bf 0.776 &0.774  &0.775 \\
MNIST  &\bf 0.802 &0.786  &0.790 &\bf 0.591 &0.581  &0.584 \\
CIFAR &0.294  &0.299  &\bf 0.305 &0.105 &0.118  &\bf 0.119 \\
ISOLET   &\bf 0.836 &0.828  &0.819 &\bf 0.759 &\bf 0.759  &0.751 \\
TINY-580K  &0.428 &\bf 0.447  &0.445  &0.144  &\bf 0.187  &0.184 \\
SIFT-1M   &0.678  &\bf 0.685  &0.683 &0.363 &\bf 0.390  &0.389 \\
\hline

  \end{tabular}
  }
\label{tab:cmp-speedup}
\end{table}

\begin{figure}[t]
\begin{tabular}{ccc}
    \centering
   \includegraphics[width=.3\linewidth, height=.24\linewidth]{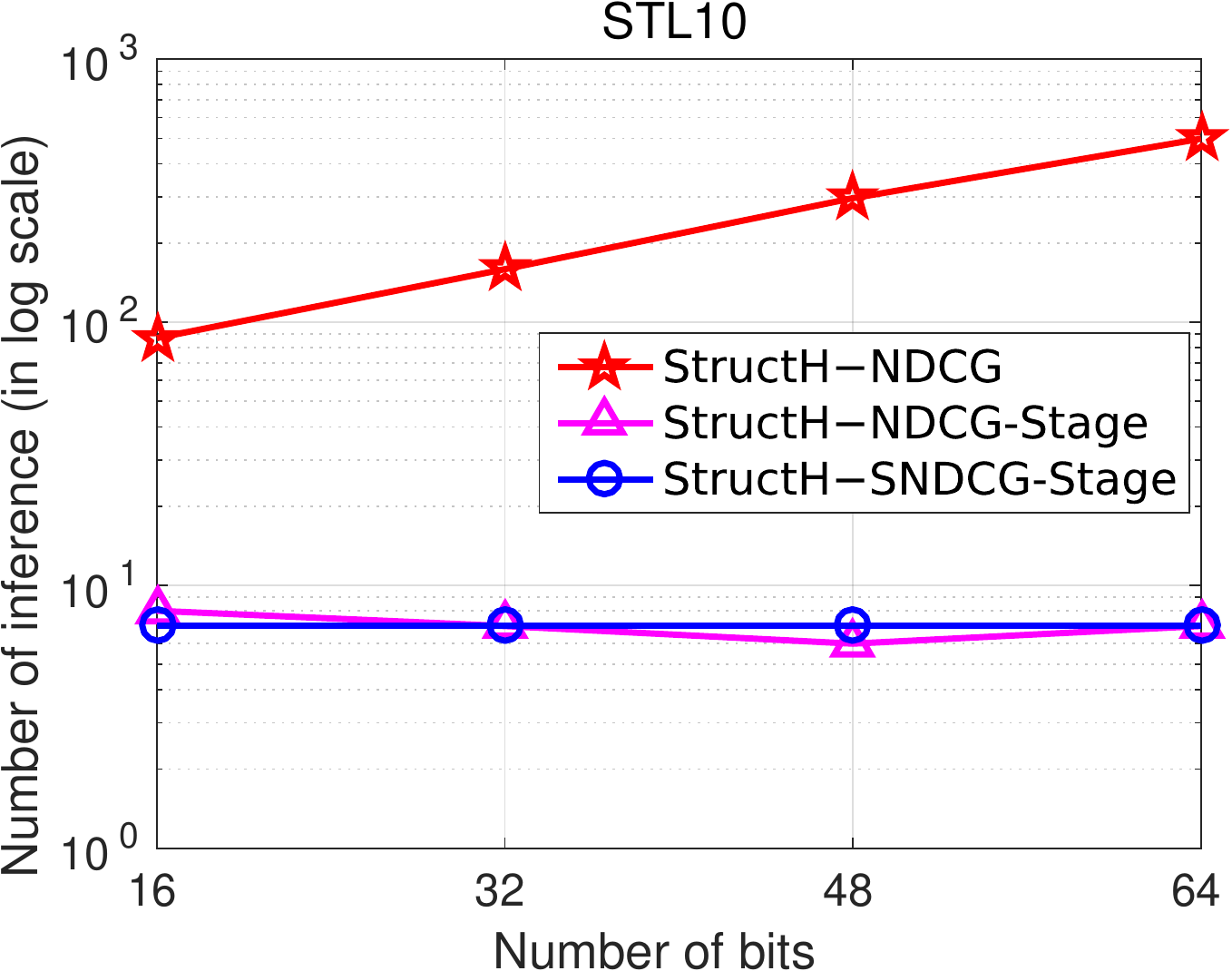}
   &\includegraphics[width=.3\linewidth, height=.24\linewidth]{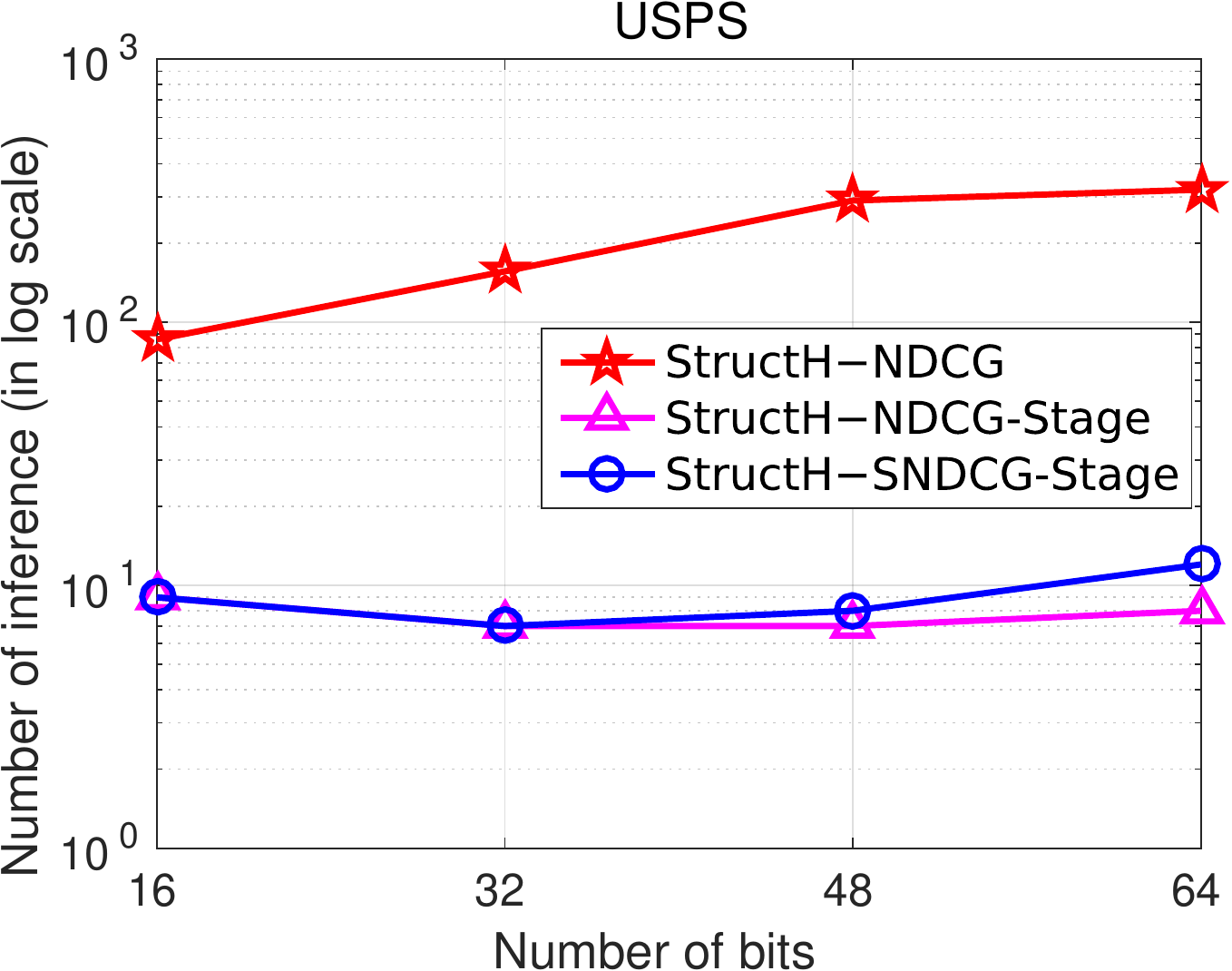}
   &\includegraphics[width=.3\linewidth, height=.24\linewidth]{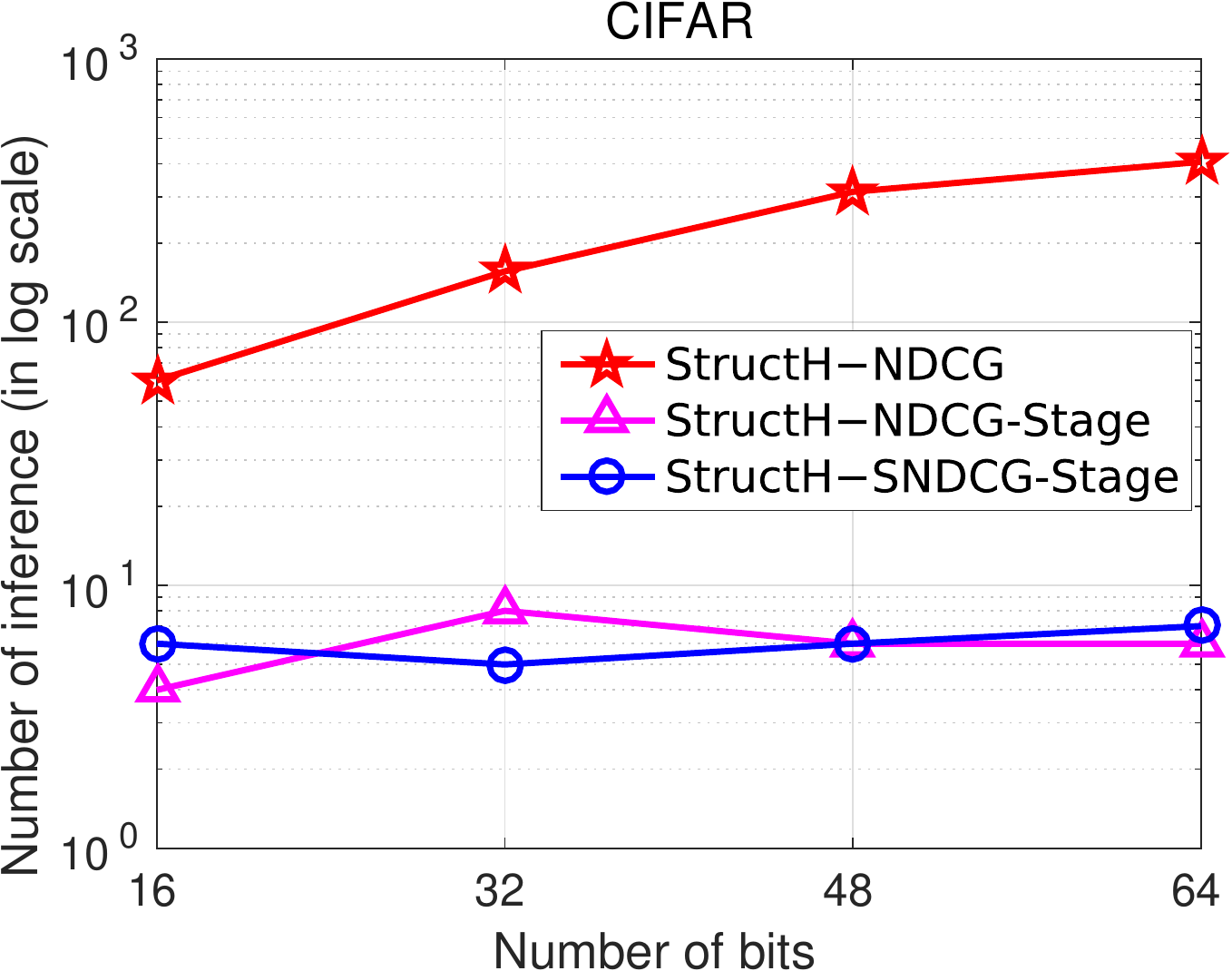} \\

\end{tabular}
 \caption{Comparisons of the number (in log scale) of inference performed in  different hashing bits on 3 datasets. The number of inference iterations for StructHash with efficient stag-wise training  generally is orders of magnitude less than that for the original StructHash.}
    \label{fig:cmp_infer}
\end{figure}

We also present the number of inference iterations performed in different hashing bits and the average time for each inference iteration, as well as the total training time (64-bit) in Table \ref{tab:cmp_time}.
As can be observed, the stage-wise training vastly reduces the inference iterations in each bit, therefore bringing orders of magnitude training speedup. 
As for the average time for each inference iteration, by using unweighted hamming distances in the stage-wise training, StructHash-NDCG-Stage consumes less computation time than the StructHash-NDCG. 
Compared to optimizing the NDCG score, optimizing the SNDCG score further reduces the inference time.   
Fig. \ref{fig:cmp_infer} plots the number of inference (in log scale) performed in different hashing bits.
It explains that the speedup of the stage-wise training is brought by the greatly reduced inference iterations performed in each bit.

\subsubsection{Training on large-scale datasets}
We further evaluate the more efficient models, \ie, the StructH-NDCG-Stage and the StructH-SNDCG-stage, on two large-scale datasets, namely, the TINY-580K and the SIFT-1M.
The results are presented in Table \ref{tab:cmp-speedup}.  
As can be observed, the StructHash with stage-wise training outperforms the original StructHash model.
Given that stage-wise \sh uses unweighted hamming distance, this may indicate that the learned hash functions are more important than the weights.
We also observe that, optimizing the SNDCG loss with stage-wise training performs on par with optimizing the original NDCG loss.

\subsubsection{Computational complexity}
\begin{figure}[t]
    \centering
    \includegraphics[width=.4\linewidth, height=.3\linewidth]{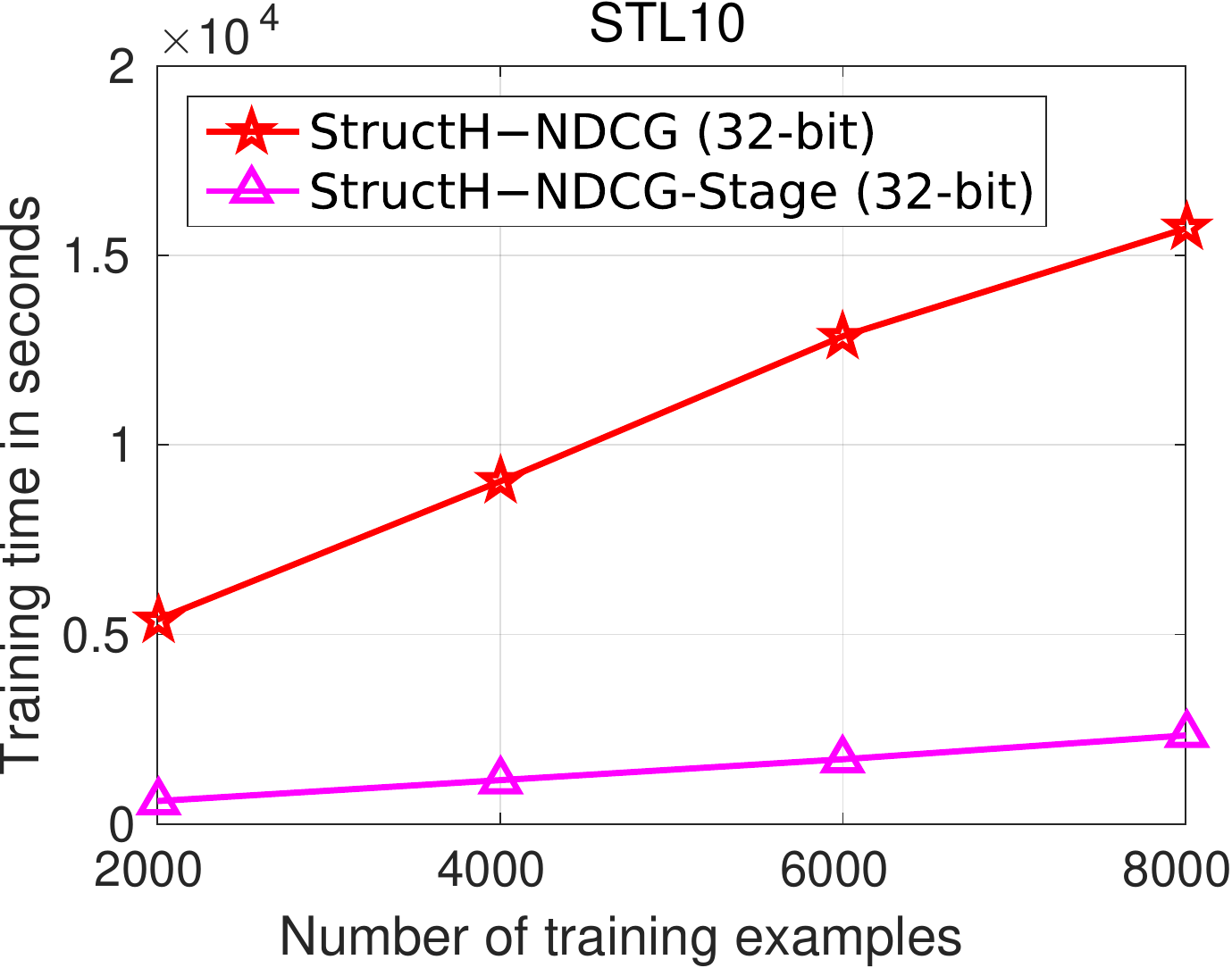}
   \includegraphics[width=.4\linewidth, height=.3\linewidth]{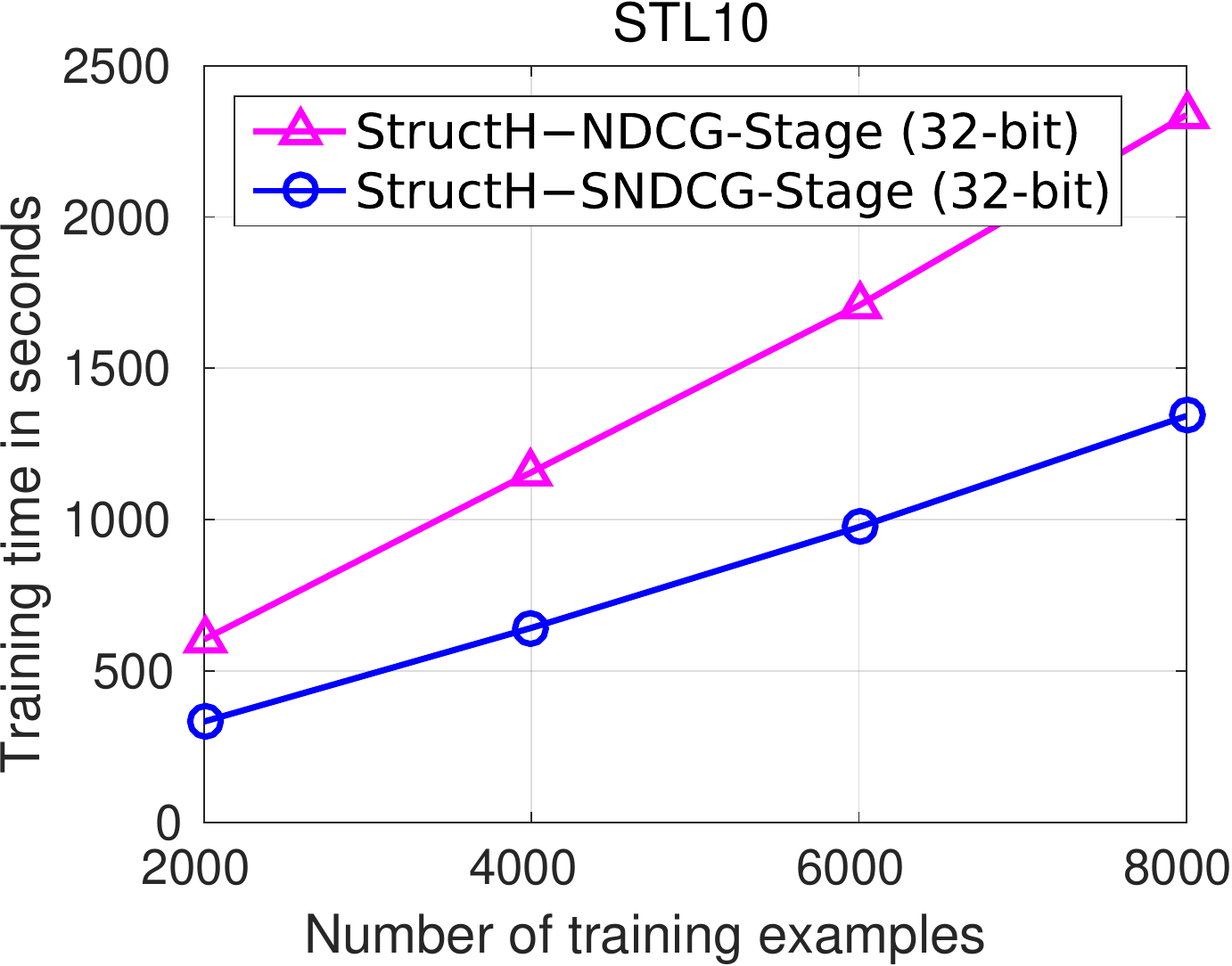} 
     \\
 \caption{Comparisons of the training time (in second) by varying the number of training examples on the STL10 dataset. 
 The left figure compares the stage-wise training with the original training of \sh, and  it shows that stage-wise training is orders of magnitudes more efficient.
 The right figure compares using simplified NDCG loss and the original NDCG loss, and clearly simplified NDCG loss is significantly more efficient.
 }
    \label{fig:cmp_trn_num}
\end{figure}

To show the scalability of the two more efficient extensions of \sh, we present the training time by varying the number of training examples in Fig. \ref{fig:cmp_trn_num}. 
We report the training time of learning 32-bit hash functions.
We also compare StructHash using stage-wise training with the original StructHash, in the left plot of Fig. \ref{fig:cmp_trn_num}.
As we can see, compared to the original StructHash model, the stage-wise training brings orders of magnitude speedup. 
The right plot in Fig.~\ref{fig:cmp_trn_num} 
compares using simplified NDCG loss and the original NDCG loss, and clearly simplified NDCG loss is significantly more efficient.

\begin{table}[t]
\caption{Comparisons on the computation time (in second) and number of inference performed in different hashing bits.
It shows that the efficient stage-wise training (StructH-N-Stage) requires much less inference iteration than the original training of \sh (StructH-N).
The inference time for using the simplified NDCG (StructH-SN-Stage) is as twice as less than using the original NDCG loss (StructH-N-Stage).
The total training time shows that the most efficient variant of \sh is StructH-SN-Stage.
}
\centering
\resizebox{.93\linewidth}{!}
  {
  \begin{tabular}{ c || c | c c c c | c | c  }
\hline
 \multirow{2}{*}{Dataset} &\multirow{2}{*}{Method} &\multicolumn{4}{c|}{Number of inference} &\multirow{2}{*}{Average time per inference iter} &\multirow{2}{*}{Total training time (64-bit)}\\
\cline{3-6}
& &16-bit &32-bit &48-bit &64-bit  & &\\
\hline
\hline
\multirow{3}{*}{STL10}  &StructH-N &87  &159 &296 &498 &  2.32 &30953.3\\
 &StructH-N-Stage  & 8  &7  &6  &7  &2.16 &1374.8\\
 &StructH-SN-Stage &7 &7  &7  &7  &1.07 &875.9\\
 \hline 
\multirow{3}{*}{USPS}   &StructH-N &86  &156 &290 &319  &1.86 &16695.7\\
 &StructH-N-Stage  &9 &7  &7  &8 &1.76 &1162.2\\
 &StructH-SN-Stage &9 &7  &8  &12 &0.75 &719.4\\
 \hline 
\multirow{3}{*}{MNIST}  &StructH-N &139 &205   &496 &294 &1.92 &30590.1 \\
 &StructH-N-Stage  &9 &7  &10 &9 &1.81 &1151.9\\
 &StructH-SN-Stage &9 &8  &7  &8 &0.73 &710 \\
 \hline
\multirow{3}{*}{CIFAR}  &StructH-N &60  &156    &313    &407 &2.64 &25670.2  \\
 &StructH-N-Stage  &4 &8  &6  &6 &2.38 &1348\\
 &StructH-SN-Stage &6 &5  &6  &7 &1.32 &887.1\\
 \hline 
\multirow{3}{*}{ISOLET}   &StructH-N &149 &246    &391    &195 &1.83 &28364.3\\
 &StructH-N-Stage  &7 &6  &9  &6 &1.68 &1038.9 \\
 &StructH-SN-Stage &9 &8  &8  &7 &0.63 &557.2 \\
\hline
  \end{tabular}
  }
\label{tab:cmp_time}
\end{table}

%% file: draft.bbl
\begin{thebibliography}{10}

\bibitem{kulis2009learning}
B.~Kulis and T.~Darrell.
\newblock Learning to hash with binary reconstructive embeddings.
\newblock {\em Proc. Adv. Neural Info. Process. Syst.}, 2009.

\bibitem{weiss2008spectral}
Y.~Weiss, A.~Torralba, and R.~Fergus.
\newblock Spectral hashing.
\newblock In {\em Proc. Adv. Neural Info. Process. Syst.}, 2008.

\bibitem{ICCV13Lin}
G.~Lin, C.~Shen, D.~Suter, and A.~{van den Hengel}.
\newblock A general two-step approach to learning-based hashing.
\newblock In {\em Proc. Int. Conf. Comp. Vis.}, Sydney, Australia, 2013.

\bibitem{CVPR13aShen}
F.~Shen, C.~Shen, Q.~Shi, A.~{van den Hengel}, and Z.~Tang.
\newblock Inductive hashing on manifolds.
\newblock In {\em Proc. Int. Conf. Comp. Vis. \& Patt. Recogn.}, Oregon, USA,
  2013.

\bibitem{liu2011hashingGraphs}
W.~Liu, J.~Wang, S.~Kumar, and S.~F. Chang.
\newblock Hashing with graphs.
\newblock In {\em Proc. Int. Conf. Mach. Learn.}, 2011.

\bibitem{CVPR14Lin}
G.~Lin, C.~Shen, Q.~Shi, A.~{van den Hengel}, and D.~Suter.
\newblock Fast supervised hashing with decision trees for high-dimensional
  data.
\newblock In {\em Proc. Int. Conf. Comp. Vis. \& Patt. Recogn.}, Columbus,
  Ohio, USA, 2014.

\bibitem{torralba2008small}
A.~Torralba, R.~Fergus, and Y.~Weiss.
\newblock Small codes and large image databases for recognition.
\newblock In {\em Proc. Int. Conf. Comp. Vis. \& Patt. Recogn.}, 2008.

\bibitem{wang2010semi}
J.~Wang, S.~Kumar, and S.F. Chang.
\newblock Semi-supervised hashing for large scale search.
\newblock {\em IEEE Trans. Patt. Anal. \& Mach. Intelli.}, 2012.

\bibitem{Strecha2012}
C.~Strecha, A.~Bronstein, M.~Bronstein, and P.~Fua.
\newblock {LDAH}ash: Improved matching with smaller descriptors.
\newblock {\em IEEE Trans. Patt. Anal. \& Mach. Intelli.}, 2012.

\bibitem{fastdection}
T.~Dean, M.~A. Ruzon, M.~Segal, J.~Shlens, S.~Vijayanarasimhan, and J.~Yagnik.
\newblock Fast, accurate detection of 100,000 object classes on a single
  machine.
\newblock In {\em Proc. Int. Conf. Comp. Vis. \& Patt. Recogn.}, 2013.

\bibitem{schultz2004learning}
M.~Schultz and T.~Joachims.
\newblock Learning a distance metric from relative comparisons.
\newblock In {\em Proc. Adv. Neural Information Processing Systems}, 2004.

\bibitem{shen12jmlr}
C.~Shen, J.~Kim, L.~Wang, and A.~{van den Hengel}.
\newblock Positive semidefinite metric learning using boosting-like algorithms.
\newblock {\em J. Machine Learning Research}, 2012.

\bibitem{mcfee10_mlr}
B.~McFee and {G. R. G.} Lanckriet.
\newblock Metric learning to rank.
\newblock In {\em Proc. Int. Conf. Mach. Learn.}, 2010.

\bibitem{Joachims2005}
T.~Joachims.
\newblock A support vector method for multivariate performance measures.
\newblock In {\em Proc. Int. Conf. Mach. Learn.}, 2005.

\bibitem{JarvelinK00}
K.~J{\"a}rvelin and J.~Kek{\"a}l{\"a}inen.
\newblock {IR} evaluation methods for retrieving highly relevant documents.
\newblock In {\em Proc. ACM Conf. SIGIR}, 2000.

\bibitem{zhangSTHs}
D.~Zhang, J.~Wang, D.~Cai, and J.~Lu.
\newblock Extensions to self-taught hashing: kernelisation and supervision.
\newblock In {\em Proc. ACM Conf. SIGIR Workshop}, 2010.

\bibitem{KSH}
W.~Liu, J.~Wang, R.~Ji, Y.G. Jiang, and S.F. Chang.
\newblock Supervised hashing with kernels.
\newblock In {\em Proc. Int. Conf. Comp. Vis. \& Patt. Recogn.}, 2012.

\bibitem{ICML13a}
X.~Li, G.~Lin, C.~Shen, A.~{van den Hengel}, and Anthony Dick.
\newblock Learning hash functions using column generation.
\newblock In {\em Proc. Int. Conf. Mach. Learn.}, 2013.

\bibitem{LinEccv14}
G.~Lin, C.~Shen, and J.~Wu.
\newblock Optimizing ranking measures for compact binary code learning.
\newblock In {\em Proc. Eur. Conf. Comp. Vis.}, 2014.

\bibitem{Gionis1999}
A.~Gionis, P.~Indyk, and R.~Motwani.
\newblock Similarity search in high dimensions via hashing.
\newblock In {\em Proc. Int. Conf. Very Large Data Bases}, 1999.

\bibitem{gong2012iterative}
Y.~Gong, S.~Lazebnik, A.~Gordo, and F.~Perronnin.
\newblock Iterative quantization: a procrustean approach to learning binary
  codes for large-scale image retrieval.
\newblock {\em IEEE Trans. Patt. Anal. \& Mach. Intelli.}, 2012.

\bibitem{KLSH}
B.~Kulis and K.~Grauman.
\newblock Kernelized locality-sensitive hashing.
\newblock {\em IEEE Trans. Patt. Anal. \& Mach. Intelli.}, 2012.

\bibitem{Norouzi11}
Mohammad Norouzi and David~J. Fleet.
\newblock Minimal loss hashing for compact binary codes.
\newblock In {\em Proc. Int. Conf. Mach. Learn.}, pages 353--360, 2011.

\bibitem{Norouzi12}
Mohammad Norouzi, David~J. Fleet, and Ruslan Salakhutdinov.
\newblock Hamming distance metric learning.
\newblock In {\em Proc. Adv. Neural Info. Process. Syst.}, 2012.

\bibitem{Wang15}
Qifan Wang, Zhiwei Zhang, and Luo Si.
\newblock Ranking preserving hashing for fast similarity search.
\newblock In {\em Proc. Int. Joint Conf. Artif. Intelli.}, 2015.

\bibitem{Shalit12}
Uri Shalit, Daphna Weinshall, and Gal Chechik.
\newblock Online learning in the embedded manifold of low-rank matrices.
\newblock {\em Journal of Mach. Learn. Res.}, 2012.

\bibitem{lim2014efficient}
Daryl Lim and Gert Lanckriet.
\newblock Efficient learning of mahalanobis metrics for ranking.
\newblock In {\em Proc. Int. Conf. Mach. Learn.}, 2014.

\bibitem{weston2010large}
Jason Weston, Samy Bengio, and Nicolas Usunier.
\newblock Large scale image annotation: learning to rank with joint word-image
  embeddings.
\newblock {\em Machine learning}, 2010.

\bibitem{lpboost}
Ayhan Demiriz, Kristin~P. Bennett, and John Shawe{-}Taylor.
\newblock Linear programming boosting via column generation.
\newblock {\em Mach. Learn.}, 2002.

\bibitem{shen10}
Chunhua Shen and Hanxi Li.
\newblock On the dual formulation of boosting algorithms.
\newblock {\em IEEE Trans. Patt. Anal. \& Mach. Intelli.}, 2010.

\bibitem{Shen2014SBoosting}
C.~Shen, G.~Lin, and A.~{van den Hengel}.
\newblock {StructBoost}: {B}oosting methods for predicting structured output
  variables.
\newblock {\em IEEE Trans. Patt. Anal. \& Mach. Intelli.}, 2014.

\bibitem{MDSH}
Y.~Weiss, R.~Fergus, and A.~Torralba.
\newblock Multidimensional spectral hashing.
\newblock In {\em Proc. Eur. Conf. Comp. Vis.}, 2012.

\bibitem{lbfgs}
C.~Zhu, R.~H. Byrd, P.~Lu, and J.~Nocedal.
\newblock Algorithm {778: L-BFGS-B}: Fortran subroutines for large-scale
  bound-constrained optimization.
\newblock {\em ACM T. Math. Softw.}, 1997.

\bibitem{demiriz2002linear}
A.~Demiriz, K.~P. Bennett, and J.~Shawe-Taylor.
\newblock Linear programming boosting via column generation.
\newblock {\em Mach. Learn.}, 2002.

\bibitem{boyd}
S.~Boyd and L.~Vandenberghe.
\newblock {\em Convex Optimization}.
\newblock Cambridge University Press, 2004.

\bibitem{kelley1960}
Jr. Kelley, J.~E.
\newblock The cutting-plane method for solving convex programs.
\newblock {\em J. Society for Industrial \& Applied Math.}, 1960.

\bibitem{JoachimsSVM}
T.~Joachims.
\newblock Training linear {SVM}s in linear time.
\newblock In {\em Proc. ACM Knowledge Discovery \& Data Mining}, 2006.

\bibitem{Shen2011Totally}
C.~Shen and H.~Li.
\newblock On the dual formulation of boosting algorithms.
\newblock {\em IEEE Trans. Patt. Anal. \& Mach. Intelli.}, 2010.

\bibitem{mosek}
MOSEK ApS.
\newblock {\em The MOSEK optimization toolbox for MATLAB manual. Version 7.1
  (Revision 28).}, 2015.

\bibitem{Chakrabarti2008}
S.~Chakrabarti, R.~Khanna, U.~Sawant, and C.~Bhattacharyya.
\newblock Structured learning for non-smooth ranking losses.
\newblock In {\em Proc. ACM Knowledge Discovery \& Data Mining}, 2008.

\bibitem{Yue2007}
Y.~Yue, T.~Finley, F.~Radlinski, and T.~Joachims.
\newblock A support vector method for optimizing average precision.
\newblock In {\em Proc. ACM Conf. SIGIR}, 2007.

\bibitem{jae2012}
J.~Heo, Y.~Lee, J.~He, S.~Chang, and S.~Yoon.
\newblock Spherical hashing.
\newblock In {\em Proc. Int. Conf. Comp. Vis. \& Patt. Recogn.}, 2012.

\end{thebibliography}
